\pdfoutput=1
\documentclass{article}

    \PassOptionsToPackage{numbers, sort, compress}{natbib}

\usepackage[final]{neurips_2021}

\usepackage[utf8]{inputenc} 
\usepackage[T1]{fontenc}    
\usepackage{hyperref}       
\usepackage{url}            
\usepackage{booktabs}       
\usepackage{amsfonts}       
\usepackage{nicefrac}       
\usepackage{microtype}      

\usepackage[dvipsnames]{xcolor}

\usepackage{amsopn,amsmath,amsthm,amssymb}
\usepackage{wrapfig}
\usepackage{multirow}
 \usepackage{mathtools}

\usepackage{anyfontsize}

\usepackage{color}
\usepackage{tikz}
\usetikzlibrary{arrows,shapes,snakes,automata,backgrounds,fit,petri}
\usepackage{adjustbox}

\newcommand{\RN}[1]{%
	\textup{\lowercase\expandafter{\it \romannumeral#1}}%
}

\makeatletter
\newcommand{\distas}[1]{\mathbin{\overset{#1}{\kern\z@\sim}}}%

\usepackage[shortlabels]{enumitem}

\usepackage{algorithm}
\usepackage{algorithmic}

\newcommand{\beq}{\vspace{0mm}\begin{equation}}
\newcommand{\eeq}{\vspace{0mm}\end{equation}}
\newcommand{\beqs}{\vspace{0mm}\begin{eqnarray}}
\newcommand{\eeqs}{\vspace{0mm}\end{eqnarray}}
\newcommand{\barr}{\begin{array}}
\newcommand{\earr}{\end{array}}

\newtheorem{theorem}{Theorem} 

\ifx\assumption\undefined
\newtheorem{assumption}{Assumption}
\fi

\ifx\definition\undefined
\newtheorem{definition}{Definition}
\fi

\ifx\remark\undefined
\newtheorem{remark}{Remark}
\fi

\usepackage{graphicx}
\usepackage{subfigure}
\usepackage{bbm}
\usepackage{multirow}
\usepackage{amssymb}
\usepackage{float}
\usepackage{verbatim}
\usepackage{amsmath}
\usepackage{amsthm}
\usepackage{mathtools}
\usepackage{longtable}
\usepackage{wrapfig}
\allowdisplaybreaks[4]

\newtheorem*{definition*}{Definition}

\def\*#1{\boldsymbol{#1}}

\newcommand{\necessary}{\Box}
\newcommand{\possible}{\Diamond}
\newcommand\algA{$\mathcal{J}$}
\newcommand\algB{$\mathcal{K}$}

\newcommand{\workingtitle}{Hyperparameter Optimization Is Deceiving Us,

and How to Stop It}

\title{\workingtitle}

\author{%
  A. Feder Cooper\thanks{Corresponding author: \url{https://cacioepe.pe}} \\
  Cornell University\\
  \texttt{afc78@cornell.edu} \\
   \And
   Yucheng Lu \\
   Cornell University\\
  \texttt{yl2967@cornell.edu} \\
  \And
   Jessica Zosa Forde \\
   Brown University\\
  \texttt{jforde2@cs.brown.edu} \\
  \And
    Christopher De Sa\\
    Cornell University \\
    \texttt{cdesa@cs.cornell.edu} \\
}

\begin{document}

\maketitle

\begin{abstract}
Recent empirical work shows that inconsistent results based on choice of hyperparameter optimization (HPO) configuration are a widespread problem in ML research. When comparing two algorithms \algA{} and \algB, searching one subspace can yield the conclusion that \algA{} outperforms \algB, whereas searching another can entail the opposite. In short, the way we choose hyperparameters can deceive us. We provide a theoretical complement to this prior work, arguing that, to avoid such deception, the process of drawing conclusions from HPO should be made more rigorous. We call this process \emph{epistemic hyperparameter optimization} (EHPO), and put forth a logical framework to capture its semantics and how it can lead to inconsistent conclusions about performance. Our framework enables us to prove EHPO methods that are guaranteed to be defended against deception, given bounded compute time budget $t$. We demonstrate our framework's utility by proving and empirically validating a defended variant of random search. 
\end{abstract}

\section{Introduction}\label{sec:introduction}

Machine learning can be informally thought of as a double-loop optimization problem. The \emph{inner loop}
is what is typically called \emph{training}: It learns the parameters
of some model by running a training algorithm on a training set. This is usually done to minimize some training loss function via an algorithm such as stochastic gradient descent (SGD). Both the inner-loop training algorithm and the model are parameterized by a vector of \emph{hyperparameters} (HPs). Unlike the learned output parameters of a ML model, HPs are inputs provided to the learning algorithm that guide the learning process, such as learning rate and network size. The \emph{outer-loop} optimization problem is to find HPs (from a set of allowable HPs) that result in a trained model that performs the best in expectation on ``fresh'' examples drawn from the same source as the training set, as measured by some loss or loss approximation. An algorithm that attempts this task is called a \emph{hyperparameter optimization} (HPO) procedure~\cite{claesen2015hyperparameter, feurer2019optimization}.

From this setup comes the natural question: How do we pick the subspace for the HPO procedure to search over? The HPO search space is enormous, suffering from the curse of dimensionality; training, which is also expensive, has to be run for each HP configuration tested. Thus, we have to make hard choices. With limited compute resources, 
we typically pick a small subspace of possible 
HPs and perform grid search or random search over that subspace. This involves comparing the empirical performance of the resulting trained models, and then reporting on the model that performs best in terms of a chosen validation metric~\cite{feurer2019optimization, john1994grid, hsu2003svm}. For grid search, the grid points are often manually set to values put forth in now-classic papers as good rules-of-thumb concerning, for example, how to set the learning rate~\cite{larochelle2007empirical, lecun1998backprop, hinton2010boltzmann, pedregosa2011scikit}. In other words, how we choose which HPs to test can seem rather ad-hoc. We may have a good rationale in mind, but we often elide the details of that rationale on paper; we choose an HPO configuration without explicitly justifying our choice.

Much recent empirical work has critiqued this practice~\cite{Sculley2018-le, dodge2019nlp, musgrave2020metric, bouthillier19reproducible, sivaprasad2019optimizer, choi2019empirical, melis2018evaluation, lipton2018troubling}. The authors examine HPO configuration choices in prior work, and find that those choices can have an outsize impact on convergence, correctness, and generalization. They therefore argue that more attention should be paid to the origins of empirical gains in ML, as it is often difficult to tell whether measured improvements are attributable to training 
or to well-chosen (or lucky) HPs. 
Yet, this empirical work does not suggest a path forward for formalizing this problem or addressing it theoretically. 

To this end, \textbf{we argue that the process of drawing conclusions using HPO should itself be an object of study}. Our contribution is to put forward, to the best of our knowledge, the first theoretically-backed characterization for making trustworthy conclusions about algorithm performance using HPO. We model theoretically the following empirically-observed problem: When comparing two algorithms, $\mathcal{J}$ and $\mathcal{K}$, searching one subspace can pick HPs that yield the conclusion that $\mathcal{J}$ outperforms $\mathcal{K}$, whereas searching another can select HPs that entail the opposite result. In short, the way we choose hyperparameters can deceive us---a problem that we call \emph{hyperparameter deception}. We formalize this problem, and prove and empirically validate a defense against it. Importantly, our proven defense does not make any promises about ground-truth algorithm performance; rather, it is guaranteed to avoid the possibility of drawing inconsistent conclusions about algorithm performance within some bounded HPO time budget $t$.  In summary, we:
\begin{itemize}[topsep=0pt, leftmargin=.25cm]
    \item Formalize the process of drawing conclusions from HPO (epistemic HPO, Section \ref{sec:ehpo}).
    \item Leverage the flexible semantics of modal logic to construct a framework for reasoning rigorously about 1) uncertainty in epistemic HPO, and 2) how this uncertainty can mislead the conclusions drawn by even the most well-intentioned researchers (Section \ref{sec:formalizing}). 
    \item Exercise our logical framework to demonstrate that it naturally suggests defenses with guarantees against being deceived by EHPO, and offer a specific, defended-random-search EHPO (Section \ref{sec:defense}).
\end{itemize}

\section{Preliminaries: Problem Intuition and Prevalence in ML Research}\label{sec:prelim}
Principled HPO methods include \emph{grid search}~\cite{john1994grid} and \emph{random search}~\cite{bergstra2012random}. For the former, we perform HPO on a grid of HP-values, constructed by picking a set for each HP and taking the Cartesian product. For the latter, the HP-values 
are randomly sampled from chosen distributions. Both of these HPO algorithms are parameterized themselves: Grid search requires inputting the spacing between different configuration points in the grid, and random search requires distributions from which to sample. We call these HPO-procedure-input values \emph{hyper-hyperparameters} (hyper-HPs).\footnote{We provide a glossary of all definitions and symbols for reference at the beginning of the Appendix.} To make HPO outputs comparable, we also introduce the notion of a \emph{log}:

\begin{definition}\label{def:log} A log $\ell$ records all the choices and measurements made during an HPO run, including the total time $T$ it took to run. It has all necessary information to make the HPO run reproducible.
\end{definition}

A log can be thought of as everything needed to produce a table in a research paper: code, random seed, choice of hyper-HPs, information about the learning task, properties of the learning algorithm, all of the observable results. We formalize all of the randomness in HPO in terms of a random seed $r$ and a pseudo-random number generator (PRNG) $G$. Given a seed, $G$ deterministically produces a sequence of pseudo-random numbers: 
all numbers lie in some set $\mathcal{I}$ (typically 64-bit integers), i.e. $r \in \mathcal{I}$ and PRNG $G: \mathcal{I} \rightarrow \mathcal{I}^{\infty}$. With this, we can now define HPO formally:
\begin{definition}
	\label{def:hpo}
	An HPO procedure $H$ is a tuple $(H_{*}, \mathcal{C}, \Lambda, \mathcal{A}, \mathcal{M}, G, X)$ where $H_{*}$ is a randomized algorithm, $\mathcal{C}$ is a set of allowable hyper-HPs (i.e., allowable configurations for $H_{*}$), $\Lambda$ is a set of allowable HPs (i.e., of HP sets $\lambda$), $\mathcal{A}$ is a training algorithm (e.g. SGD), $\mathcal{M}$ is a model (e.g. VGG16), $G$ is a PRNG, and $X$ is some dataset (usually split into train and validation sets). When run, $H_{*}$ takes as input a hyper-HP configuration $c \in \mathcal{C}$ and a random seed $r \in \mathcal{I}$, then proceeds to run $\mathcal{A}_{\lambda}$ (on $\mathcal{M}_{\lambda}$ using $G(r)$ and data\footnote{Definition \ref{def:hpo} does not preclude cross-validation, as this can be part of $H_{*}$. The input dataset $X$ can be split in various ways, as a function of the random seed $r$.} from $X$) some number of times for different HPs $\lambda \in \Lambda$. Finally, $H_{*}$ outputs a tuple $(\lambda^*, \ell)$, where $\lambda^*$ is the HP configuration chosen by HPO and $\ell$ is the log documenting the run.
\end{definition}


Running $H$ is a crucial part of model development. As part of an empirical, scientific procedure, we specify different training algorithms and a learning task, run potentially many HPO passes, and try to make general conclusions about overall algorithm performance. That is, we aim to develop knowledge regarding whether one of the algorithms outperforms the others. However,
recent empirical findings indicate that it is actually really challenging to pick hyper-HPs that yield reliable knowledge about general algorithm performance. In fact, it is a surprisingly common occurrence to be able to draw inconsistent conclusions based on our choice of hyper-HPs \cite{choi2019empirical, dodge2019nlp, sivaprasad2019optimizer, lipton2018troubling}. 

\begin{figure}[h!]
 \begin{center}

    \includegraphics[width=0.44\textwidth]{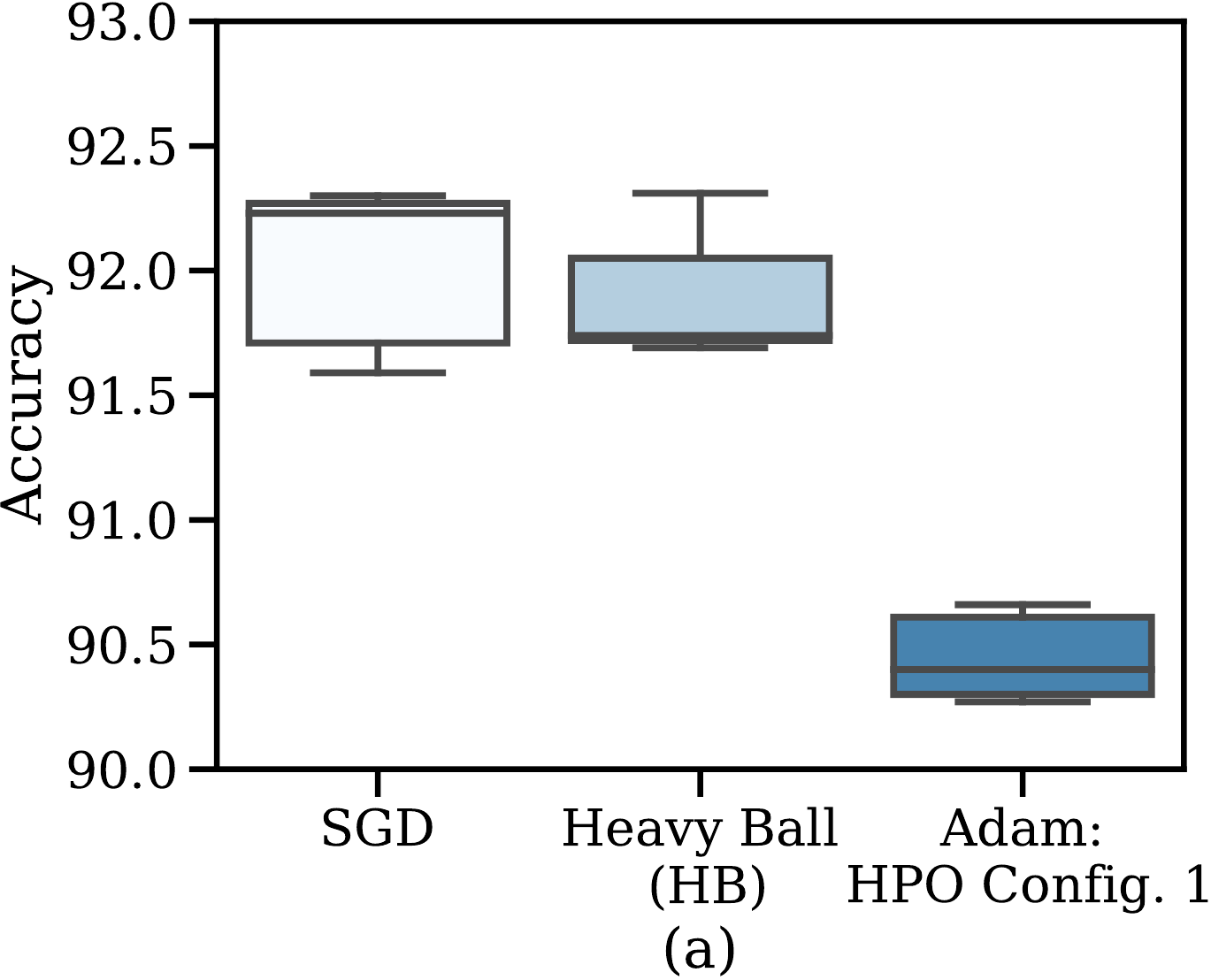} \hspace{.5cm}
\includegraphics[width=0.44\textwidth]{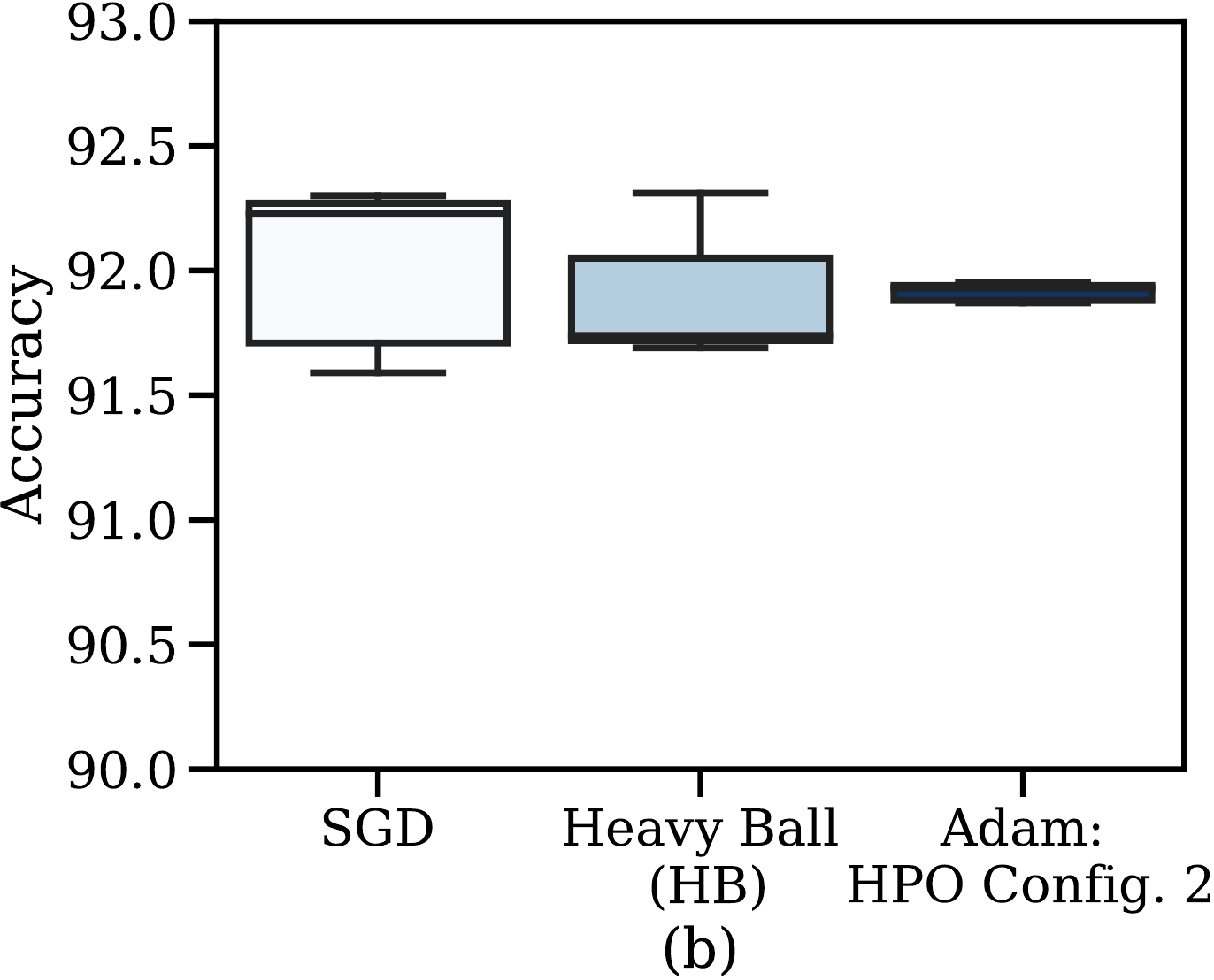}
 \vspace{-.2cm}

    \caption{Demonstrating the possibility of drawing inconsistent conclusions from HPO (what we shorthand \emph{hyperparameter deception}) when training VGG16 on CIFAR-10. Each box plot represents a log. In {(a)}, we replicate \citet{wilson2017marginal} and show the best-performing results:  One can reasonably conclude that Adam under-performs non-adaptive methods. In (b), we change the HPO search space for Adam, and similarly show the best-performing results: In contradiction, one can reasonably conclude that Adam performs just as well as non-adaptive methods in terms of test accuracy.}
    \label{fig:wilson_deception}
  \end{center}
\end{figure}

\textbf{An example illustrating the possibility of drawing inconsistent conclusions from HPO.} As a first step to studying HPO as a procedure for developing reliable knowledge, 
we provide an example of how being inadvertently deceived by HPO is a real problem, even in excellent research (we give an additional example in the Appendix).\footnote{All code can be found at \url{https://github.com/pasta41/deception}.} 
We first reproduce~\citet{wilson2017marginal}, in which the authors trained VGG16 with different optimizers on CIFAR-10 (Figure \ref{fig:wilson_deception}a). This experiment uses grid search, with a powers-of-2 grid for the learning rate $\alpha$ crossed with the default HPs for Adam. Based on the best-performing HPO per algorithm ($\alpha=1$), it is reasonable to conclude that non-adaptive methods (e.g., SGD) perform better than adaptive ones (e.g., Adam~\cite{kingma2014adam}), as the non-adaptive optimizers demonstrate higher test accuracy. 

However, this setting of grid search's hyper-HPs directly informs this particular conclusion; using different hyper-HPs makes it possible to conclude the opposite. Inspired by~\citet{choi2019empirical}, we perform grid search over a different subspace, tuning both learning rate and Adam's $\epsilon$ parameter. Our results entail the logically opposite conclusion: Non-adaptive methods \emph{do not} outperform adaptive ones. Rather, when choosing the HPs that maximize test accuracy, all of the optimizers essentially have equivalent performance (Figure \ref{fig:wilson_deception}b, Appendix). Notably, as we can see from the confidence intervals in Figure \ref{fig:wilson_deception}, \textbf{satisfying statistical significance is not sufficient to avoid being deceived about comparative algorithm performance} \cite{young2018crisis}. Thus, we will require additional tools aside from statistical tests to reason about this, which we discuss in Sections \ref{sec:formalizing} \& \ref{sec:defense}.

\vspace{-.2cm}
\vspace{-.2cm}

This example is not exceptional, or even particularly remarkable, in terms of illustrating the hyperparameter deception problem. We simply chose it for convenience: The experiment does not require highly-specialized ML sub-domain expertise to understand it, and it is arguably broadly familiar, as it very well-cited \cite{wilson2017marginal}. However, we emphasize that hyperparameter deception is rather common. Additional examples can be found in numerous empirical studies across ML subfields~\cite{choi2019empirical, sivaprasad2019optimizer, melis2018evaluation, musgrave2020metric, bouthillier19reproducible, schneider2019deepobs, dodge2019nlp, Lucic2018-dr} (Appendix). This work shows that reported results tend to be impressive for the tested hyper-HP configurations, but that modifying HPO can lead to vastly different performance outcomes that entail contradictory conclusions. 
More generally, it is possible to develop results that are wrong about performance, or else correct about performance but for the wrong reasons (e.g., by picking ``lucky'' hyperparameters). Neither of these outcomes constitutes reliable knowledge~\cite{gettier1963justified, lehrer1979gettier}. 
As scientists, this is disheartening. We want to have confidence in the conclusions we draw from our experiments. We want to trust that we are deriving reliable knowledge about algorithm performance. 
\textbf{In the sections that follow, our aim is to study HPO in this reliable-knowledge sense: We want to develop ways to reason rigorously and confidently about how we derive knowledge from empirical investigations involving HPO.}

\section{Epistemic Hyperparameter Optimization}\label{sec:ehpo}

Our discussion in Section~\ref{sec:prelim} shows that applying standard HPO methodologies can be deceptive: Our beliefs about algorithm performance can be controlled by happenstance, wishful thinking, or, even worse, potentially by an adversary trying to trick us with a tampered set of HPO logs. This leaves us in a position where the ``knowledge'' we derived may not be knowledge at all---since we could have easily (had circumstances been different) concluded the opposite. To address this, we propose that the process of drawing conclusions using HPO should itself be an object of study. We formalize this reasoning process, which we call \emph{epistemic hyperparameter optimization} (EHPO), and we provide an intuition for how EHPO can help us think about the hyperparameter deception problem. 

\begin{definition} \label{def:ehpo}
An \textbf{epistemic hyperparameter optimization procedure (EHPO)} is a tuple $(\mathcal{H}, \mathcal{F})$ where $\mathcal{H}$ is a set of HPO procedures $H$ (Definition \ref{def:hpo}) and $\mathcal{F}$ is a function that maps a set of HPO logs $\mathcal{L}$ (Definition \ref{def:log}) to a set of logical formulas $\mathcal{P}$, i.e. $\mathcal{F}(\mathcal{L}) = \mathcal{P}$. An execution of EHPO involves running each $H \in \mathcal{H}$ some number of times (each run produces a log $\ell$), and then evaluating $\mathcal{F}$ on the logs $\mathcal{L}$ produced in order to output the conclusions $\mathcal{F}(\mathcal{L})$ we draw from all of the HPO runs.


\end{definition}


In practice, it is common to run EHPO for two training algorithms, \algA{} and \algB{}, and to compare their performance to conclude which is better-suited for the task at hand. $\mathcal{H}$ contains at least one HPO that runs \algA{} and at least one HPO that runs \algB{}. The possible conclusions in 
output $\mathcal{P}$ include $p$ = ``\algA{} performs better than \algB{}'', and $\lnot p$ = ``\algA{} does not perform better than \algB{}''. 
Intuitively, EHPO is deceptive whenever it could produce $p$ and also could (if configured differently or due to randomness) produce $\lnot p$. That is, we can be deceived if the EHPO procedure we use to derive knowledge about algorithm performance could entail logically inconsistent results. 

Our example in Section \ref{sec:prelim} is deceptive because using different hyper-HP-configured grid searches for $\mathcal{H}$ could produce contradictory conclusions.
We ran two variants of EHPO $(\mathcal{H}, \mathcal{F})$: The first replicated \citet{wilson2017marginal}'s original $\mathcal{H}$ of 3 grid-searches on SGD, HB, and Adam (Figure \ref{fig:wilson_deception}a), and the second used 3 grid-searches with a modified grid search for Adam that also tuned $\epsilon$ (Figure \ref{fig:wilson_deception}b). Each EHPO produced a $\mathcal{L}$ with 3 logs.  
For both, to draw conclusions $\mathcal{F}$ picks the best-performing HP-config per $\mathcal{A}$ and maps them to formulas including ``SGD outperforms Adam." From the 3 logs in Figure \ref{fig:wilson_deception}a, we conclude $p$: ``Non-adaptive optimizers outperform adaptive ones"; from the  3 logs in Figure \ref{fig:wilson_deception}b, we conclude $\lnot p$: ``Non-adaptive methods do not outperform adaptive ones." How can we formally reason about EHPO to avoid this possibility of drawing inconsistent conclusions---to guard against deceiving ourselves about algorithm performance when running EHPO?

\textbf{Framing an adversary who can deceive us.} To begin answering this question, we take inspiration from~\citeauthor{descartes1996evildemon}' deceptive demon thought experiment (Appendix). We frame the problem in terms of a powerful adversary trying to deceive us---one that can cause us to doubt ourselves and our conclusions. Notably, the demon is not a real adversary; rather, it models a worst-case setting of configurations and randomness that are usually set arbitrarily or by happenstance in EHPO. 

Imagine an evil demon who is trying to deceive us about the relative performance of different algorithms via running EHPO. At any time, the demon maintains a set $\mathcal{L}$ of HPO logs, which it can modify either by running an HPO $H \in \mathcal{H}$ with whatever hyper-HPs $c \in \mathcal{C}$ and seed $r \in \mathcal{I}$ it wants (producing a new log $\ell$, which it adds to $\mathcal{L}$) or by erasing some of the logs in its set. Eventually, it stops and presents us with $\mathcal{L}$, from which we will draw some conclusions using $\mathcal{F}$, i.e. $\mathcal{F}(\mathcal{L})$.
The demon's EHPO could deceive us via the conclusions we draw from the set of logs it produces. 
For example, $\mathcal{L}$ may lead us to conclude that one algorithm performs better than another, when in fact picking a different set of hyper-HPs could have generated logs that would lead us to conclude differently. We want to be sure that we will not be deceived by any logs the demon \emph{\textbf{could}} produce. 
Of course, this intuitive definition is lacking: It is not clear what is meant by \emph{\textbf{could}}. Our contribution in the sections that follow is to pin down a formal, reasonable definition of \emph{\textbf{could}} in this context, so that we can suggest an EHPO procedure that can defend against such a maximally powerful adversary. We intentionally imagine such a powerful adversary because, if we can defend against it, then we will also be defended against weaker or accidental deception.


\vspace{-.25cm}
\section{A Logic for Reasoning about EHPO}\label{sec:formalizing}
The informal notion of \emph{\textbf{could}} established above encompasses numerous sources of uncertainty. There is the time to run EHPO and the choices of random seed, algorithms to compare, HPO procedures, hyper-HPs, and learning task. Then, once we have completed EHPO and have a set of logs, we have to digest those logs into logical formulas from which we base our conclusions. This introduces more uncertainty, as we need to reason about whether we believe those conclusions or not. 
Our formalization needs to capture all of these sources of uncertainty, and needs to be sufficiently expressive to capture how they could combine to cause us to believe deceptive conclusions. It needs to be expansive enough to handle the common case---of a well-intentioned researcher with limited resources making potentially incorrect conclusions---and the rarer, worst case---of gaming results.

\textbf{Why not statistics?} As the common toolkit in ML, statistics might seem like the right choice for modeling all this uncertainty. However, statistics is great for reasoning about uncertainty that is \emph{quantifiable}. For this problem, not all of the sources of uncertainty are easily quantifiable. In particular, it is very difficult to quantify the different hyper-HP possibilities. It is not reasonable to model hyper-HP selection as a random process; 
we do not sample from a distribution and, even if we wanted to, it is not clear how we would pick the distribution from which to sample. Moreover, as we saw in our example in Section \ref{sec:prelim}, testing for statistical significance is not sufficient to prevent deception. 
While the results under consideration may be statistically significant, they can still fail to prevent the possibility of yielding inconsistent conclusions. For this reason, when it comes to deception, statistical significance can even give us false confidence in the conclusions we draw.

\textbf{Why modal logic?} Modal logic is the standard mathematical tool for formalizing reasoning about uncertainty~\cite{chellas1980book, emerson1991temporal}---for formalizing the thus far informal notion of what the demon \textit{\textbf{could}} bring about running EHPO. 
It is meant precisely for dealing with different types of uncertainty, particularly uncertainty that is difficult to quantify, and has been successfully employed for decades in AI~\cite{halpern2017book, halpern1991apm, blackburn2006modal}, programming languages~\cite{clarke1986tls, pnueli1977temporal, lamport1980semantics}, and distributed systems~\cite{hawblitzel2015ironfleet, fischer1988uncertaintyds, owicki1982liveness}. In each of these computer science fields, modal logic's flexible semantics has been indispensable for writing proofs about higher-level specifications with multiple sources of not-precisely-quantifiable, lower-level uncertainty. For example, in distributed computing, it lets us write proofs about overall system correctness, abstracting away from the specific non-determinism introduced by each lower-level computing process \cite{fischer1988uncertaintyds}. Analogously, modal logic can capture the uncertainty in EHPO without being prescriptive about particular hyper-HP choices. 
Our notion of correctness, which we want to reason about and guarantee, is not being deceived. Therefore, while modal logic may be an atypical choice for ML, it comes with a huge payoff. By constructing the right semantics, we can capture all the sources of uncertainty described above and we can write simple proofs about whether we can be deceived by the EHPO we run. In Section \ref{sec:defense}, it is this formalization that ultimately enables us to naturally suggest a defense against being deceived.

\subsection{Introducing  our logic: syntax and semantics overview} \label{sec:logicintro}

Modal logic inherits the tools of more-familiar propositional logic and adds two operators:  $\possible$ to represent \emph{possibility} and $\necessary$ to represent \emph{necessity}. These operators enable reasoning about \emph{possible worlds}---a semantics for representing how the world \emph{is} or \emph{could be}, making modal logic the natural choice to express the ``could'' intuition from Section \ref{sec:ehpo}. The well-formed formulas $\phi$ of modal logic are given recursively in Backus-Naur form, where $P$ is any atomic proposition:
\begin{align*}
    \phi \coloneqq P\;|\;\lnot \phi\;|\;\phi \land \phi \;|\; \possible \phi
\end{align*}
$\possible p$ reads, ``It is possible that $p$.'';  $p$ is true at \emph{some} possible world, which we \emph{could} reach (Appendix). Note that $\necessary$ is syntactic sugar, with $\necessary p \equiv \lnot \possible \lnot p$. Similarly, ``or'' has $p \lor q \equiv \lnot (\lnot p \land \lnot q)$ and ``implies'' has $p \rightarrow q \equiv \lnot p \lor q$. The axioms of modal logic are as follows:
\begin{align*}
    \vdash Q &\rightarrow \necessary Q & \textit{(necessitation)}. &&
    \necessary(Q \rightarrow R) &\rightarrow (\necessary Q \rightarrow \necessary R) & \textit{(distribution)}.
\end{align*}

where $Q$ and $R$ are any formula, and $\vdash Q$ means $Q$ is a theorem of propositional logic. We can now provide the syntax and an intuitive notion of the semantics of our logic for reasoning about deception. 

\textbf{Syntax.} Our logic requires an extension of standard modal logic. We need \emph{two} modal operators to reckon with two overarching modalities: the possible results of the demon running EHPO ($\possible_t$) and our beliefs about conclusions from those results ($\mathcal{B}$). Combining these modalities yields well-formed formulas $\psi$ where, for any atomic proposition $P$ and any positive real $t$,
\[
    \psi \coloneqq P\;|\;\lnot \psi\;|\;\psi \land \psi \;|\; \possible_t \psi \;|\; \mathcal{B} \psi
\]
Note the EHPO modal operator here is \emph{indexed}: $\possible_t$ captures  ``how possible'' ($\possible$) something is, quantified by the compute capabilities of the demon ($t$) \cite{blackburn2006modal, emerson1991temporal, heifeitz1998indexed}. 

\textbf{Semantics intuition.}  
We suppose that an EHPO user has in mind some atomic propositions (propositions of the background logic unrelated to possibility or belief, such as 
``the best-performing log for \algA{} has lower loss than the best-performing log for \algB{}'') with semantics that are already defined.
$\land$ and $\lnot$ inherit their semantics from ordinary propositional logic, which can combine propositions to form formulas. A set of EHPO logs $\mathcal{L}$ (Definition \ref{def:log}) can be digested into such logical formulas. That is, we define our semantics using logs $\mathcal{L}$ as models over formulas $p$:  $\mathcal{L} \models p$, which reads ``$\mathcal{L}$ models $p$'', means that $p$ is true for the set of logs $\mathcal{L}$. We will extend this intuition to give semantics for possibility $\possible_t$ (Section \ref{sec:ehpologic}) and belief $\mathcal{B}$ (Section \ref{sec:belieflogic}), culminating in a tool that lets us reason about whether or not EHPO can deceive us by possibly yielding inconsistent conclusions (Section \ref{sec:deceptionlogic}).

\textbf{Using our concrete example to ground us.} To clarify our presentation below, we will map our semantics to the example from Section \ref{sec:prelim}, providing an informal intuition before formal definitions.



\subsection{Expressing the possible outcomes of EHPO using $\possible_t$} \label{sec:ehpologic}
Our formalization for possible EHPO is based on the demon of Section~\ref{sec:ehpo}.
Recall, the demon models a worst-case scenario. In practice, we deal with the easier case of well-intentioned ML researchers.  The notion of possibility we define here gives limits on what possible world a demon \emph{with bounded EHPO time} could reliably bring about. We first define a \emph{strategy} the demon can execute for EHPO:

\begin{definition} \label{def:strategy}
A randomized \textbf{strategy} $\boldsymbol{\sigma}$ is a function that specifies which action the demon will take. Given $\mathcal{L}$, its current set of logs, $\boldsymbol{\sigma(\mathcal{L})}$ gives a distribution over concrete actions, where each action is either 1) running a new $H$ with its choice of hyper-HPs $c$ and seed $r$ 
2) erasing some logs, or 3) returning. We let $\Sigma$ denote the set of all such strategies. 
\end{definition}
The demon we model controls the hyper-HPs $c$ and the random seed $r$, but importantly does \emph{not} fully control the PRNG $G$. From the adversary's perspective, for a strategy $\sigma$ to be reliable it must succeed regardless of the specific $G$. Informally, the demon cannot hack the PRNG.\footnote{We do not consider adversaries that can directly control how data is ordered and submitted to the algorithms under evaluation. This distinction shows that our logical construction non-trivial: We are able to defend against strong adversaries that can game the output of EHPO, which is separate from cheating by hacking the PRNG.}

\textbf{Informally}, we now want to \emph{execute a strategy} to bring about a particular outcome $p$. In Section \ref{sec:prelim}, our good-faith strategy was simple: We ran each $H$ with its own hyper-HPs and random seed, then returned. The demon is trickier: It is adopting a strategy to try to bring about a deceptive outcome. 
\textbf{Formally}, we model the 
demon executing strategy $\sigma$ on logs $\mathcal{L}$ with a PRNG unknown to the demon as follows. Let $\mathcal{G}$ denote the distribution over PRNGs $G: \mathcal{I} \rightarrow \mathcal{I}^{\infty}$, in which all number sequence elements are drawn independently and uniformly from $\mathcal{I}$ (recall, $\mathcal{I}$ is typically the 64-bit integers).  
First, draw $G$ from $\mathcal{G}$, conditioned on $G$ being consistent with all the runs in $\mathcal{L}$.\footnote{i.e., All random events recorded in $\mathcal{L}$ should agree with the corresponding random numbers produced by $G$.}
The demon then performs a random action drawn from $\sigma(\mathcal{L})$, using $G$ as the PRNG when running a new HPO $H$, and continues---updating the working set of logs $\mathcal{L}$ as it goes---until the ``return'' action is chosen. 

Using this process, we define what outcomes $p$ the demon can reliably bring about (i.e., what is possible, $\possible$) in the EHPO output logs $\mathcal{L}$ by running this random strategy $\sigma$ in bounded time $t$. \textbf{Informally}, $\possible_t p$ means that an adversary could adopt a strategy $\sigma$ that is guaranteed to cause the desired outcome $p$ to be the case while taking time at most $t$ in expectation. In Section \ref{sec:prelim}, where $p$ is ``Non-adaptive methods outperform adaptive ones", Figure \ref{fig:wilson_deception}a shows $\possible_t p$. \textbf{Formally},
\begin{definition} \label{def:ehpologic}
Let $\boldsymbol{\sigma[\mathcal{L}]}$ denote the logs output from executing strategy $\sigma$ on logs $\mathcal{L}$, and let $\boldsymbol{\tau_\sigma(\mathcal{L})}$ denote the total time spent during execution. $\tau_\sigma(\mathcal{L})$ is equivalent to the sum of the times $T$ it took each HPO procedure $H \in \mathcal{H}$ executed in strategy $\sigma$ to run.
Note that both $\sigma[\mathcal{L}]$ and $\tau_\sigma(\mathcal{L})$ are random variables, as a function of the randomness of selecting $G$ and the actions sampled from $\sigma(\mathcal{L})$. 
For any formula $p$ and any $t \in \mathbb{R}_{>0}$, we say $\mathcal{L} \models \possible_t p$, i.e. ``$\mathcal{L}$ models that it is possible $p$ in time $t$,'' if 
\[
    \text{there exists a strategy } \sigma \in \Sigma, \text{ such that } \;\; \mathbb{P}(\sigma[\mathcal{L}] \models p) = 1 \; \text{ and } \; \mathbb{E}[\tau_\sigma(\mathcal{L})] \le t.
\]
\end{definition}

We will usually choose $t$ to be an upper bound on what is considered a reasonable amount of time to run EHPO. It does not make sense for $t$ to be unbounded, since this corresponds to the unrealistic setting of having infinite compute time to perform HPO runs. 
We model our budget in terms of time; however, we could use this setup to reason about other monotonically increasing resource costs, such as energy usage. 
Our indexed modal logic inherits many axioms of modal logic, with indexes added (Appendix), e.g.:
\begin{align*}
\vdash (p \rightarrow q) \rightarrow (\possible_t p \rightarrow \possible_t q) && \textit{(necess. + distribution)} && p \rightarrow \possible_t p && \textit{(reflexivity)} \\
\possible_t\possible_s p \rightarrow \possible_{t+s} p && \textit{(transitivity)} &&
\possible_s\necessary_t p \rightarrow \necessary_t p && \textit{(symmetry)}\\
\possible_t(p \land q) \rightarrow (\possible_t p \land \possible_t q) && \textit{(dist. over $\land$)},
\end{align*}
\textbf{To summarize}: The demon knows all possible hyper-HPs; it can pick whichever ones it wants to run EHPO within a bounded time budget $t$ to realize the outcome $p$ it wants. That is, if with some probability the demon can deceive us in some amount of time, then the demon can reliably deceive us with any larger time budget: If the demon fails to produce a deceptive result, it can use the strategy of just re-running until it yields the result it desires. Since $\possible_t$ models the worst-case all-powerful demon, it can also model any weaker EHPO user with time budget $t$.

\subsection{Expressing how we draw conclusions using $\mathcal{B}$} \label{sec:belieflogic}

We employ the modal operator $\mathcal{B}$ from the logic of belief\footnote{$\mathcal{B}$ is syntactically analogous to the $\necessary$ modal operator in standard modal logic~\cite{hintikka1962doxastic,vanBenthem2006epistemic, segerberg1999logicofbelief} (Appendix).} to model ourselves as an observer who believes in the truth of the conclusions drawn from running EHPO. $\mathcal{B}p$ reads ``It is concluded that $p$.'' For example, when comparing the performance of two algorithms for a task, $p$ could be ``$\mathcal{J}$ is better than $\mathcal{K}$" and thus $\mathcal{B}p$ would be understood as, ``It is concluded that $\mathcal{J}$ is better than $\mathcal{K}$.''   

We model ourselves as a consistent \emph{Type 1} reasoner~\cite{smullyan1986belief}. \textbf{Informally}, this means we believe all propositional tautologies (necessitation), our belief distributes over implication (distribution), and we do not derive contradictions (consistency). We do not require completeness: We allow the possibility of not concluding anything about $p$ (i.e., neither $\mathcal{B}p$ nor $\mathcal{B}\lnot p$). \textbf{Formally}, for any formulas $p$ and $q$,
\begin{align*}
    \vdash p \rightarrow & \mathcal{B} p & \textit{(necess.)}; &&
    \mathcal{B} (p \rightarrow q) \rightarrow & (\mathcal{B} p \rightarrow \mathcal{B} q) &  \textit{(dist.)}; && 
    &\lnot ( \mathcal{B}p \land \mathcal{B}\lnot p ) & \textit{(consistency)}.
\end{align*}

To understand our belief semantics, recall that EHPO includes a function $\mathcal{F}$, which maps a set of output logs $\mathcal{L}$ to our conclusions (i.e., $\mathcal{F}(\mathcal{L}) = \mathcal{P}$ is our set of conclusions). 
\textbf{Informally}, when our conclusion set $\mathcal{F}(\mathcal{L})$ contains a formula $p$, we say the set of logs $\mathcal{L}$ models our belief $\mathcal{B}$ in that formula $p$. In Section \ref{sec:prelim}, the logs of Figure \ref{fig:wilson_deception}a model $\mathcal{B}p$ and the logs of Figure \ref{fig:wilson_deception}b model $\mathcal{B}\lnot p$. \textbf{Formally},

\begin{definition}\label{def:conclusionfunction}  For any formula $p$, we say
$\mathcal{L} \models \mathcal{B} p$, ``$\mathcal{L}$ models our belief in $p$'', if $p \in \mathcal{F}(\mathcal{L})$.
\end{definition}

Note we constrain what $\mathcal{F}$ can output. For a reasonable notion of belief, $\mathcal{F}$ 
must model the consistent \emph{Type 1} reasoner axioms above. Otherwise, deception aside, $\mathcal{F}$ is an unreasonable way to draw conclusions, since it is not even compatible with our belief logic. 

\subsection{Expressing hyperparameter deception} \label{sec:deceptionlogic}

So far we have defined the semantics of our two separate modal operators, $\possible_t$ and $\mathcal{B}$. We now begin to reveal the benefit of using modal logic for our formalization. These operators can interact to formally express what we informally illustrated in Section \ref{sec:prelim}: a notion of hyperparameter deception. It is a well-known result that we can combine modal logics~\cite{scott1970multimodal} (Appendix).  We do so to define an axiom that, if satisfied, guarantees EHPO will not be able to deceive us. For any formula $p$,
\begin{align*}
    & \lnot \left( \possible_t \mathcal{B} p \; \land \; \possible_t \mathcal{B} \lnot p \right) & & \textit{($t$-non-deceptive)}.
\end{align*}
\textbf{Informally}, our running example can be considered a proof by exhibition: It violates this axiom because Figure \ref{fig:wilson_deception}a's logs model $\possible_t \mathcal{B} p$ and Figure \ref{fig:wilson_deception}b's logs model $\possible_t \mathcal{B} \lnot p$. That is, $\possible_t \mathcal{B} p \land \possible_t \mathcal{B} \lnot p$ using grid search for this task. 

For the worst-case, \emph{$t$-non-deceptiveness} expresses the following: \textbf{If there exists a strategy $\sigma$ by which the demon could get us to conclude $p$ in $t$ expected time, then there can exist no $t$-time strategy by which the demon could have gotten us to believe $\lnot p$}. 
To make this concrete, suppose our $t$-non-deceptive axiom holds for an EHPO method that results in $p$.  
Intuitively, given a maximum reasonable time budget $t$, if there is no adversary that can consistently control whether we believe $p$ 
or its negation when running that EHPO, then the EHPO is defended against deception. Conversely, if an adversary could consistently control our conclusions, then the EHPO is potentially gameable. That is, if our $t$-non-deceptive axiom does not hold (i.e., we can be deceived, $\possible_t \mathcal{B} p \land \possible_t \mathcal{B} \lnot p$), then even if we conclude $p$ after running EHPO, we cannot claim to \emph{know} $p$. Our belief as to the truth-value of $p$ could be under the complete control of an adversary---or just a result of happenstance.

\textbf{To summarize}: An EHPO is $t$-non-deceptive if it satisfies all of the axioms above. Our example in Section \ref{sec:prelim} is $t$-deceptive because the axioms do not hold. The semantics of these axioms capture all of the possible uncertainty from the process of drawing conclusions from EHPO--and how that uncertainty can combine to cause us to believe $t$-deceptive conclusions.

\section{Constructing Defended EHPO }\label{sec:defense}

Now that we have a formal notion of what it means for EHPO to be (non)-deceptive, 
\textbf{we can write proofs about what it means for an EHPO method to be guaranteed to be deception-free}. 
Importantly, these proofs will increase our confidence that our conclusions from EHPO are not due to the happenstance of picking a particular set of hyper-HPs. 

To talk about defenses, we need to understand what it means to construct a ``defended reasoner." In other words, for an EHPO $(\mathcal{H}, \mathcal{F})$, we need $\mathcal{F}$ to yield conclusions that we can defend against deception. 
Recall from Definition \ref{def:conclusionfunction} that logs $\mathcal{L}$ model our belief in a formula $p$, i.e. $\mathcal{L} \models \mathcal{B} p \equiv p \in \mathcal{F}(\mathcal{L})$. With this in mind, we begin by supposing we have a naive EHPO $(\mathcal{H}, \mathcal{F}_{\text{n}})$ featuring a naive reasoner $\mathcal{B}_{\text{n}}$ with corresponding belief function $\mathcal{F}_{\text{n}}$. We want to construct a new ``defended reasoner'' $\mathcal{B}_{*}$ that has a ``skeptical'' belief function $\mathcal{F_{*}}$.  $\mathcal{F_{*}}$ should weaken the conclusions of $\mathcal{F}_{\text{n}}$ (i.e., $\mathcal{F_{*}}(\mathcal{L}) \subseteq \mathcal{F}_{\text{n}}(\mathcal{L})$ for any $\mathcal{L}$) and result in an EHPO $(\mathcal{H}, \mathcal{F_*})$ that is guaranteed to be $t$-non-deceptive. In other words, defended reasoner $\mathcal{B}_*$ never concludes more than the naive reasoner $\mathcal{B}_{\text{n}}$. 
\textbf{Informally}, a straightforward way to do this is to have $\mathcal{B}_{*}$ conclude $p$ only if both the naive $\mathcal{B}_{\text{n}}$ would have concluded $p$, and it is impossible for an adversary to get $\mathcal{B}_{\text{n}}$ to conclude $\lnot p$ in time $t$. \textbf{Formally}, construct $\mathcal{B}_*$ such that for any $p$, $\mathcal{B}_{*} p \equiv \mathcal{B}_{\text{n}} p \land \lnot \possible_t \mathcal{B}_{\text{n}} \lnot p$ \; (\hypertarget{ax:defendedreasoner}{1}).

Directly from our axioms (Section \ref{sec:formalizing}), we can now prove $\mathcal{B}_{*}$ is defended. We will suppose it is possible for $\mathcal{B}_{*}$ to be deceived, demonstrate a contradiction, and thereby guarantee that $\mathcal{B}_{*}$ is $t$-non-deceptive. Suppose $\mathcal{B}_{*}$ can be deceived in time $t$, i.e. $\textcolor{red}{\possible_t \mathcal{B}_{*} p} \land \textcolor{red}{\possible_t \mathcal{B}_{*} \lnot p}$ is $\mathsf{True}$. Starting with the left, $\textcolor{red}{\possible_t \mathcal{B}_{*}p}$ :

\begingroup
\setlength{\tabcolsep}{3.5pt} 
\renewcommand{\arraystretch}{1} 
\begin{table}[H]
\vspace{-.2cm}
\small
\begin{center}
      \centering
        \begin{tabular}{r c l c c}
\toprule
\; & \; & \; & \; & \textbf{Rule} \\
\midrule
$\textcolor{red}{\possible_t \mathcal{B}_{*} p}$ & $\equiv$ & $\possible_t \left( \mathcal{B}_{\text{n}} p \land \lnot \possible_t \mathcal{B}_{\text{n}} \lnot p \right)$   &  \; & \text{Applying $\possible_t$ to the definition of $\mathcal{B}_{*} p$ \; (\hyperlink{ax:defendedreasoner}{1})} \\

\midrule
\; & $\rightarrow$ & $\possible_t \left( \lnot \possible_t \mathcal{B}_{\text{n}} \lnot p \right)$  & \;   &  \text{Reducing a conjunction to either of its terms: $(a \land b) \rightarrow b$} \\

\midrule
\; & $\rightarrow$ & $\textcolor{blue}{\lnot \possible_t \mathcal{B}_{\text{n}} \lnot p}$  & \;   & \text{Symmetry; dropping all but the right-most operator: $\possible_t(\possible_t a)\rightarrow \possible_t a$} \\
\bottomrule
\end{tabular}
\end{center}
\vspace{-.5cm}
\end{table}

We then pause to apply our axioms to the right side of the conjunction, $\textcolor{red}{\possible_t \mathcal{B}_{*}\lnot p}$ :

\begingroup
\setlength{\tabcolsep}{3pt} 
\renewcommand{\arraystretch}{1} 
\begin{table}[H]
\vspace{-.2cm}
\small
\begin{center}
      \centering
        \begin{tabular}{r c l c c}
\toprule
\; & \; & \; & \; & \textbf{Rule} \\
\midrule
$\textcolor{red}{\possible_t \mathcal{B}_{*} \lnot p}$ & $\equiv$ & $\possible_t \left( \mathcal{B}_{\text{n}} \lnot p \land \lnot \possible_t \mathcal{B}_{\text{n}} p \right)$   &  \; & \text{Applying $\possible_t$ to the definition of $\mathcal{B}_{*} \lnot p$ \; (\hyperlink{ax:defendedreasoner}{1})} \\

\midrule
\; & $\rightarrow$ & $\possible_t\mathcal{B}_{\text{n}} \lnot p \land \possible_t\lnot \possible_t \mathcal{B}_{\text{n}} p$  & \;   &  \text{Distributing $\possible_t$ over $\land$: $\possible_t (a \land b) \rightarrow (\possible_t a \land \possible_t b)$} \\

\midrule
\; & $\rightarrow$ & $\textcolor{blue}{\possible_t \mathcal{B}_{\text{n}} \lnot p}$  & \;   & \text{Reducing a conjunction to either of its terms: $(a \land b) \rightarrow a$} \\
\bottomrule
\end{tabular}
\end{center}
\vspace{-.35cm}
\end{table}

We now bring both sides of the conjunction back together: $\textcolor{red}{\possible_t \mathcal{B}_{*} p} \land  \textcolor{red}{\possible_t \mathcal{B}_{*} \lnot p}
\; \equiv \;
\textcolor{blue}{\lnot \possible_t \mathcal{B}_{\text{n}} \lnot p} \land  \textcolor{blue}{\possible_t \mathcal{B}_{\text{n}} \lnot p}$.
The \textcolor{blue}{right-hand side} is of the form $\lnot a \land a$, which must be $\mathsf{False}$. This contradicts our initial assumption that $\mathcal{B}_{*}$ is $t$-deceptive (i.e., $\textcolor{red}{\possible_t \mathcal{B}_{*} p} \land \textcolor{red}{\possible_t \mathcal{B}_{*} \lnot p}$ is $\mathsf{True}$). Therefore, $\mathcal{B}_{*}$ is $t$-non-deceptive. 

This example illustrates the power of our choice of formalization. In just a few lines of simple logic, we can validate defenses against deception. \textbf{This analysis shows that a $t$-defended reasoner $\mathcal{B}_{*}$ \emph{is always possible}}, and it does so without needing to refer to the particular underlying semantics of an EHPO. However, we intend this example to only be illustrative, as it may not be practical to compute $\mathcal{B}_{*}$ as defined in (\hyperlink{ax:defendedreasoner}{1}) if we cannot easily evaluate whether $\possible_t \mathcal{B}_{\text{n}} \lnot p$. We next suggest a concrete EHPO with a defended $\mathcal{B_{*}}$, and show how deception can be avoided in our Section \ref{sec:prelim} example by using this EHPO instead of grid search. 

\textbf{A defended random search EHPO.} Random search takes two hyper-HPs, a distribution $\mu$ over the HP space and a number of trials $K \in \mathbb{N}$ to run. HPO consists of $K$ independent trials of training algorithms $\mathcal{A}_{\lambda_1}, \mathcal{A}_{\lambda_2}, \ldots, \mathcal{A}_{\lambda_K}$, where the HPs $\lambda_k$ are independently drawn from $\mu$, taking expected time proportional to $K$. When drawing conclusions, we usually look at the ``best'' run for each algorithm. For simplicity, we suppose there is only one algorithm, $\mathcal{A}$. We bound how much the choice of hyper-HPs can affect the HPs, and define a defended EHPO based on a variant of random search.

\begin{definition}\label{def:defended_random}
Suppose that we are given a naive EHPO procedure $(\{H\}, \mathcal{F}_{\text{n}})$, in which $H$ is random search and is the only HPO in our EHPO, and $\mathcal{F}_{\text{n}}$ is a ``naive'' belief function associated with a naive reasoner $\mathcal{B}_{\text{n}}$. 
For any $K, R \in \mathbb{N}$, we define the ``$(K,R)$-defended'' belief function $\mathcal{F}_*$ for a skeptical reasoner $\mathcal{B}_{*}$ as the following conclusion-drawing procedure.
First, $\mathcal{F}_*$ only makes conclusion set $\mathcal{P}_*$ from a single log $\hat \ell$ with $K*R$ trials; otherwise, it concludes nothing, outputting $\emptyset$.
Second, $\mathcal{F}_{*}$ splits the single $\hat \ell$ into $R$ logs $\ell_1, \ell_2, \ldots, \ell_R$, each containing $K$ independent-random-search trials.\footnote{This is not generally allowable. $\mathcal{F}_*$ can do this because random-search logs contain interchangeable trials.}
Finally, $\mathcal{F}_{*}$ outputs the intersection of what the naive reasoner would have output on each log $\ell_i$,
\[
    \mathcal{F}_{*}(\{\hat \ell\}) = \mathcal{P}_* \equiv \mathcal{F}_{\text{n}}(\{ \ell_1 \})
    \cap \mathcal{F}_{\text{n}}(\{ \ell_2 \}) \cap \cdots \cap \mathcal{F}_{\text{n}}(\{ \ell_R \}).
\]
Equivalently, $\{\hat \ell\} \models \mathcal{B_{*}}p$ only if $\{\ell_i\} \models \mathcal{B}_{\text{n}}p$ for all $i$.
\end{definition}


\textbf{Informally}, to draw a conclusion using this EHPO, $\mathcal{B}_{*}$ splits a random-search-trial log of size $K*R$ into $R$ groups of $K$-trial logs, passing each $K$-trial log to one of an ensemble of $R$ naive reasoners $\mathcal{B}_{\text{n}}$. $\mathcal{B}_{*}$ only concludes $p$ if all $R$ naive reasoners unanimously agree on $p$. We can guarantee this EHPO to be $t$-non-deceptive by assuming a bound on how much the hyper-HPs can affect the HPs.

\begin{theorem} \label{thm:defendedhpo}
Suppose that the set of allowable hyper-HPs $\mathcal{C}$ of $H$ is constrained, such that any two allowable random-search distributions $\mu$ and $\nu$ have Renyi-$\infty$-divergence at most a constant, i.e. $D_{\infty}(\mu \| \nu) \le \gamma$. The $(K,R)$-defended random-search EHPO of Definition \ref{def:defended_random} is guaranteed to be $t$-non-deceptive if we set $R \ge \sqrt{t \exp(\gamma K)/K} = O(\sqrt{t})$.
\end{theorem}

We prove Theorem \ref{thm:defendedhpo} in the Appendix. This result shows that our defense is \emph{actually} a defense, and moreover it defends with a log size $K*R$---and compute requirement for good-faith EHPO---that scales sublinearly in $t$. A good-faith actor can, in sublinear-in-$t$ time, produce a log (of length $K*R$) that will allow our $t$-non-deceptive reasoner to reach conclusions. This means that we defend against adversaries with much larger compute budgets than are expected from good-faith actors.
 
\textbf{Validating our defense empirically and selecting hyper-HPs.} Any defense ultimately depends on the hyper-HPs it uses. Thus, we should have a reasonable belief that choosing differently would not have led an opposite conclusion. We therefore run a two-phased search~\cite{choi2019empirical, henderson2017reinforcement, Riquelme2018-cn}, repeating our VGG16-CIFAR10 experiment from Section \ref{sec:prelim}. First, we run a coarse-grained, dynamic protocol to find reasonable hyper-HPs for Adam's $\epsilon$; second, we use those hyper-HPs to run our defended random search. We start with a distribution to search over $\epsilon$, and note that the performance is best on the high end. We  change the hyper-HPs, shifting the distribution until Adam's performance starts to degrade, and use the resulting hyper-HPs ($\epsilon \in [10^{10}, 10^{12}]$) to run our defense (Appendix). 

We now run a modified version of our defended EHPO in Definition \ref{def:defended_random}, described in Algorithm \ref{algo:def}, with $K*R=600$ ($200$ logs for each optimizer). Using a budget of $M=10000$ iterations, we subsample $\kappa=11$ logs and pass them to an ensemble of $\kappa$ naive reasoners $\mathcal{B}_{\text{n}}$. We use $\kappa$ logs, relaxing the requirement of using all $K*R$ logs in Definition \ref{def:defended_random}, for efficiency. Each iteration $m$ concludes the majority conclusion of the $\kappa$-sized $\mathcal{B}_{\text{n}}$ ensemble. This is why we set $\kappa$ to an odd number---to avoid ties. $\mathcal{B}_{*}$ draws conclusions based on the results of the $M$-majority conclusions. That is, we further relax the requirements of Definition \ref{def:defended_random}: Instead of requiring unanimity, 
$\mathcal{B}_{*}$ only requires agreement on the truth-value of $p$ for a fractional subset of $M$. We set this fraction using parameter $\delta \in [0,1]$, where $\delta$ controls how skeptical our defended reasoner $\mathcal{B}_{*}$ is (lower $\delta$ corresponding to more skepticism). $\mathcal{B}_{*}$ concludes $p$ when at least $(1 - \delta)$ of our $M$ subsampled runs concluded $p$. When this threshold is not met, $\mathcal{B}_{*}$ remains skeptical and concludes nothing. We summarize our final results in Table \ref{table:defense}, and provide complete results in the Appendix. Given how similar the optimizers all perform on this task (similar to Figure \ref{fig:wilson_deception}), being more skeptical increases the likelihood that we do not conclude anything.

\begin{minipage}{.53\linewidth}
\vspace{-.2cm}
      \centering
      \begin{algorithm}[H]
	\caption{Defense with Random Search}\label{algo:def}
	\begin{algorithmic}[1]
		\REQUIRE Set of ${K*R}$ random-search logs $\{\mathcal{L}_i\}_{i=1}^{KR}$, \\ defense subsampling budget $M$, \\ criterion constant $\delta$, subsample size $\kappa$.
		\FOR{$m=1,\cdots, M$}
			\STATE Subsample $\kappa$ logs: $\{\mathcal{L}_i\}_{i=1}^\kappa\sim\{\mathcal{L}_i\}_{i=1}^{KR}$.
			\STATE Obtain conclusions $\{\mathcal{P}_i\}_{i=1}^\kappa$ from $\{\mathcal{L}_i\}_{i=1}^\kappa$.
			\STATE Obtain output conclusion for $m$: \\
			\;\;\;\;$\mathcal{P}^{(m)}\leftarrow \text{Majority}(\{\mathcal{P}_i\}_{i=1}^\kappa)$
		\ENDFOR
		\IF{$\exists p$ s.t. $\geq (1-\delta)M$ of $\{\mathcal{P}^{(m)}\}_{i=1}^M$ conclude $p$}
		    \STATE Conclude $p$.
		\ELSE
		    \STATE Conclude nothing.
		\ENDIF
	\end{algorithmic}
\end{algorithm}

\end{minipage}%
\hspace{.4cm}
\begin{minipage}{.42\linewidth}
\vspace{-.2cm}
      \centering
      \begin{table}[H]
\caption{Results from repeating our Section \ref{sec:prelim} experiment, using Algorithm \ref{algo:def} instead of grid search. $p$ = ``Non-adaptive optimizers (SGD and HB) perform better than the adaptive optimizer Adam''.}
\label{table:defense}
\small
\begin{center}
\begin{tabular}{c c c  c  ccc}
\toprule
\; &   $p$ &   $\lnot p$   &   $1 - \delta$  &   Conclude    \\
\midrule
\multirow{3}[4]{*}{\shortstack{SGD\; \\vs.\;\\Adam\;}}   &   \multirow{3}[4]{*}{$0.213$}  &   \multirow{3}[4]{*}{$0.788$\;}  &   0.75    &   $\neg p$    \\
\cmidrule{4-5}
   &   &  & 0.8 &   Nothing \\
\cmidrule{4-5}
   &   &  & 0.9 &   Nothing \\
\midrule
\multirow{3}[4]{*}{\shortstack{HB\; \\vs.\;\\Adam\;}}   &   \multirow{3}[4]{*}{$0.168$}  &   \multirow{3}[4]{*}{$0.832$\;}  &   0.75    &   $\lnot p$   \\
\cmidrule{4-5}
   &   &  & 0.8 &   $\lnot p$ \\
\cmidrule{4-5}
   &   &  & 0.9 &   Nothing \\
\bottomrule
\end{tabular}
\end{center}
\end{table}
\end{minipage} 

\section{Conclusion and Practical Takeaways}\label{sec:conclusion}
Much recent empirical work illustrates that it is easy to draw inconsistent conclusions from HPO~\cite{choi2019empirical, sivaprasad2019optimizer, melis2018evaluation, musgrave2020metric, bouthillier19reproducible, schneider2019deepobs, dodge2019nlp, Lucic2018-dr}. We call this problem \emph{hyperparameter deception} and, to derive a defense, \textbf{argue that the process of drawing conclusions using HPO should itself be an object of study}. Taking inspiration from \citeauthor{descartes1996evildemon}' demon, we formalize a logic for studying an epistemic HPO procedure. 
The demon can run any number of reproducible HPO passes to try to get us to believe a particular notion about algorithm performance. 
Our formalization enables us to not believe deceptive notions: It naturally suggests how to guarantee that an EHPO is defended against deception. We offer recommendations to avoid hyperparameter deception in practice (we expand on this in the Appendix):
\begin{itemize}[topsep=0pt, leftmargin=.25cm]
    \item \textbf{Researchers should construct their own notion of skepticism $\mathcal{B_*}$, appropriate to their specific task.} There is no one-size-fits-all defense solution. Our results are \emph{\textbf{broad insights}} about defended EHPO: A defended EHPO is \emph{\textbf{always possible}}, but finding an efficient one will depend on the task.
    \item \textbf{Researchers should make explicit how they choose hyper-HPs.} 
    What is reasonable is ultimately a function of what the ML community accepts. Being explicit, rather than eliding hyper-HP choices, is essential for helping decide what is reasonable. As a heuristic, we recommend setting hyper-HPs such that they include HPs for which the optimizers' performance starts to degrade, as we do above. 
    \item \textbf{Avoiding hyperparameter deception is just as important as reproducibility}. We have shown that reproducibility \cite{henderson2017reinforcement, pineau2019checklist, gundersen2018reproducibility, bouthillier19reproducible, Sinha2020-aw} is only part of the story for ensuring reliability. While necessary for guarding against brittle findings, it is not sufficient. We can replicate results---even statistically significant ones---that suggest conclusions that are altogether wrong.
\end{itemize}

More generally, our work is a call to researchers to reason more rigorously about their beliefs concerning algorithm performance. In relation to EHPO, this is akin to challenging researchers to reify their notion of $\mathcal{B}$---to justify their belief in their conclusions from the HPO. 
Such epistemic rigor concerning drawing conclusions from empirical studies has a long history in more mature branches of science and computing, including evolutionary biology~\cite{gould1996mismeasure}, statistics~\cite{gelman2014garden, gelman2019fishing}, programming languages~\cite{mytkowicz2009pl}, and computer systems~\cite{friedman1996bias} (Appendix). We believe that applying similar rigor will contribute significantly to the ongoing effort of making ML more robust and reliable. 

\section*{Acknowledgements}
A. Feder Cooper is supported by the Artificial Intelligence Policy and Practice initiative at Cornell University, the Digital Life Initiative at Cornell Tech, and the John D. and Catherine T. MacArthur Foundation. Jessica Zosa Forde is supported by ONR PERISCOPE MURI N00014-17-1- 2699. We would additionally like to thank the following individuals for feedback on earlier ideas, drafts, and iterations of this work: Prof. Rediet Abebe, Harry Auster, Prof. Solon Barocas, Jerry Chee, Prof. Jonathan Frankle, Dr. Jack Goetz, Prof. Hoda Heidari, Kweku Kwegyir-Aggrey, Prof. Karen Levy, Prof. Helen Nissenbaum, and Prof. Gili Vidan. We also thank Meghan Witherow for her whimsical interpretation of \citeauthor{descartes1996evildemon}' evil demon, which we include in the Appendix.

\bibliography{references}
\bibliographystyle{plainnat}

\newpage
\appendix
\onecolumn
\setcounter{theorem}{0}
\setcounter{definition}{0}

\section*{Checklist}

\begin{enumerate}

\item For all authors...
    \begin{enumerate}
      \item Do the main claims made in the abstract and introduction accurately reflect the paper's contributions and scope?
        \answerYes{}
      \item Did you describe the limitations of your work?
        \answerYes{See Section~\ref{sec:defense}.}
      \item Did you discuss any potential negative societal impacts of your work?
        \answerYes{See Section~\ref{sec:conclusion}. We also expand further in the Appendix.}
      \item Have you read the ethics review guidelines and ensured that your paper conforms to them?
        \answerYes{}
    \end{enumerate}
    
    \item If you are including theoretical results...
    \begin{enumerate}
      \item Did you state the full set of assumptions of all theoretical results?
        \answerYes{See Sections \ref{sec:ehpo}, \ref{sec:formalizing} \& \ref{sec:defense}. For extended theoretical results and assumptions, see the Appendix.}
    	\item Did you include complete proofs of all theoretical results?
        \answerYes{See Section \ref{sec:defense} and the Appendix.}
    \end{enumerate}
    
    \item If you ran experiments...
    \begin{enumerate}
      \item Did you include the code, data, and instructions needed to reproduce the main experimental results (either in the supplemental material or as a URL)?
        \answerYes{We provide detailed information concerning the code and data in the Appendix. We release the code at \url{https://github.com/pasta41/deception} and provide further instructions in the README.}
      \item Did you specify all the training details (e.g., data splits, hyperparameters, how they were chosen)?
        \answerYes{See the Appendix.}
    	\item Did you report error bars (e.g., with respect to the random seed after running experiments multiple times)?
        \answerYes{See Section \ref{sec:prelim} and the Appendix.}
    	\item Did you include the total amount of compute and the type of resources used (e.g., type of GPUs, internal cluster, or cloud provider)?
        \answerYes{See Appendix.}
    \end{enumerate}
    
    \item If you are using existing assets (e.g., code, data, models) or curating/releasing new assets...
    \begin{enumerate}
      \item If your work uses existing assets, did you cite the creators?
        \answerYes{See Section \ref{sec:prelim} and the Appendix.}
      \item Did you mention the license of the assets?
        \answerYes{See Appendix.}
      \item Did you include any new assets either in the supplemental material or as a URL?
        \answerYes{We provide detailed information concerning the code and data in the Appendix. We release the code at \url{https://github.com/pasta41/deception} and provide further instructions in the README.}
      \item Did you discuss whether and how consent was obtained from people whose data you're using/curating?
        \answerNA{We used benchmark datasets.}
      \item Did you discuss whether the data you are using/curating contains personally identifiable information or offensive content?
        \answerNA{}
    \end{enumerate}
    
    \item If you used crowdsourcing or conducted research with human subjects...
    \begin{enumerate}
      \item Did you include the full text of instructions given to participants and screenshots, if applicable?
        \answerNA{}
      \item Did you describe any potential participant risks, with links to Institutional Review Board (IRB) approvals, if applicable?
        \answerNA{}
      \item Did you include the estimated hourly wage paid to participants and the total amount spent on participant compensation?
        \answerNA{}
    \end{enumerate}
\end{enumerate}

\newpage
\section{Glossary} \label{app:sec:glossary}
We provide a glossary of terms, definitions, and symbols that we introduce throughout the paper, streamlined in one place.

\subsection{Definitions Reference}

\emph{hyper-hyperparameters} (hyper-HPs) are HPO-procedure-input values, such as the spacing between different points in the grid for grid search and the distributions to sample from in random search. 

\vspace{.25cm}

\begin{definition} A log $\ell$ records all the choices and measurements made during an HPO run, including the total time $T$ it took to run. It has all necessary information to make the HPO run reproducible.
\end{definition}

\vspace{.25cm}

\begin{definition}
	An HPO procedure $H$ is a tuple $(H_{*}, \mathcal{C}, \Lambda, \mathcal{A}, \mathcal{M}, G, X)$ where $H_{*}$ is a randomized algorithm, $\mathcal{C}$ is a set of allowable hyper-HPs (i.e., allowable configurations for $H_{*}$), $\Lambda$ is a set of allowable HPs (i.e., of HP sets $\lambda$), $\mathcal{A}$ is a training algorithm (e.g. SGD), $\mathcal{M}$ is a model (e.g. VGG16), $G$ is a PRNG, and $X$ is some dataset (usually split into train and validation sets). When run, $H_{*}$ takes as input a hyper-HP configuration $c \in \mathcal{C}$ and a random seed $r \in \mathcal{I}$, then proceeds to run $\mathcal{A}_{\lambda}$ (on $\mathcal{M}_{\lambda}$ using $G(r)$ and data\footnote{Definition \ref{def:hpo} does not preclude cross-validation, as this can be part of $H_{*}$. The input dataset $X$ can be split in various ways, as a function of the random seed $r$.} from $X$) some number of times for different HPs $\lambda \in \Lambda$. Finally, $H_{*}$ outputs a tuple $(\lambda^*, \ell)$, where $\lambda^*$ is the HP configuration chosen by HPO and $\ell$ is the log documenting the run.
\end{definition}

\vspace{.25cm}

\begin{definition}
An \textbf{epistemic hyperparameter optimization procedure (EHPO)} is a tuple $(\mathcal{H}, \mathcal{F})$ where $\mathcal{H}$ is a set of HPO procedures $H$ (Definition \ref{def:hpo}) and $\mathcal{F}$ is a function that maps a set of HPO logs $\mathcal{L}$ (Definition \ref{def:log}) to a set of logical formulas $\mathcal{P}$, i.e. $\mathcal{F}(\mathcal{L}) = \mathcal{P}$. An execution of EHPO involves running each $H \in \mathcal{H}$ some number of times (each run produces a log $\ell$) and then evaluating $\mathcal{F}$ on the set of logs $\mathcal{L}$ produced in order to output the conclusions $\mathcal{F}(\mathcal{L})$ we draw from all of the HPO runs.
\end{definition}

\begin{definition}
A randomized \textbf{strategy} $\boldsymbol{\sigma}$ is a function that specifies which action the demon will take. Given $\mathcal{L}$, its current set of logs, $\boldsymbol{\sigma(\mathcal{L})}$ gives a distribution over concrete actions, where each action is either 1) running a new $H$ with its choice of hyper-HPs $c$ and seed $r$ 2) erasing some logs, or 3) returning. We let $\Sigma$ denote the set of all such strategies.
\end{definition}

\vspace{.25cm}

\begin{definition}
Let $\boldsymbol{\sigma[\mathcal{L}]}$ denote the logs output from executing strategy $\sigma$ on logs $\mathcal{L}$, and let $\boldsymbol{\tau_\sigma(\mathcal{L})}$ denote the total time spent during execution. $\tau_\sigma(\mathcal{L})$ is equivalent to the sum of the times $T$ it took each HPO procedure $H \in \mathcal{H}$ executed in strategy $\sigma$ to run.
Note that both $\sigma[\mathcal{L}]$ and $\tau_\sigma(\mathcal{L})$ are random variables, as a function of the randomness of selecting $G$ and the actions sampled from $\sigma(\mathcal{L})$. For any formula $p$ and any $t \in \mathbb{R}_{>0}$, we say $\mathcal{L} \models \possible_t p$, i.e. ``$\mathcal{L}$ models that it is possible $p$ in time $t$,'' if 
\[
    \text{there exists a strategy } \sigma \in \Sigma, \text{ such that } \;\; \mathbb{P}(\sigma[\mathcal{L}] \models p) = 1 \; \text{ and } \; \mathbb{E}[\tau_\sigma(\mathcal{L})] \le t.
\]
\end{definition}

\vspace{.25cm}

\begin{definition}
For any formula $p$, we say $\mathcal{L} \models \mathcal{B} p$, ``$\mathcal{L}$ models our belief in $p$'', if $p \in \mathcal{F}(\mathcal{L})$.
\end{definition}

\vspace{.25cm}

\begin{definition}
Suppose that we are given a naive EHPO procedure $(\{H\}, \mathcal{F}_{\text{n}})$, in which $H$ is random search and is the only HPO in our EHPO, and $\mathcal{F}_{\text{n}}$ is a ``naive'' belief function associated with a naive reasoner $\mathcal{B}_{\text{n}}$. 
For any $K, R \in \mathbb{N}$, we define the ``$(K,R)$-defended'' belief function $\mathcal{F}_*$ for a skeptical reasoner $\mathcal{B}_{*}$ as the following conclusion-drawing procedure.
First, $\mathcal{F}_*$ only makes conclusion set $\mathcal{P}_*$ from a single log $\hat \ell$ with $K*R$ trials; otherwise, it concludes nothing, outputting $\emptyset$.
Second, $\mathcal{F}_{*}$ splits the single $\hat \ell$ into $R$ logs $\ell_1, \ell_2, \ldots, \ell_R$, each containing $K$ independent-random-search trials.\footnote{This is not generally allowable. $\mathcal{F}_*$ can do this because random-search logs contain interchangeable trials.}
Finally, $\mathcal{F}_{*}$ outputs the intersection of what the naive reasoner would have output on each log $\ell_i$,
\[
    \mathcal{F}_{*}(\{\hat \ell\}) = \mathcal{P}_* \equiv \mathcal{F}_{\text{n}}(\{ \ell_1 \})
    \cap \mathcal{F}_{\text{n}}(\{ \ell_2 \}) \cap \cdots \cap \mathcal{F}_{\text{n}}(\{ \ell_R \}).
\]
Equivalently, $\{\hat \ell\} \models \mathcal{B_{*}}p$ only if $\{\ell_i\} \models \mathcal{B}_{\text{n}}p$ for all $i$.
\end{definition}

\newpage
\subsection{Symbols and Acronyms Reference}

\begingroup
\setlength{\tabcolsep}{8pt} 
\renewcommand{\arraystretch}{1.1} 
\begin{table}[H]
\small
\begin{center}
      \centering
        \begin{tabular}{p{0.1\linewidth}p{0.5\linewidth}p{0.3\linewidth}}
\toprule
\textbf{Term} & \textbf{Explanation} & \textbf{Example} \\
\midrule
HPO & Acronym for hyperparameter optimization & \; \\
$\mathcal{J}$, $\mathcal{K}$ & Used as examples of arbitrary optimizers & \\
$P$ & Arbitrary atomic proposition & \\
$p$, $q$, $\phi$ & Used as arbitrary or (when specified) specific logical formulas & $p$ = ``Non-adaptive optimizers have higher test accuracy than adaptive optimizers." \\
HP(s) & Acronym for hyperparameter(s) & \\
$\ell \in \mathcal{L}$,  & Log (Definition \ref{def:log}); log set & Figure \ref{tab:sgd-hb}; Log for running HPO using SGD \\
$T$ & The total time it took to run HPO to produce a log $\ell$  & \; \\
$\mathcal{I}$ & A set of integers & Typically the 64-bit integers \\
$r$ & Random seed; $r \in \mathcal{I}$ & \\
$G$ & Pseudo-random number generator; $G(r)$; $G: \mathcal{I} \rightarrow \mathcal{I}^{\infty}$ & \\
PRNG & Acronym for pseudo-random number generator; $G$ & \\
$H$ & HPO procedure (Definition \ref{def:hpo}) & SGD, VGG-16 grid search experiment \\
$H_*$ & A randomized algorithm used in $H$ & Random search \\
$c \in \mathcal{C}$ & Hyper-HP configuration; of set of allowable such configurations for $H_{*}$ & powers-of-2 grid spacing; configurations the demon has access to\\
$\lambda \in \Lambda$; $\lambda^{*}$ & HP config. used to run an HPO pass; of allowable HP configs., determined by $c$; $\lambda^*$ is the output HP config. that performs the best & $\alpha=1$ in Wilson; allowable $\alpha$ values, e.g. $[.001, .01, 1]$  \\
$\mathcal{A}$; $\mathcal{A}_{\lambda}$ & Training algorithm; parameterized by HPs $\lambda$ & SGD; SGD with $\alpha=1$\\
$\mathcal{M}$; $\mathcal{M}_{\lambda}$ & Model; parameterized by HPs $\lambda$ & VGG16 \\
$X$ & A dataset & CIFAR-10 \\
$\alpha$ & Learning rate & Figure \ref{fig:full-our}, $\alpha = 1$ \\
$\epsilon$ & Adam-specific HP & Figure \ref{fig:full-our}, we set $\epsilon = 10^{12}$ \\
EHPO & Epistemic HPO (Definition \ref{def:ehpo}) & Our defended random search in Section \ref{sec:defense}\\
$\mathcal{H}$ & Set of HPO procedures $H$  &  \\
$\mathcal{P}$ & Set of concluded logical formulas; $p \in \mathcal{P}$ & \\
$\mathcal{F}$ & A function that maps a set of HPO logs $\mathcal{L}$ to a set of logical formulas $\mathcal{P}$  & $\mathcal{F}_*$ (skeptical belief function); $\mathcal{F}_{\text{n}}$ (naive belief function) \\
$\necessary$ & Modal logic operator for ``necessary"  &  $\necessary p$ reads ``It is necessary that $p$\\
$\possible$ & Modal logic operator for ``possible"  &  $\possible p$ reads ``It is possible that $p$\\
$\vdash$ & Indicates a theorem of propositional logic  &  $\vdash Q \rightarrow \necessary Q$ (necessitation) \\
$\possible_t$ & EHPO modal operator (Section \ref{sec:ehpologic}; Definition \ref{def:ehpologic})  &  \\
$\mathcal{B}$ & Belief modal operator  &  $\mathcal{B}_*$ (skeptical belief); $\mathcal{B}_{\text{n}}$ (naive belief)\\
$\sigma \in \Sigma$ & A randomized strategy function that specifies EHPO actions; set of all such strategies (Section \ref{sec:ehpologic}, Definition \ref{def:strategy}) &  \\
$\sigma(\mathcal{L})$ & Distribution over concrete actions for log set &  \\
$\sigma[\mathcal{L}]$ & The logs output from running $\sigma$ on $\mathcal{L}$ &  \\
$\tau_{\sigma}(\mathcal{L})$ & Total time spent executing $\sigma[\mathcal{L}]$ &  \\
$\models$ & Denotes ``models" &  $\mathcal{L}\models\possible_t p$: $\mathcal{L}$ model that $p$ is possible in  $t$ \\
$\gamma$ & Renyi-$\infty$-divergence constant upper bound (Theorem \ref{thm:defendedhpo}) & \\
$K$, $R$ & Numbers of independent random search trials (Section \ref{sec:defense}) & \\
$\kappa$ & Subsampling size (Algorithm \ref{algo:def}) & We set $\kappa=11$ (Section \ref{sec:defense}) \\
$M$ & Subsampling budget (Algorithm \ref{algo:def}) & We set $M=10000$ (Section \ref{sec:defense}) \\
$\delta$ & Skeptical reasoner conclusion threshold (Algorithm \ref{algo:def}) & See Table \ref{table:defense} \\
\bottomrule
\end{tabular}
\end{center}
\end{table}


\section{Section \ref{sec:prelim} Appendix: Additional Notes on the Preliminaries} \label{app:sec:prelim}

The code for running these experiments can be found at \url{https://github.com/pasta41/deception}.

\subsection{Empirical Deception Illustration using \citet{wilson2017marginal}}

\subsubsection{Why we chose \citet{wilson2017marginal}}

We elaborate on why we specifically chose \citet{wilson2017marginal} as our running example of hyperparameter deception. There are four main reasons why we thought this was the right example to focus on for an illustration. First, the experiment involves optimizers known across ML (e.g. SGD, Adam), a model frequently used for benchmark tasks (VGG16) and a commonly-used benchmark dataset (CIFAR-10). Unlike other examples of hyperparameter deception, one does not need highly-specialized domain knowledge to understand the issue \cite{dodge2019nlp, Lucic2018-dr}. Second, the paper is exceptionally well-cited and known in the literature, so many folks in the community are familiar with its results. Third, we were certain that we could demonstrate hyperparameter deception before we ran our experiments; we observe that Adam's update rule basically simulates Heavy Ball when its $\epsilon$ parameter is set high enough. So, we were confident that we could (at the very least) get Adam to perform as well as Heavy Ball via changing hyper-HPs, which would demonstrate hyperparameter deception. We then found further support for this observation in concurrent work \cite{sivaprasad2019optimizer}, which cited earlier work \cite{choi2019empirical} that also observes this. Fourth, the claim in \citet{wilson2017marginal} is fairly broad. They make a claim about adaptive vs. non-adaptive optimizers, more generally. If the claim had been narrower -- about small $\epsilon$ values for numerical stability, then perhaps hyperparameter deception would not have occurred. In general, we note that narrower claims could help avoid deception. 

\subsection{Expanded empirical results}

We elaborate on the results we present in Section \ref{sec:prelim}.

\subsubsection{Experimental setup}\label{sec:wilson:setup}

We replicate and run a variant of \citet{wilson2017marginal}'s VGG16 experiment on CIFAR-10, using SGD, Heavy Ball, and Adam as the optimizers. 

We launch each run on a local machine configured with a 4-core 2.6GHz Inter (R) Xeon(R) CPU, 8GB memory and an NIVIDIA GTX 2080Ti GPU. 

Following the exact configuration from \citet{wilson2017marginal}, we set the mini-batch size to be 128, the momentum constant to be 0.9 and the weight decay to be 0.0005. 

The learning rate is scheduled to follow a linear rule: The learning rate is decayed by a factor of 10 every 25 epochs. The total number of epochs is set to be 250.

For the CIFAR-10 dataset, we apply random horizontal flipping and normalization. Note that \citet{wilson2017marginal} does not apply random cropping on CIFAR-10; thus we omit this step to be consistent with their approach. We adopt the standard cross entropy loss.

For each HPO setting, we run 5 times and average the results and include error bars two standard deviations above and below the mean. 

\subsubsection{Associated results and logs}\label{app:sec:logs}

In line with our notion of a \emph{log} (Definition \ref{def:log}), we provide data tables (Figures \ref{tab:sgd-hb}, \ref{tab:wilson-adam-1}, and \ref{tab:wilson-adam-2}) that correspond with our results graphed in the Figures \ref{fig:full-our}, \ref{fig:full-wilson}, \ref{fig:full-our-Adam}.

\newpage
\begin{figure}[H]
\centering
\includegraphics[scale=.35]{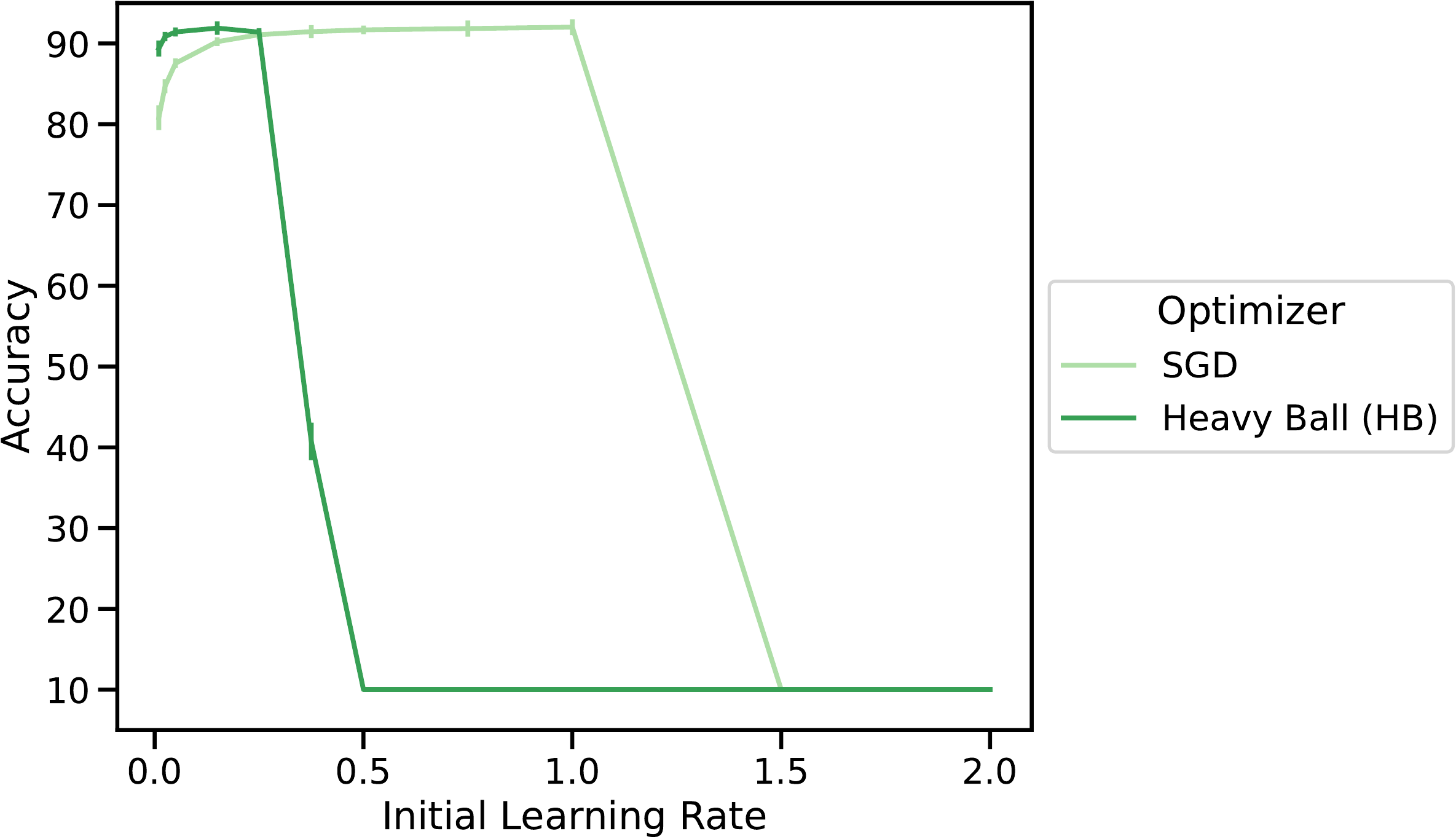}
\caption{Full test accuracy results of VGG-16 on CIFAR-10 for SGD and Heavy Ball learning rate ($\alpha$) HPO. Error bars indicate two standard deviations above and below the mean. Each HPO setting is measured with five replicates. We achieve similar performance as \citet{wilson2017marginal}.}
\label{fig:full-our}
\end{figure}

\begin{figure}[H]

\includegraphics[scale=.29]{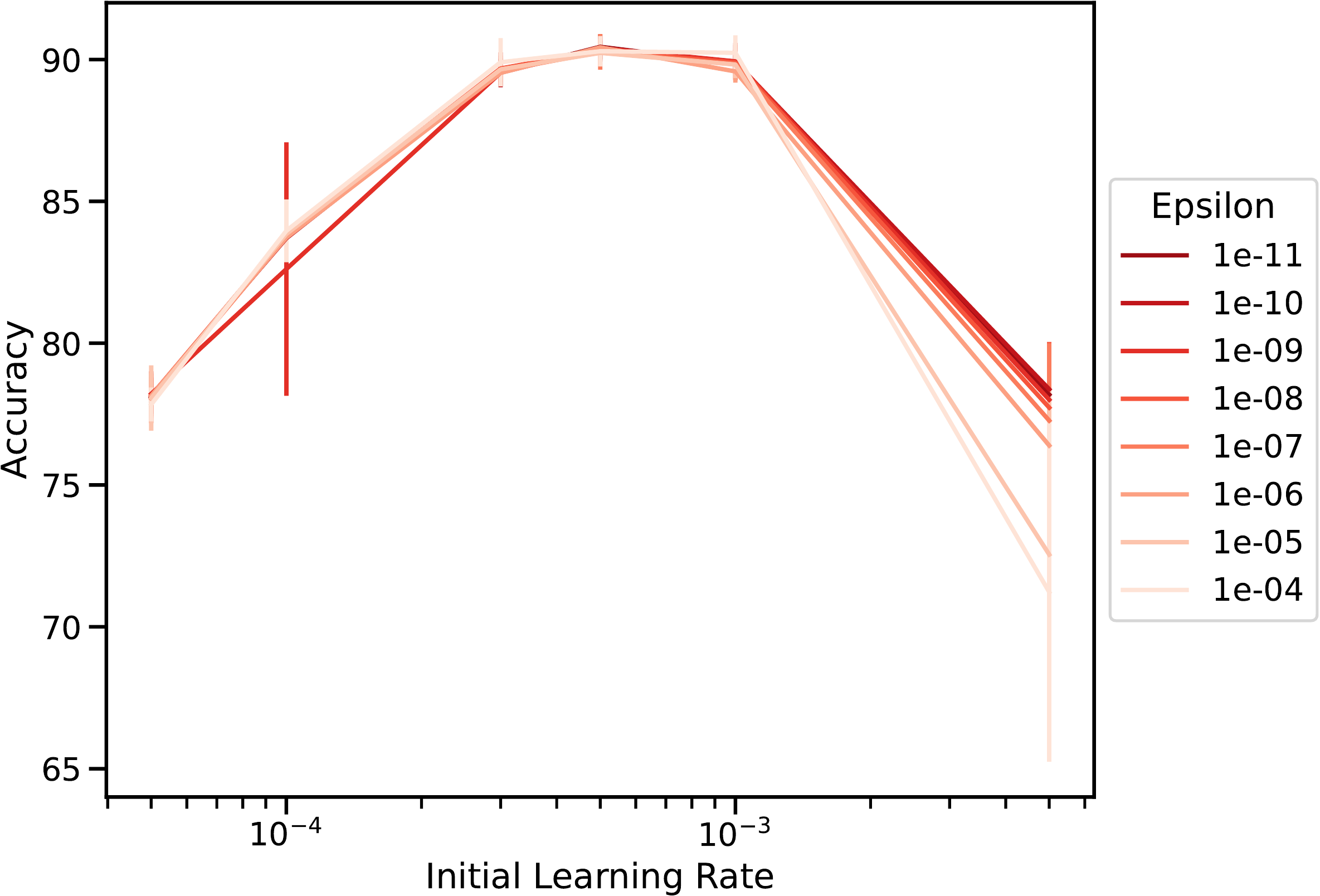}
\includegraphics[scale=.29]{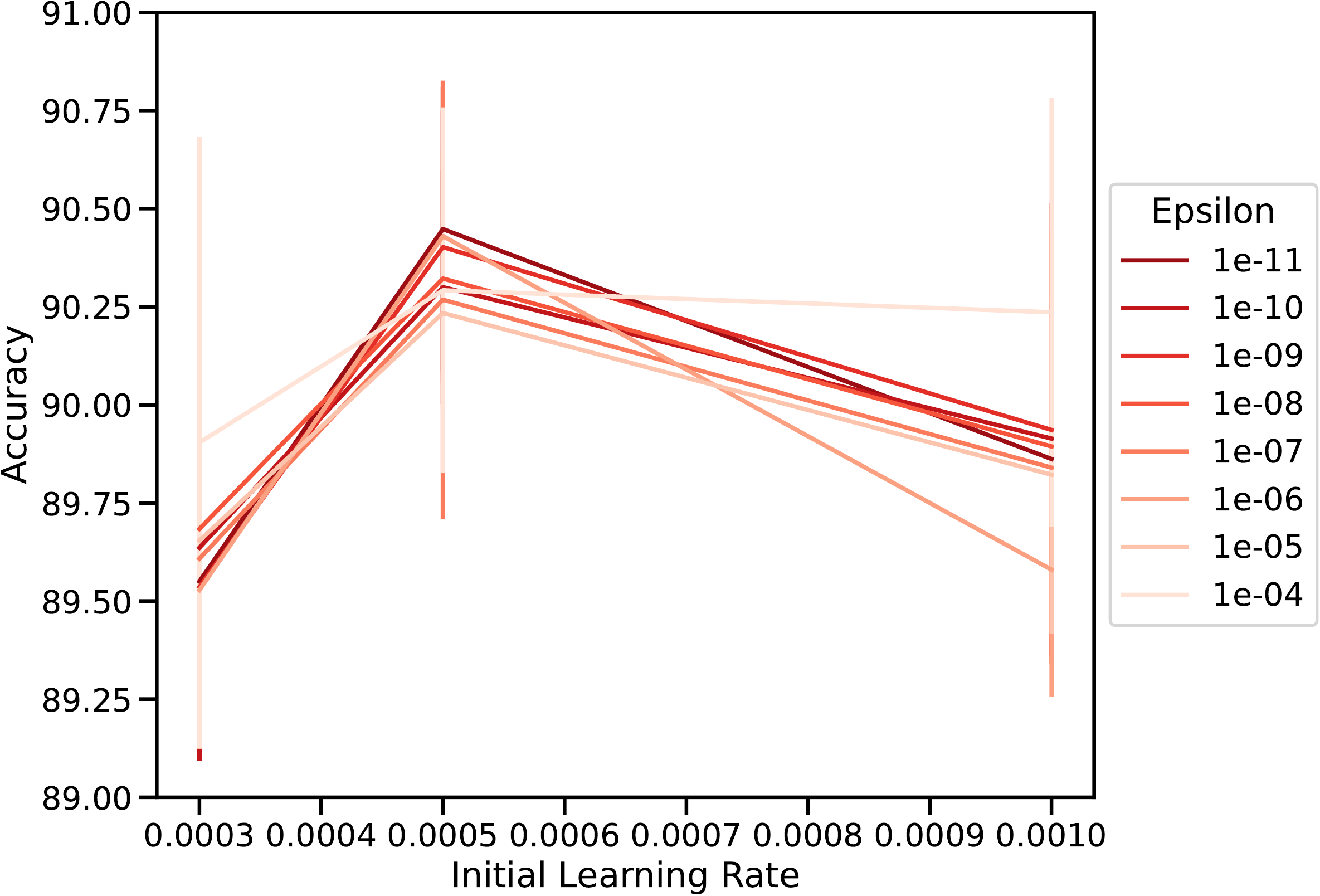}
\caption{Tuning over learning rate for different small values of $\epsilon$. On the left, we show a wide range of learning rates tested. On the right, we zoom in on the portion of results where the best test accuracy occurs. These results reflect what \citet{wilson2017marginal} showed, but with tuning over $\epsilon$ (small values).  Each HP setting is used to train VGG-16 on CIFAR-10 five times, and the error bars represent two standard deviations above and below the mean test accuracy.}
\label{fig:full-wilson}
\end{figure}

\begin{figure}[H]
\centering
\includegraphics[scale=.35]{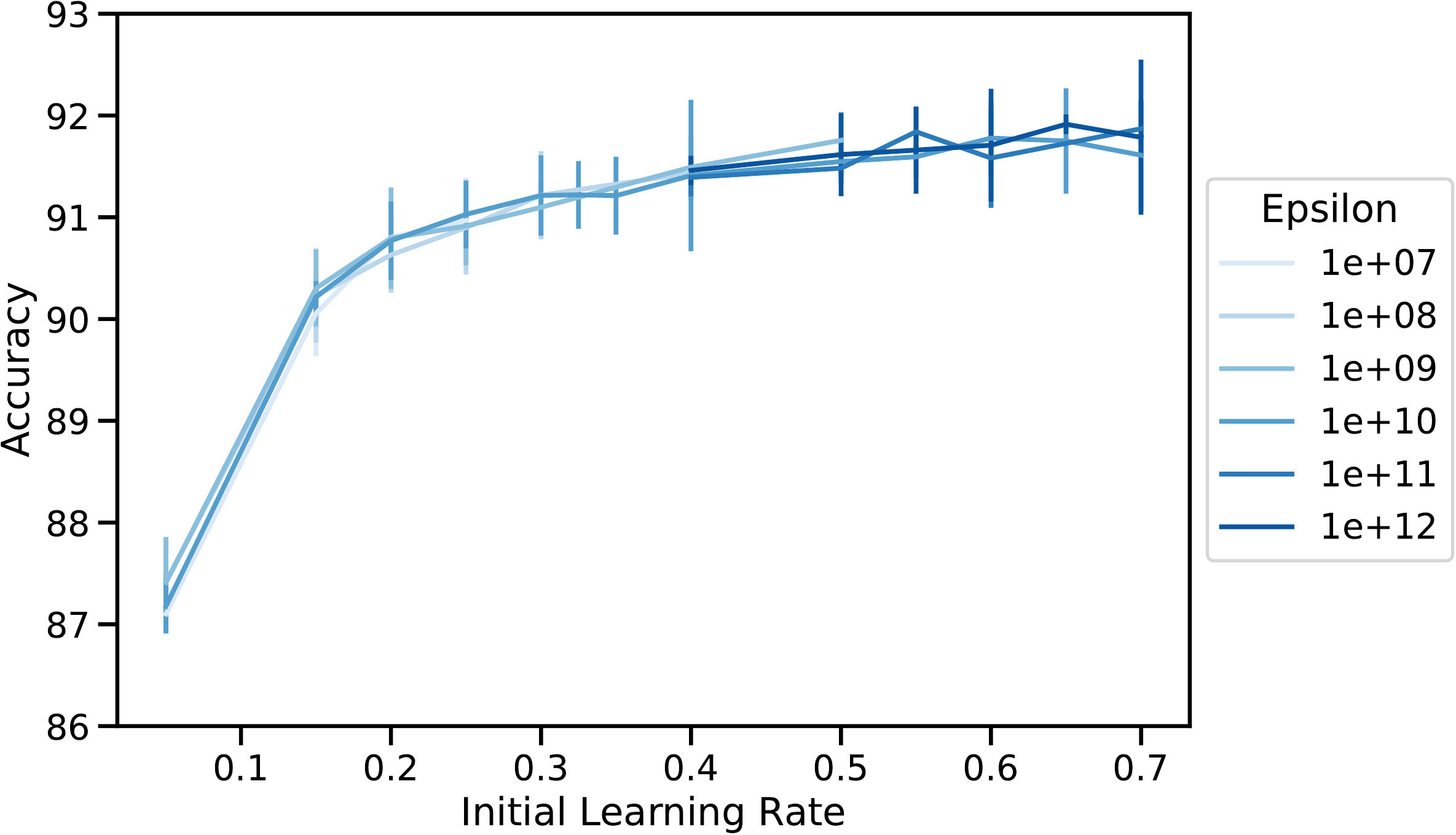}
\caption{Results for our expanded search over large $\epsilon$ values for Adam. We show test accuracy on CIFAR-10 as a function of different learning rates $\alpha$ for the different large $\epsilon$ values. Error bars show two standard deviations above and below mean test accuracy for five replicates for each HP setting.}
\label{fig:full-our-Adam}
\end{figure}

\begin{figure}[H]
\includegraphics[scale=.5]{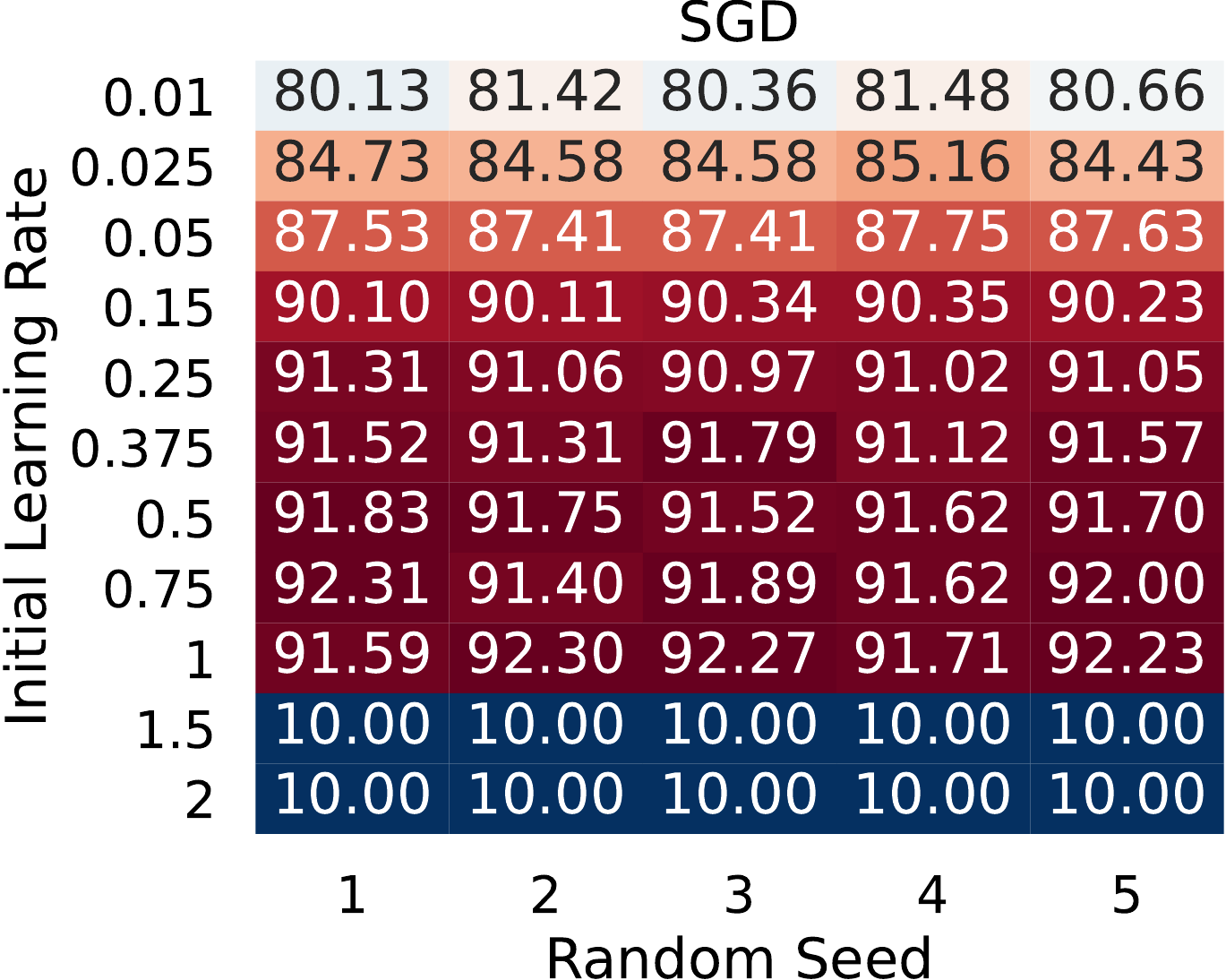}
\includegraphics[scale=.5]{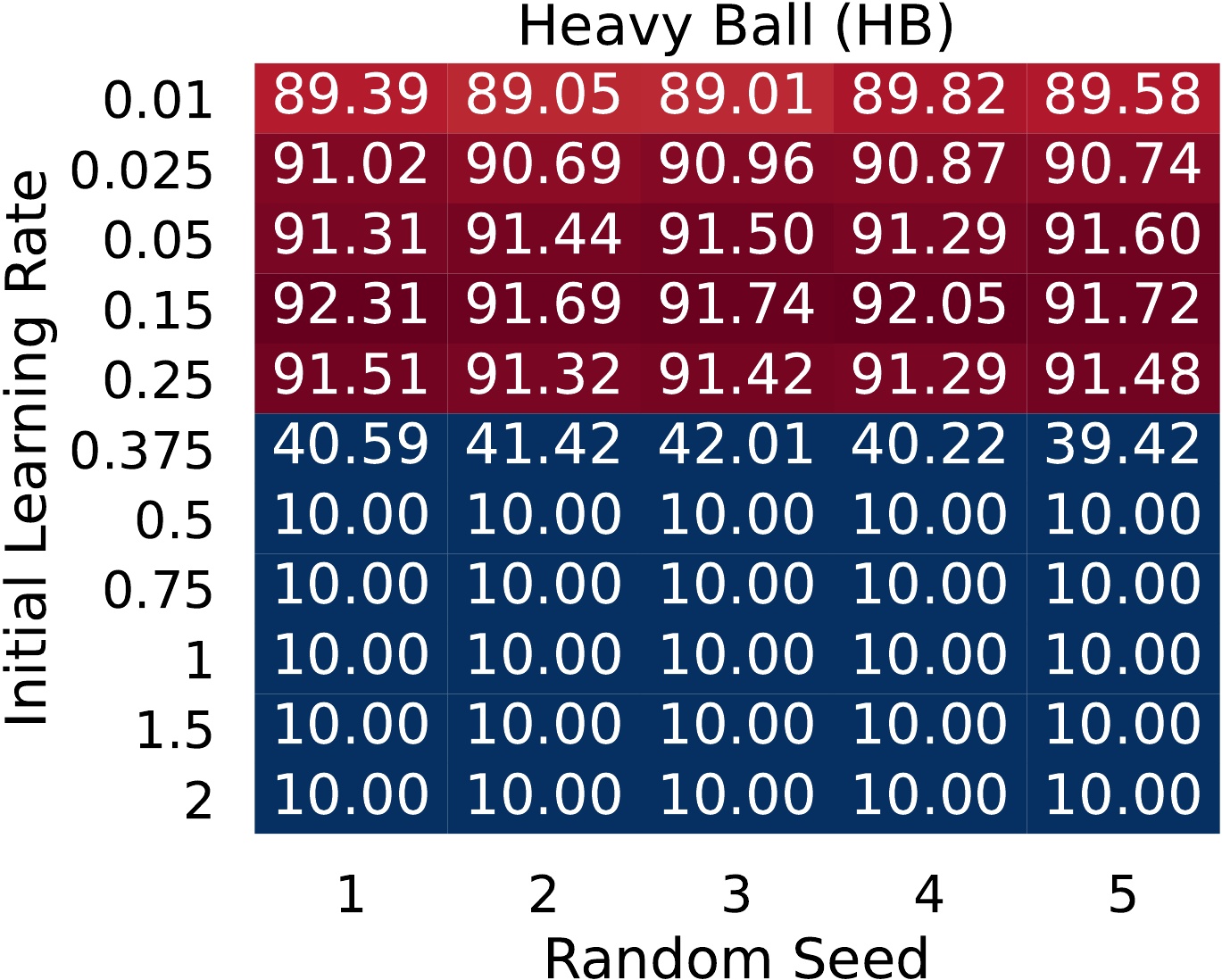}
\vspace{-.35cm}
\caption{Heatmap logs of test accuracy of VGG-16 on CIFAR-10 for SGD and Heavy Ball for each initial learning rate and random seed. These logs correspond to the results graphed in Figure \ref{fig:full-our}.}
\vspace{.35cm}
\label{tab:sgd-hb}


\includegraphics[scale=.5]{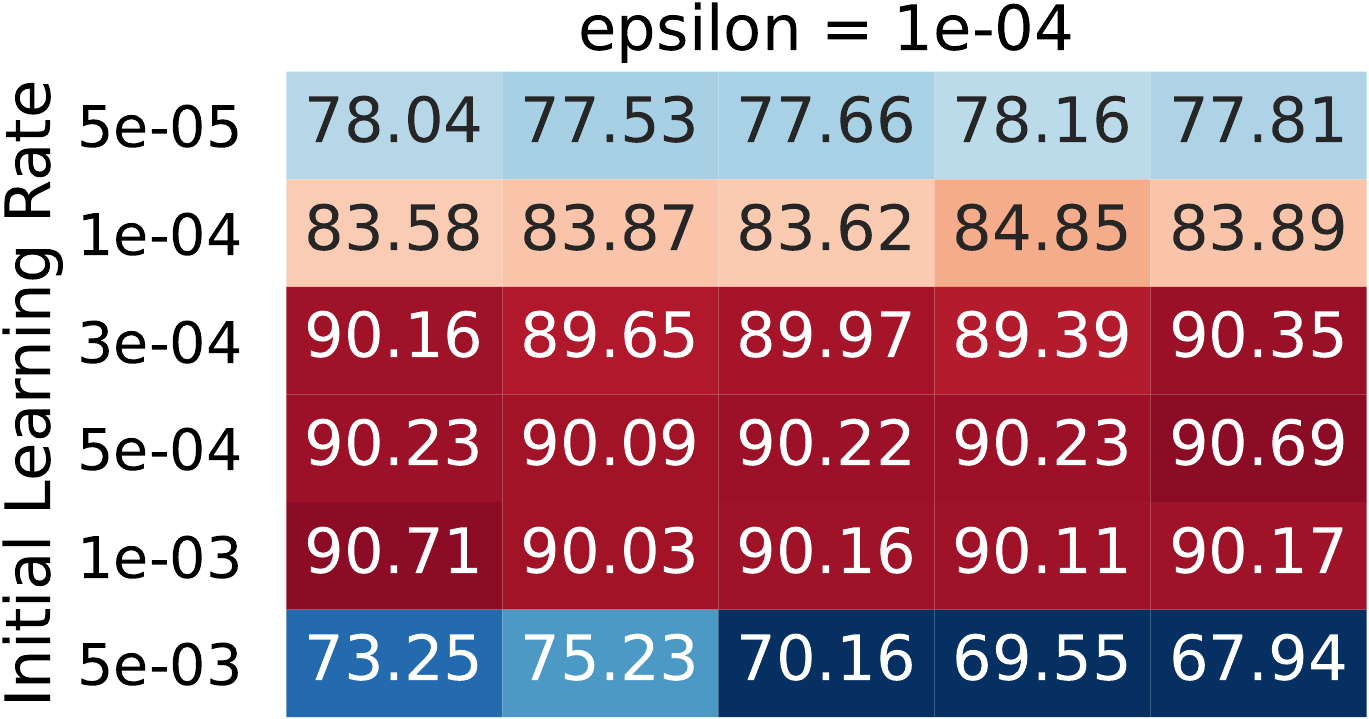}
\includegraphics[scale=.5]{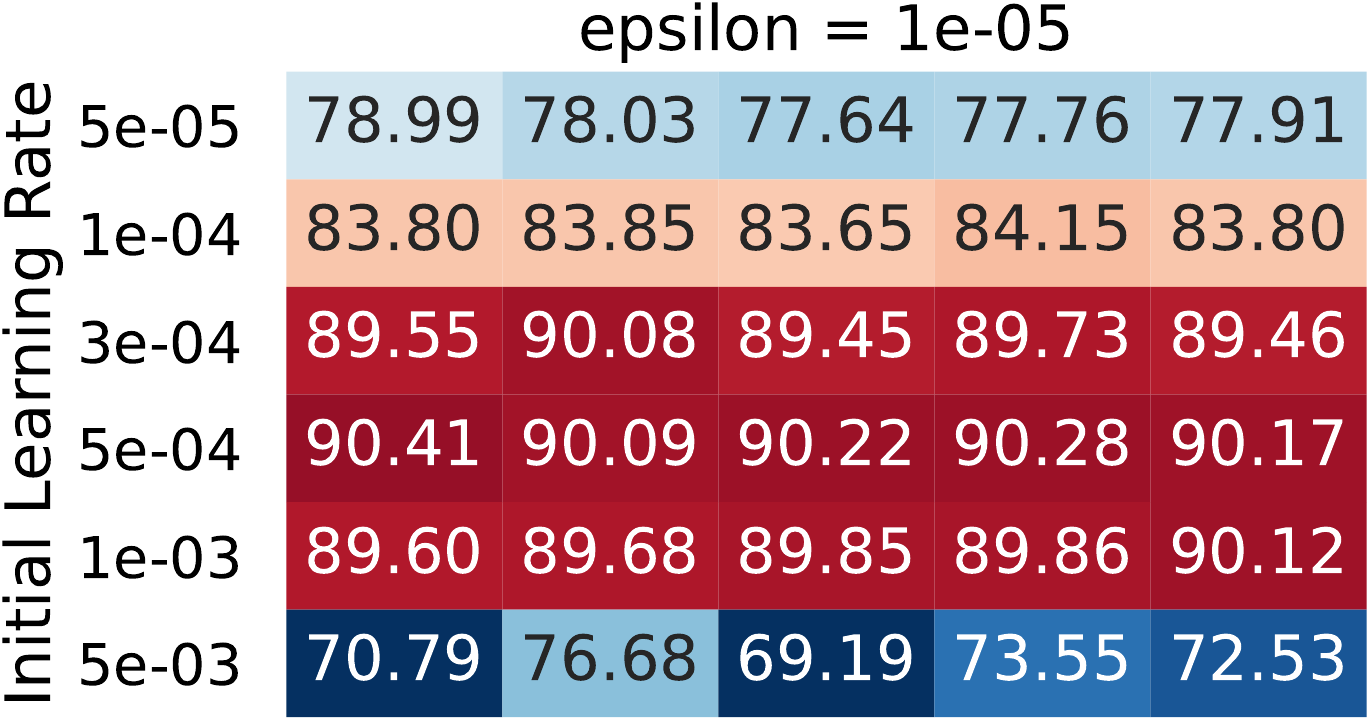}

\includegraphics[scale=.5]{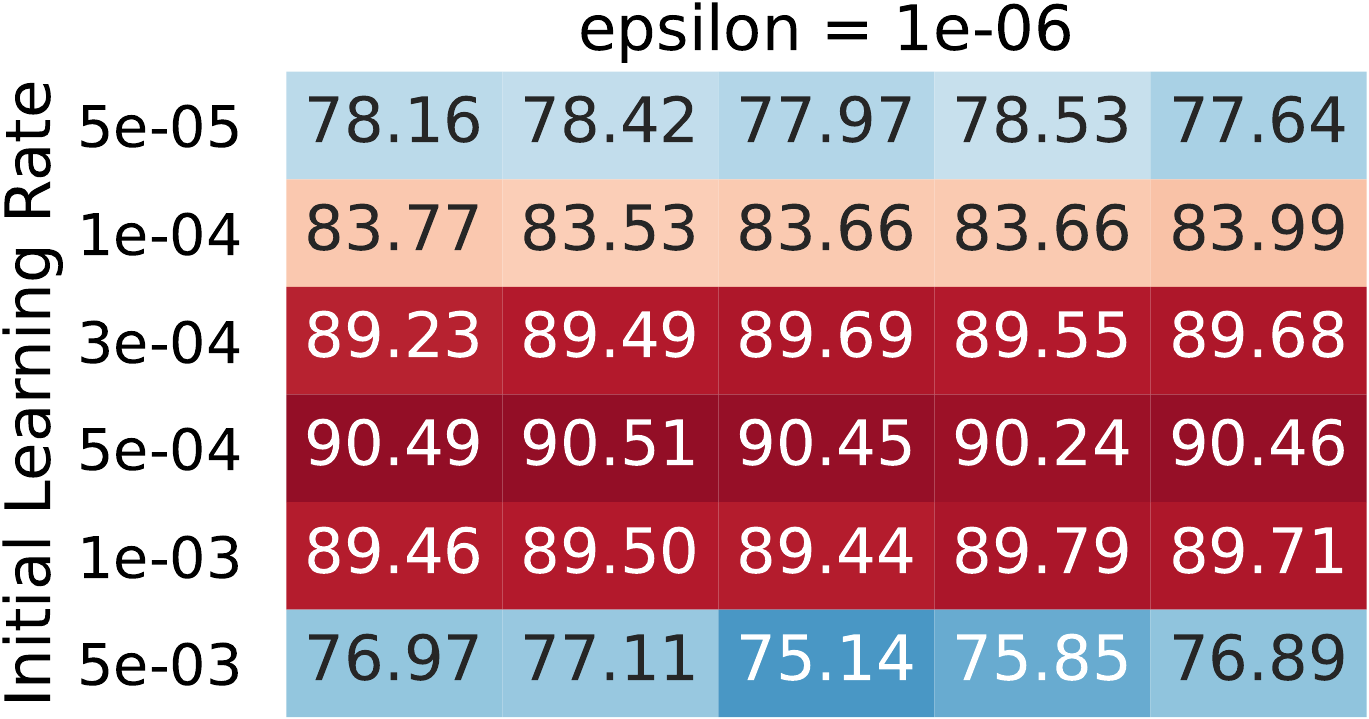}
\includegraphics[scale=.5]{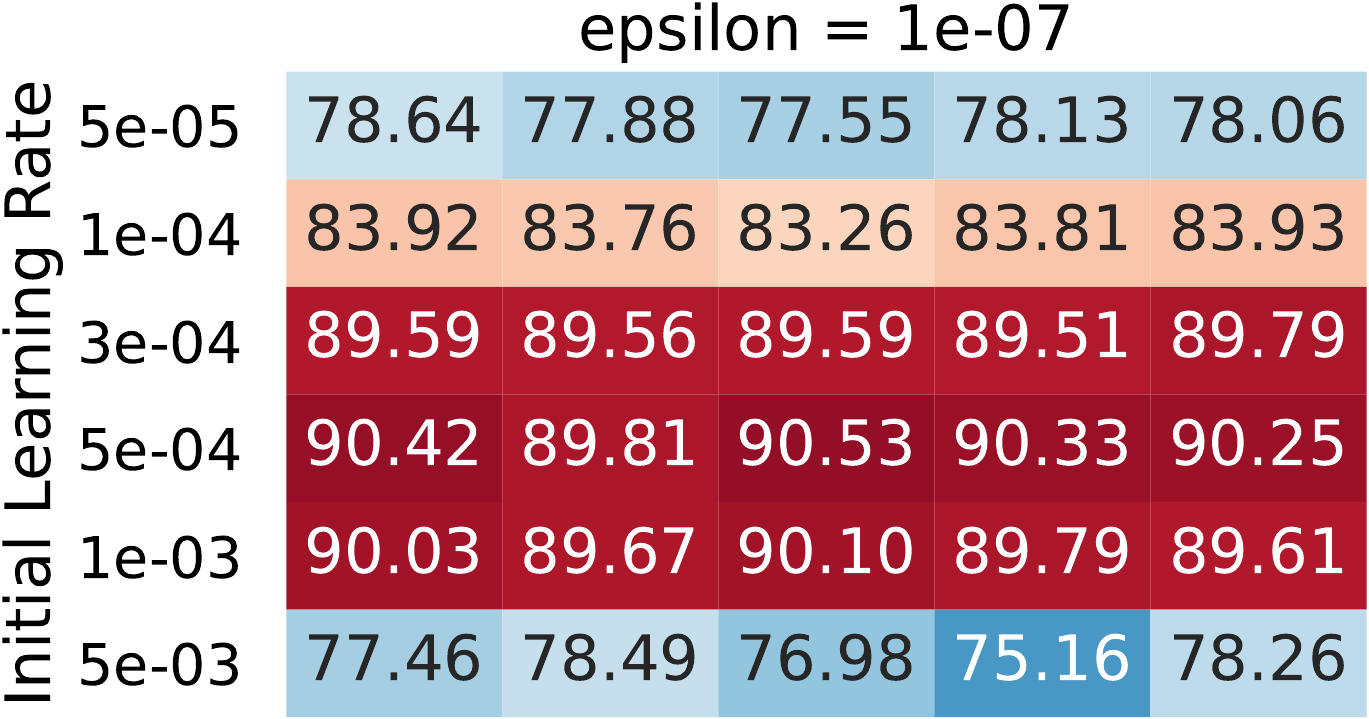}

\includegraphics[scale=.5]{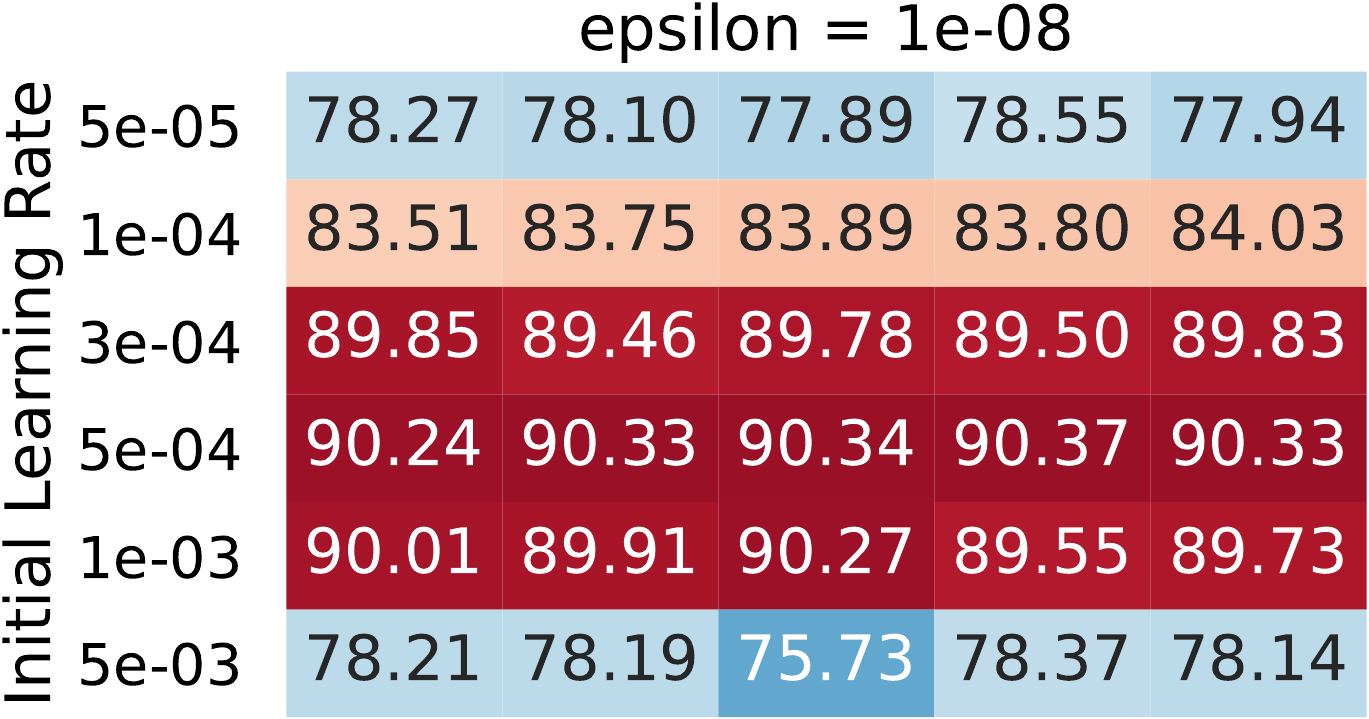}
\includegraphics[scale=.5]{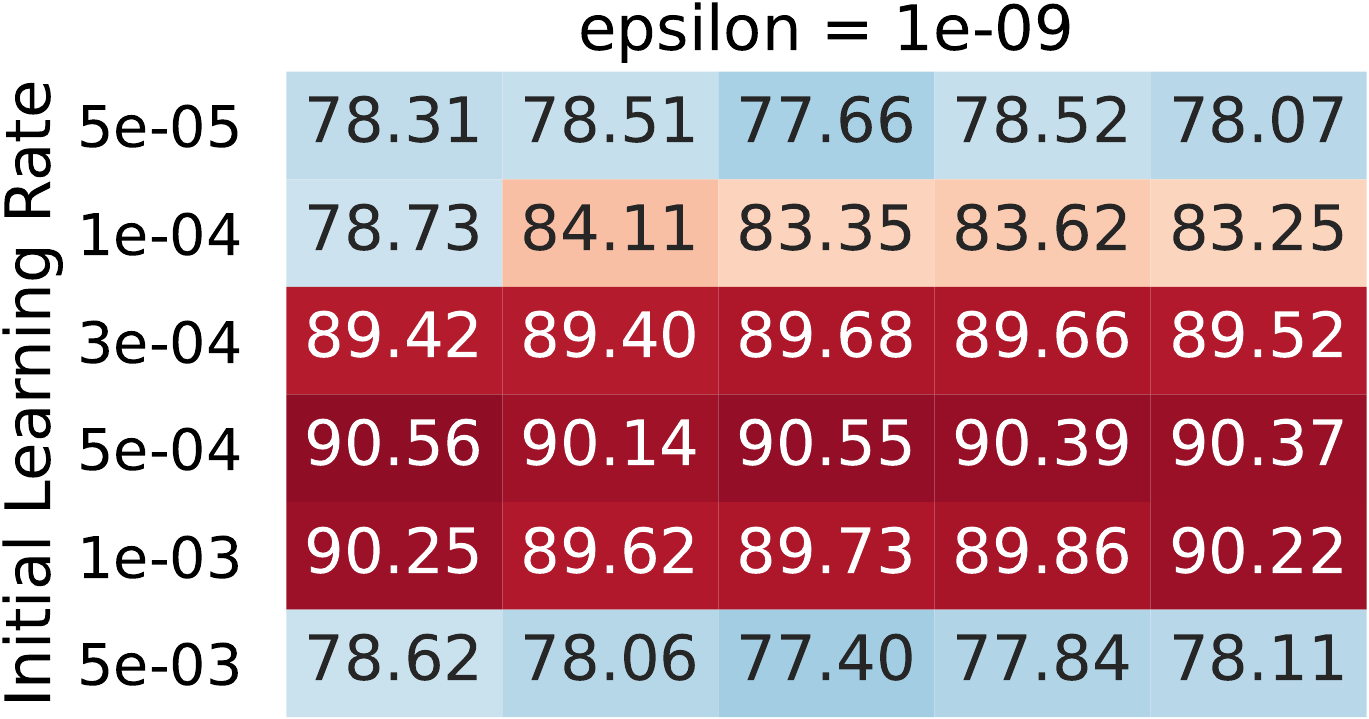}

\includegraphics[scale=.5]{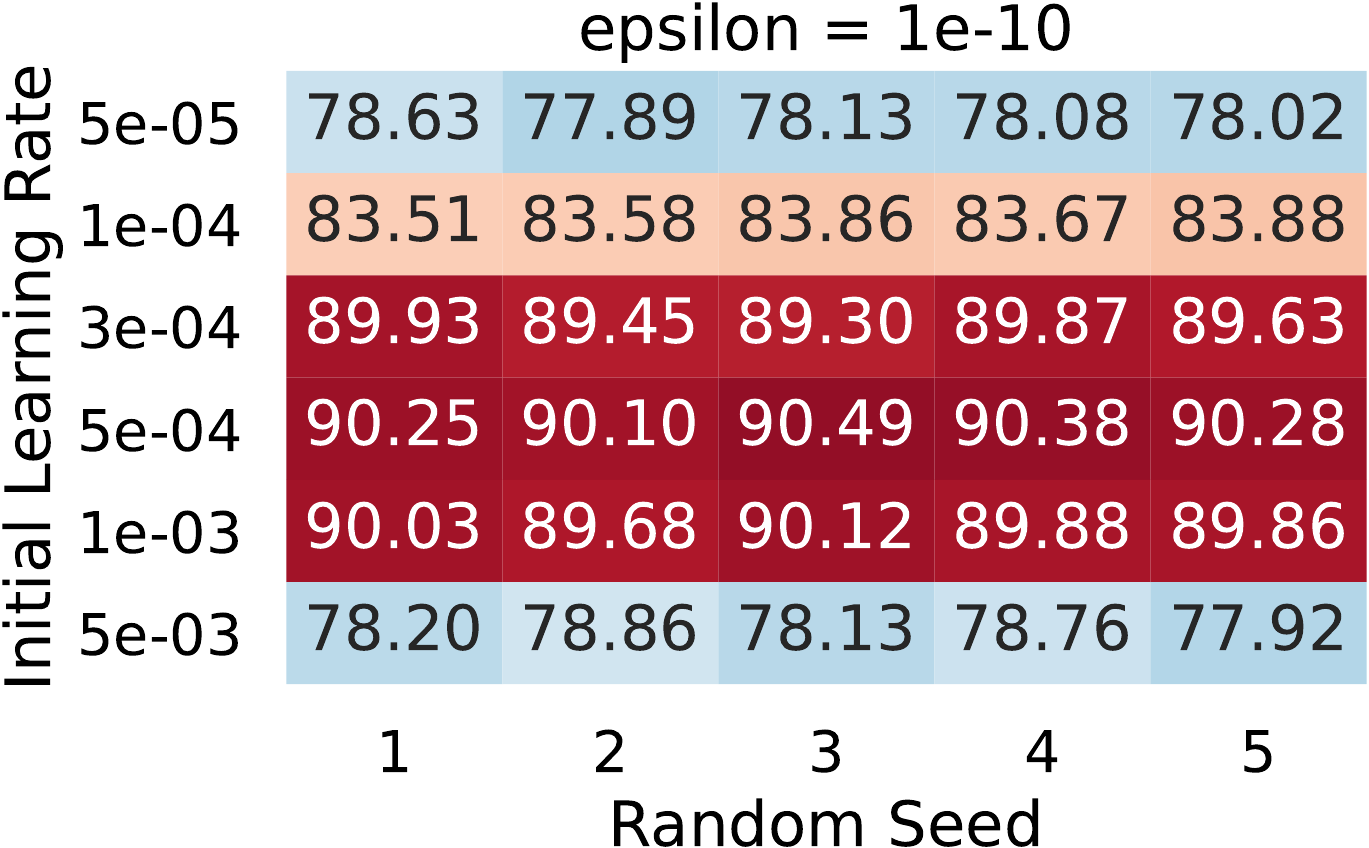}
\includegraphics[scale=.5]{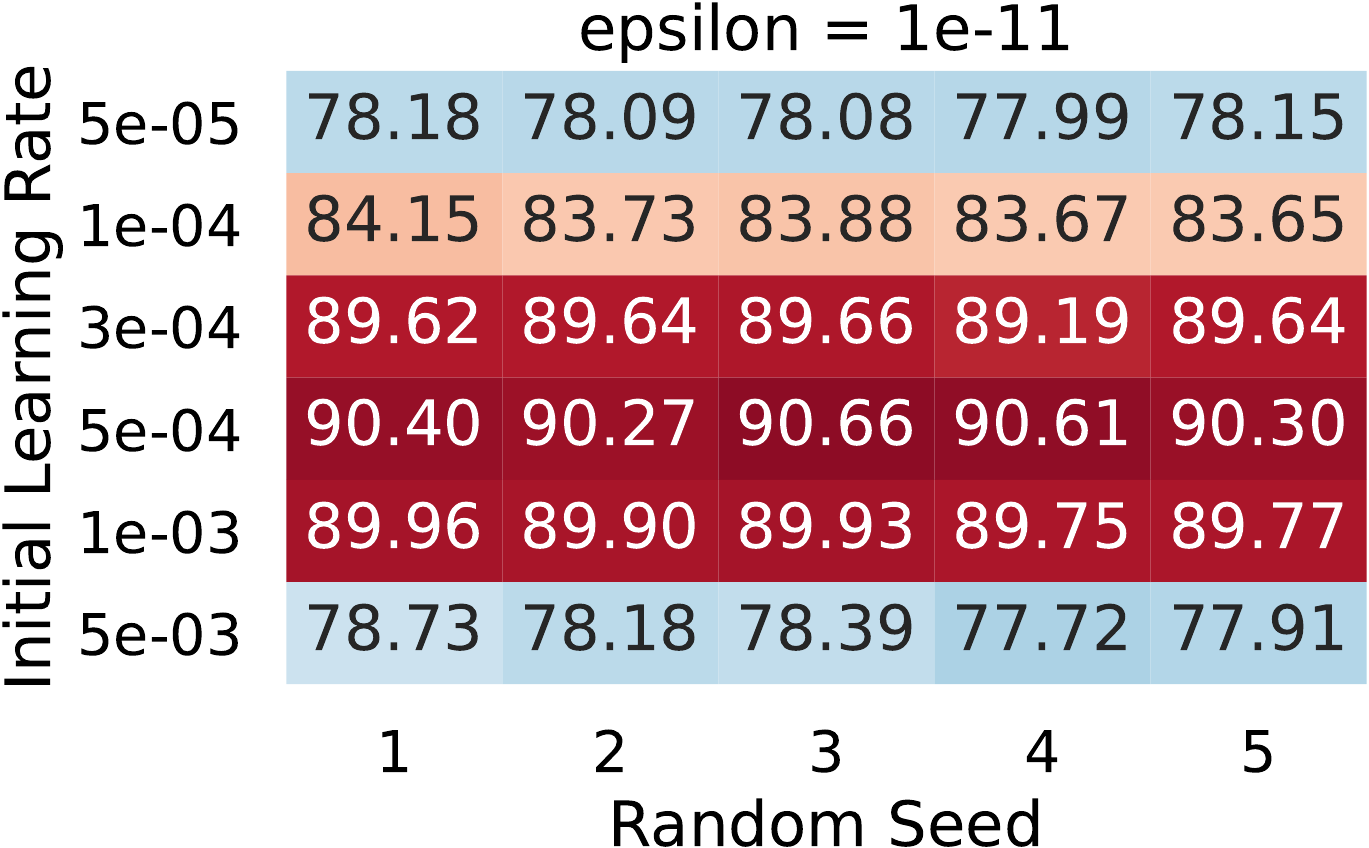}
\vspace{-.35cm}
\caption{Heatmap logs of test accuracy of VGG-16 on CIFAR-10 for Adam for each initial learning rate and random seed for different small values of Adam's $\epsilon$ HP. These results correspond to those graphed in Figure \ref{fig:full-wilson}.}
\label{tab:wilson-adam-1}
\end{figure}

\begin{figure}[H]

\includegraphics[scale=.5]{section/figure/adam_tuned_table_1e-04-crop.pdf}
\includegraphics[scale=.5]{section/figure/adam_tuned_table_1e-05-crop.pdf}

\includegraphics[scale=.5]{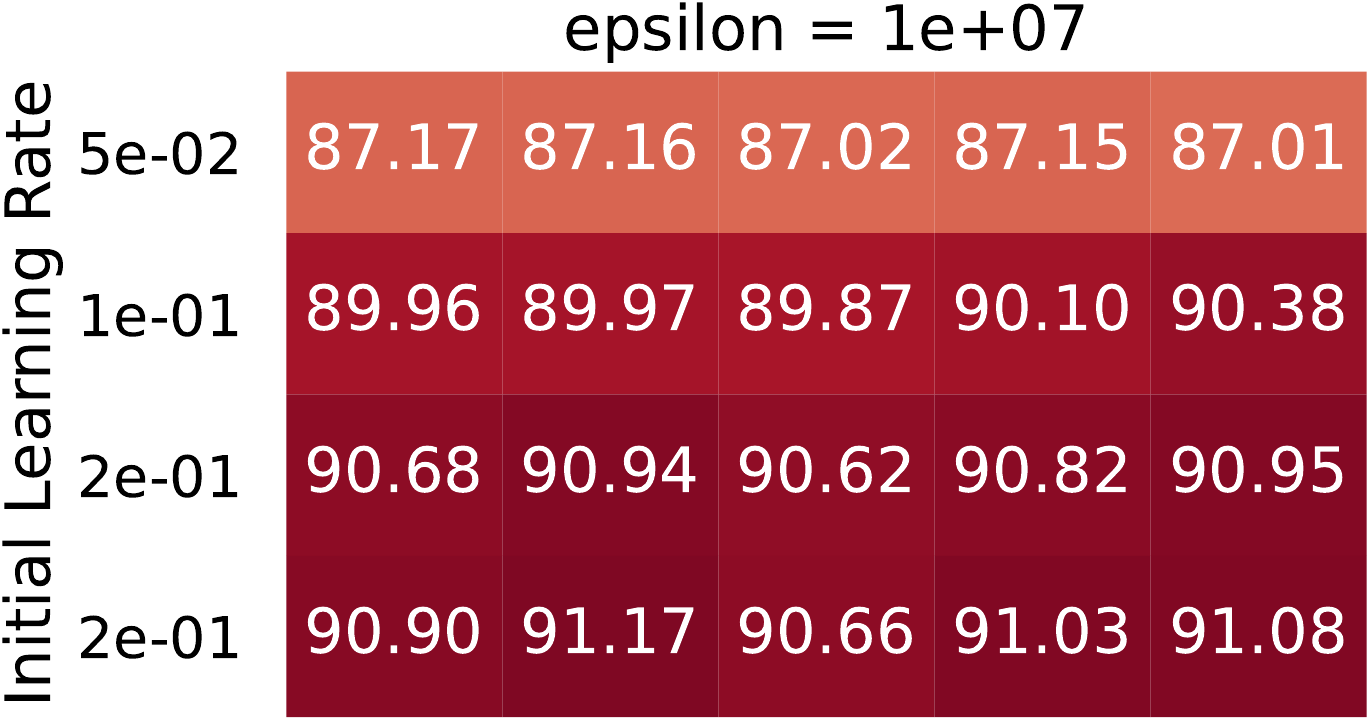}
\includegraphics[scale=.5]{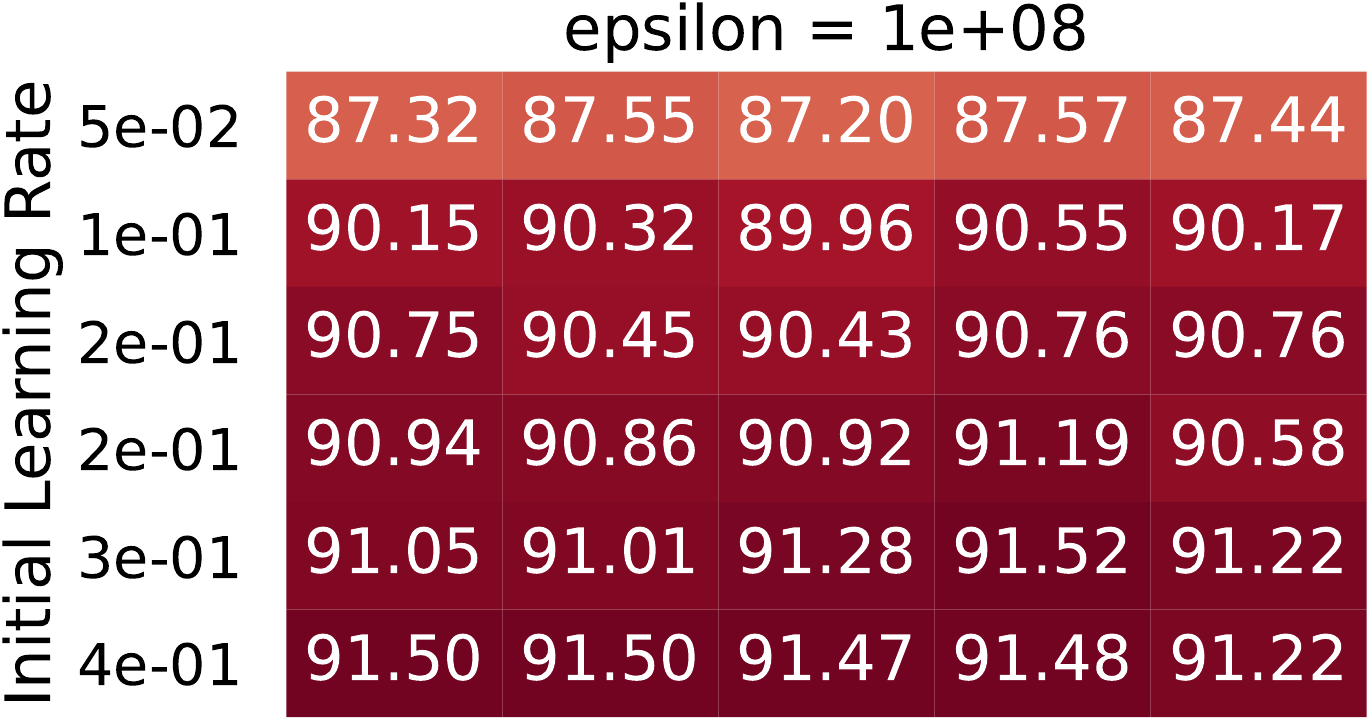}

\includegraphics[scale=.5]{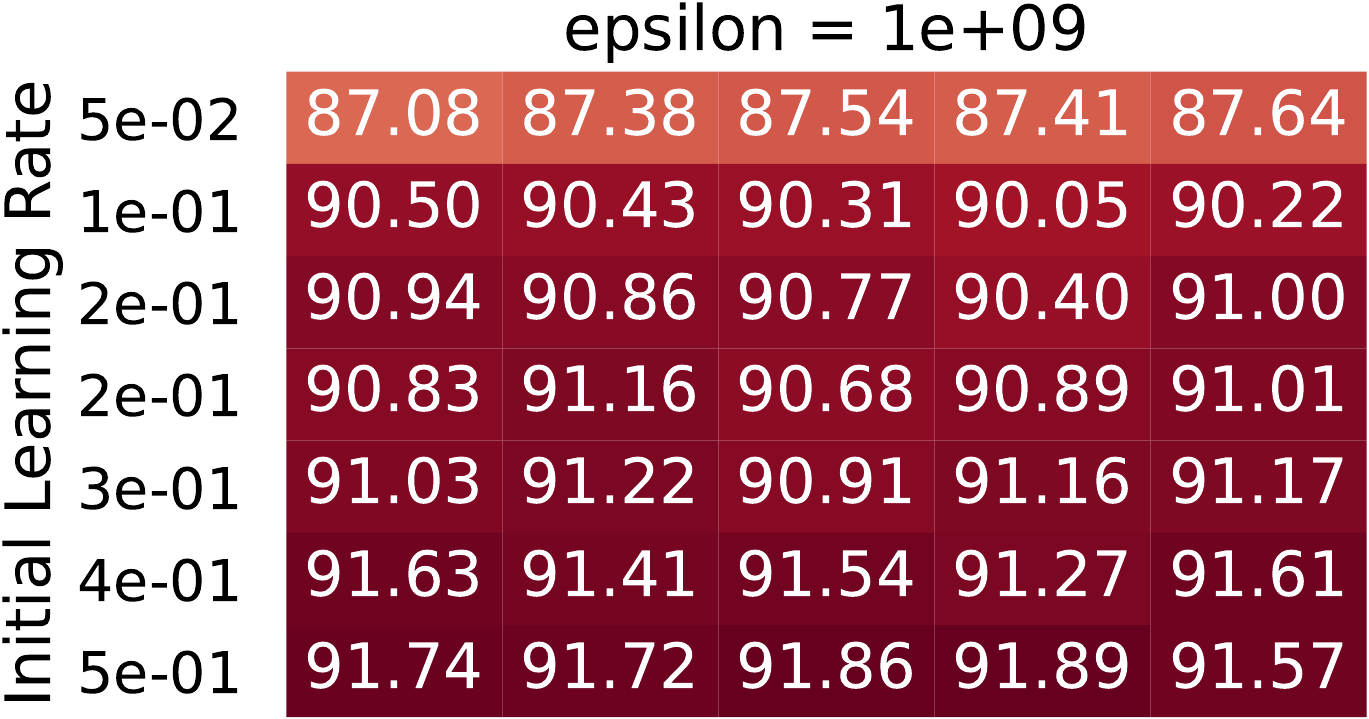}
\includegraphics[scale=.5]{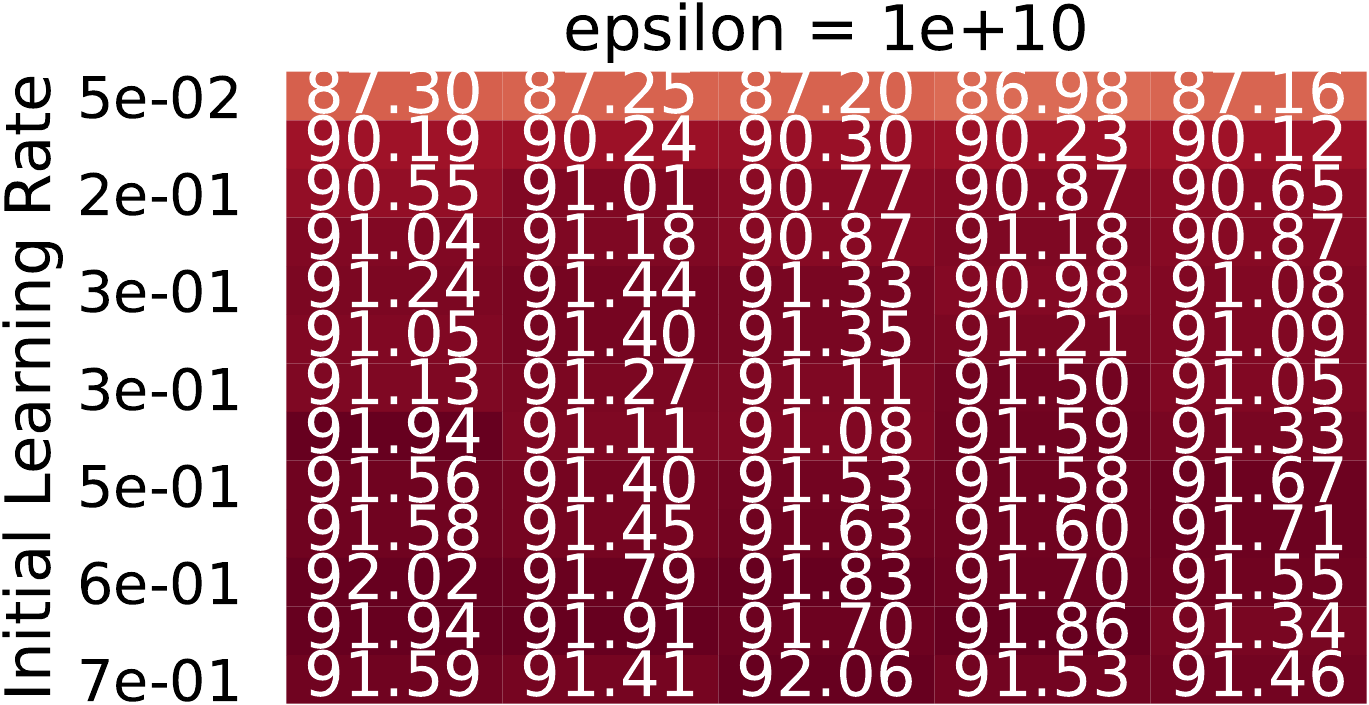}

\includegraphics[scale=.5]{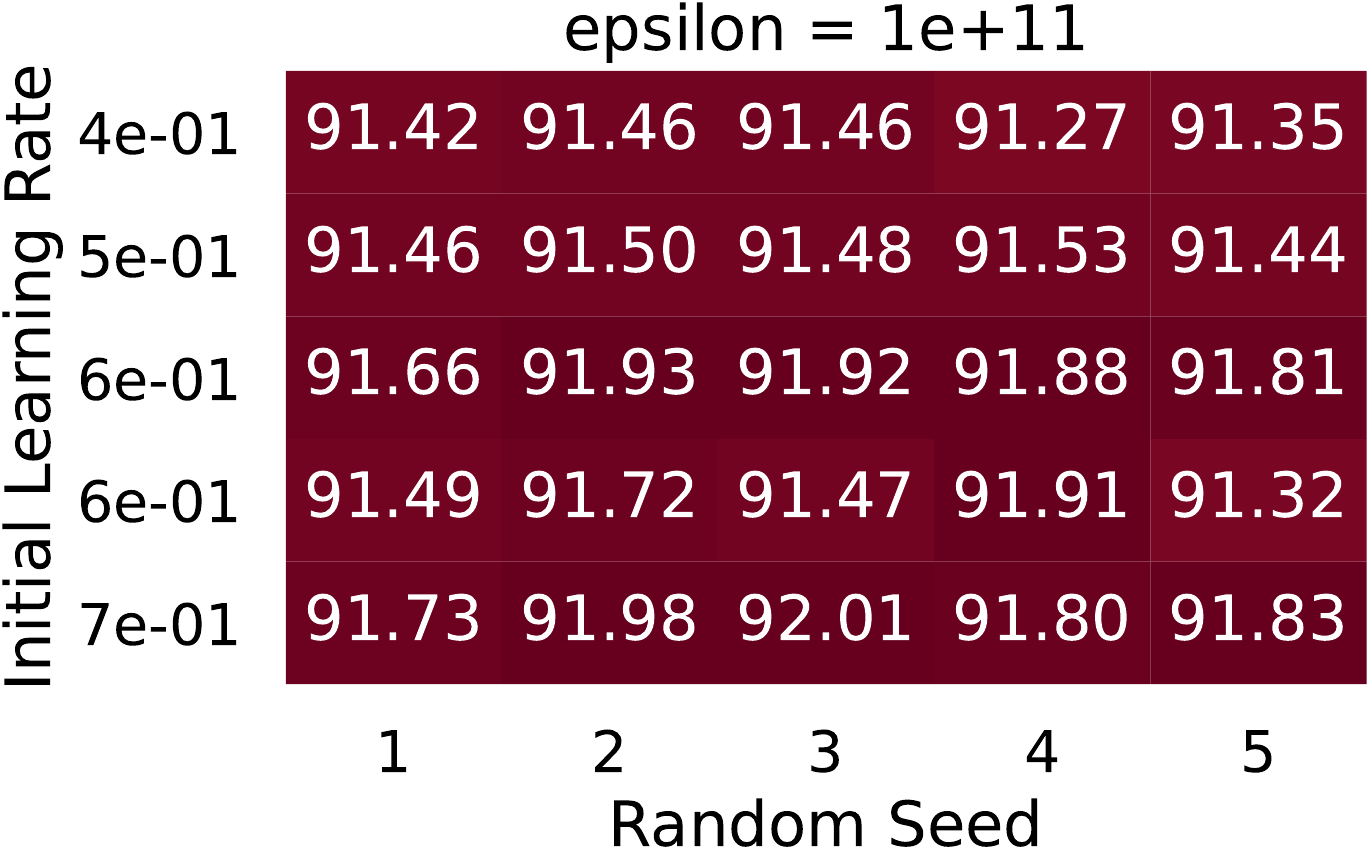}
\includegraphics[scale=.5]{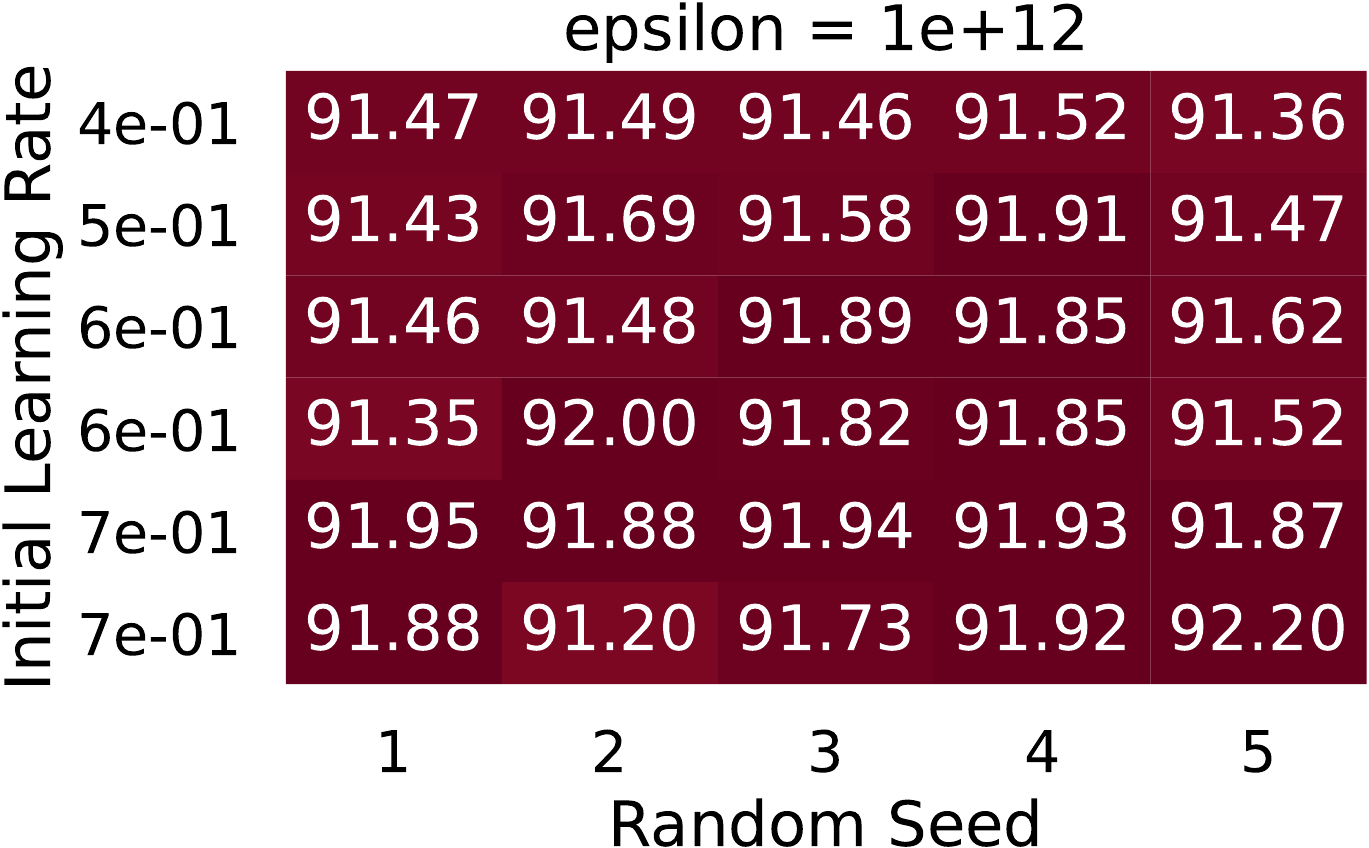}
\caption{Heatmap logs of test accuracy of VGG-16 on CIFAR-10 for Adam for each initial learning rate and random seed for different values of Adam's $\epsilon$ using our expanded search space. These logs reflect the results graphed in Figure \ref{fig:full-our-Adam}.}
\label{tab:wilson-adam-2}
\end{figure}


\newpage
\subsection{Empirical Deception Illustration using \citet{merity2016pointer}}
In addition to the computer vision experiments of \citet{wilson2017marginal}, we also show a separate line of experiments from NLP: training an LSTM on Wikitext-2 using Nesterov and Heavy Ball as the optimizers. We illustrate deception (i.e., the possibility of drawing inconsistent conclusions) using two different sets of hyper-HPs to configure HPO grids for tuning the learning rate. We run ten replicates for each optimizer / grid combination (a total of 40 runs). 
We run these experiments using the same hardware as described in Appendix \ref{sec:wilson:setup}.

\vspace{-.2cm}
\begin{figure}[h!]
 \begin{center}

    \includegraphics[width=0.44\textwidth]{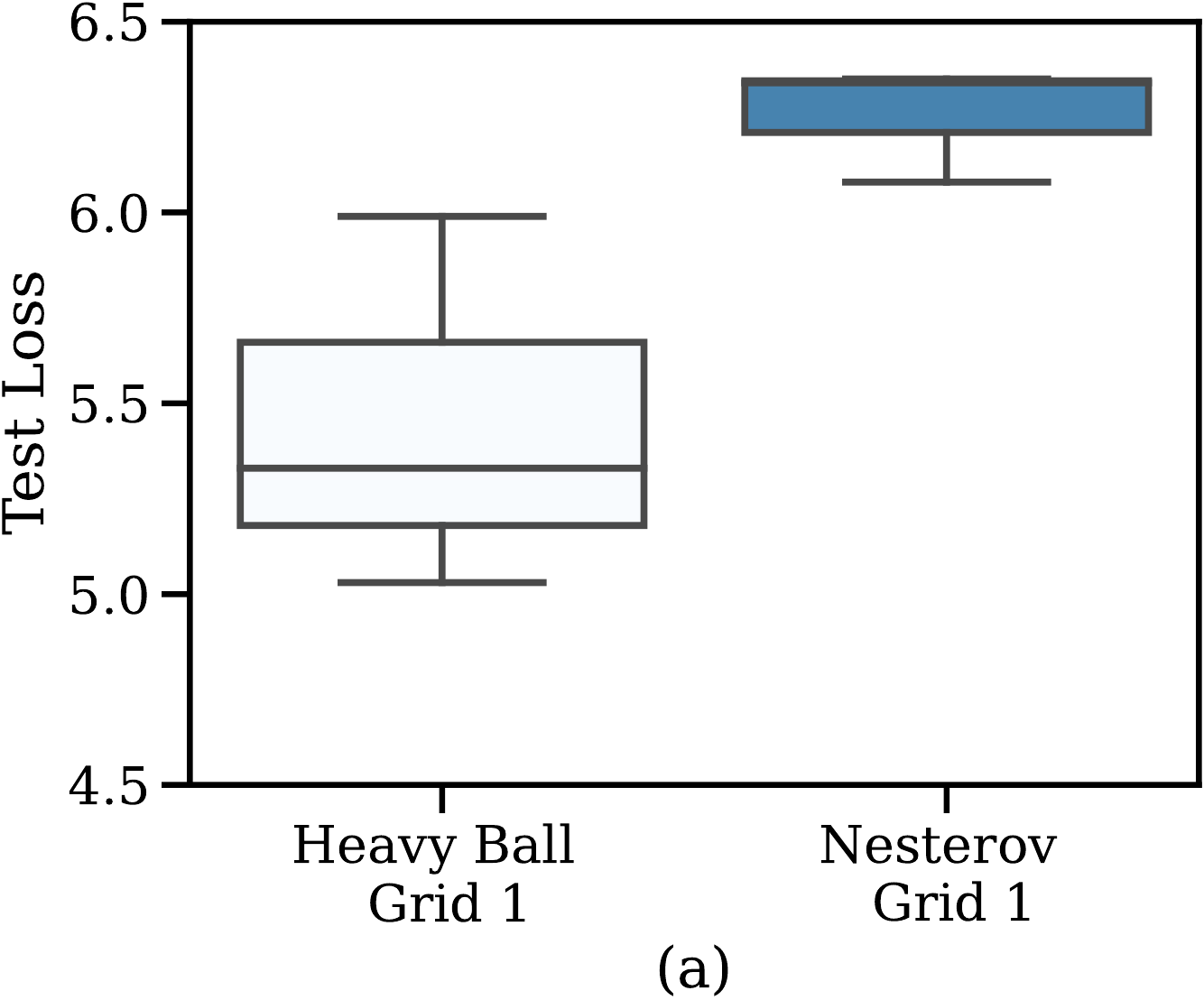} \hspace{.5cm}
\includegraphics[width=0.44\textwidth]{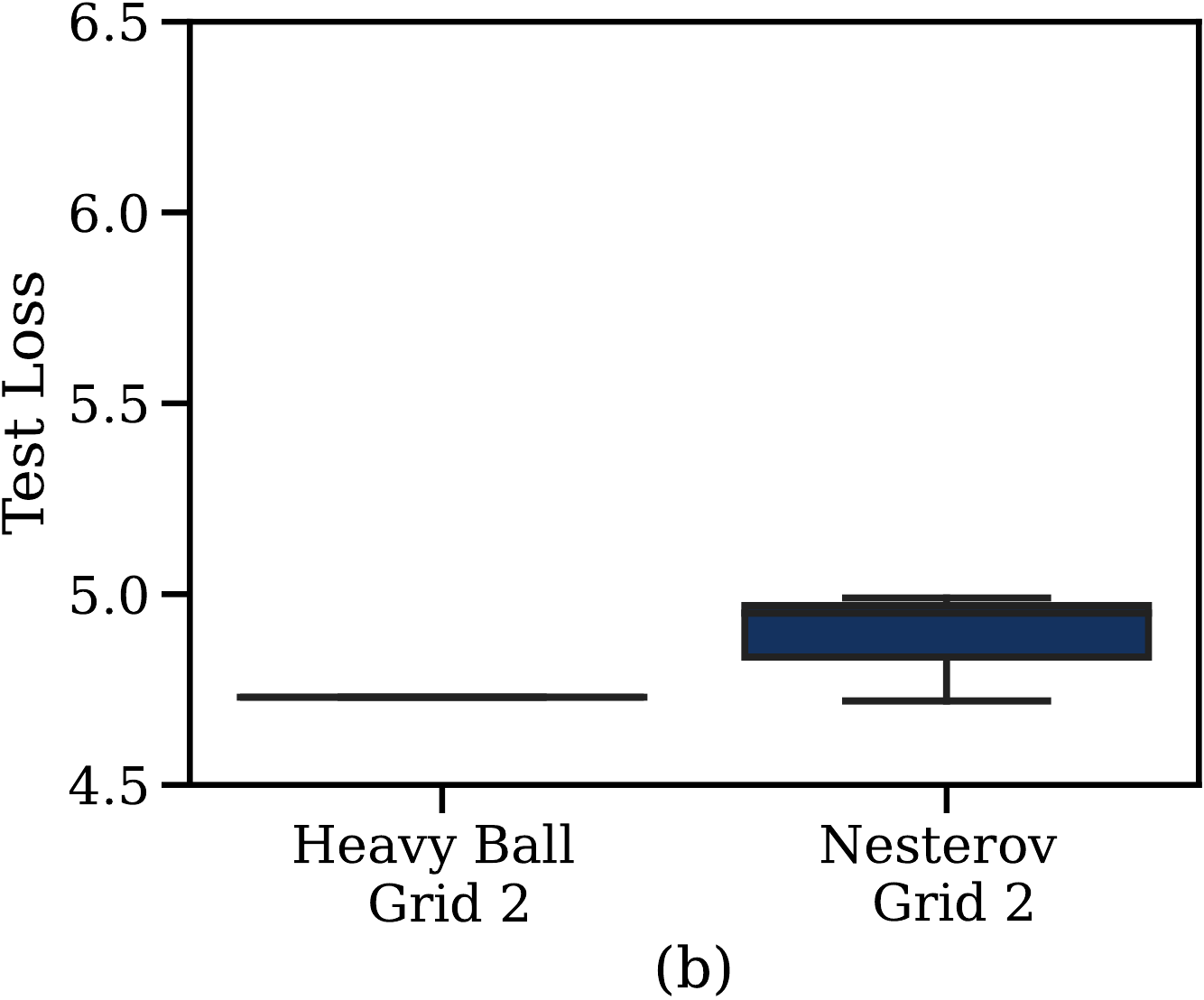}
 \vspace{-.2cm}

    \caption{Demonstrating the possibility of drawing inconsistent conclusions from HPO (what we shorthand \emph{hyperparameter deception}) LSTM on Wikitext-2 using Nesterov and Heavy Ball as the optimizers. Each box plot represents a log. In (a), we use the grid $\alpha={1,5,10,15,20,25,30,35,40}$, from which we can reasonably conclude that Nesterov outperforms HB. In (b), we use the grid $\alpha={10,20,30,40}$, from which we can reasonably conclude that HB outperforms Nesterov.}
    \label{fig:merity_deception}
  \end{center}
\end{figure}
\vspace{-.2cm}

\section{Section \ref{sec:ehpo} Appendix: Epistemic Hyperparameter Optimization} \label{app:sec:ehpo}

\subsection{Additional concrete interpretations of EHPO}

For concision, in the main text we focus on examples of EHPO procedures that compare the performance of different optimizers. However, it is worth noting that our definition of EHPO (Definition \ref{def:ehpo}) is more expansive than this setting. For example, it is possible to run EHPO to compare different models (perhaps, though not necessarily, keeping the optimizer fixed), to draw conclusions about the relative performance of different models on different learning tasks. 

\subsection{\citeauthor{descartes1996evildemon}' Evil Demon Thought Experiment}

Our formalization was inspired by \citeauthor{descartes1996evildemon}' evil genius/demon thought experiment. This experiment more generally relates to his use of systematic doubt in \emph{The Meditations} more broadly. It is this doubt/skepticism (and its relationship to possibility) that we find useful for the framing of an imaginary, worst-case adversary. In particular, we draw on the following quote, from which we came up with the term \emph{hyperparameter deception}:

\begin{quote}
    \emph{I will suppose...an evil genius, supremely powerful and clever, who has directed his entire effort at deceiving me. I will regard the heavens, the air, the earth, colors, shapes, sounds, and all external things as nothing but the bedeviling hoaxes of my dreams, with which he lays snares for my credulity...even if it is not within my power to know anything true, it certainly is within my power to take care resolutely to withhold my assent to what is false, lest this deceiver, however powerful, however clever he may be, have any effect on me.} -- \citeauthor{descartes1996evildemon}
\end{quote}

\begin{figure}[H]
  \begin{center}
    \includegraphics[width=.2\textwidth]{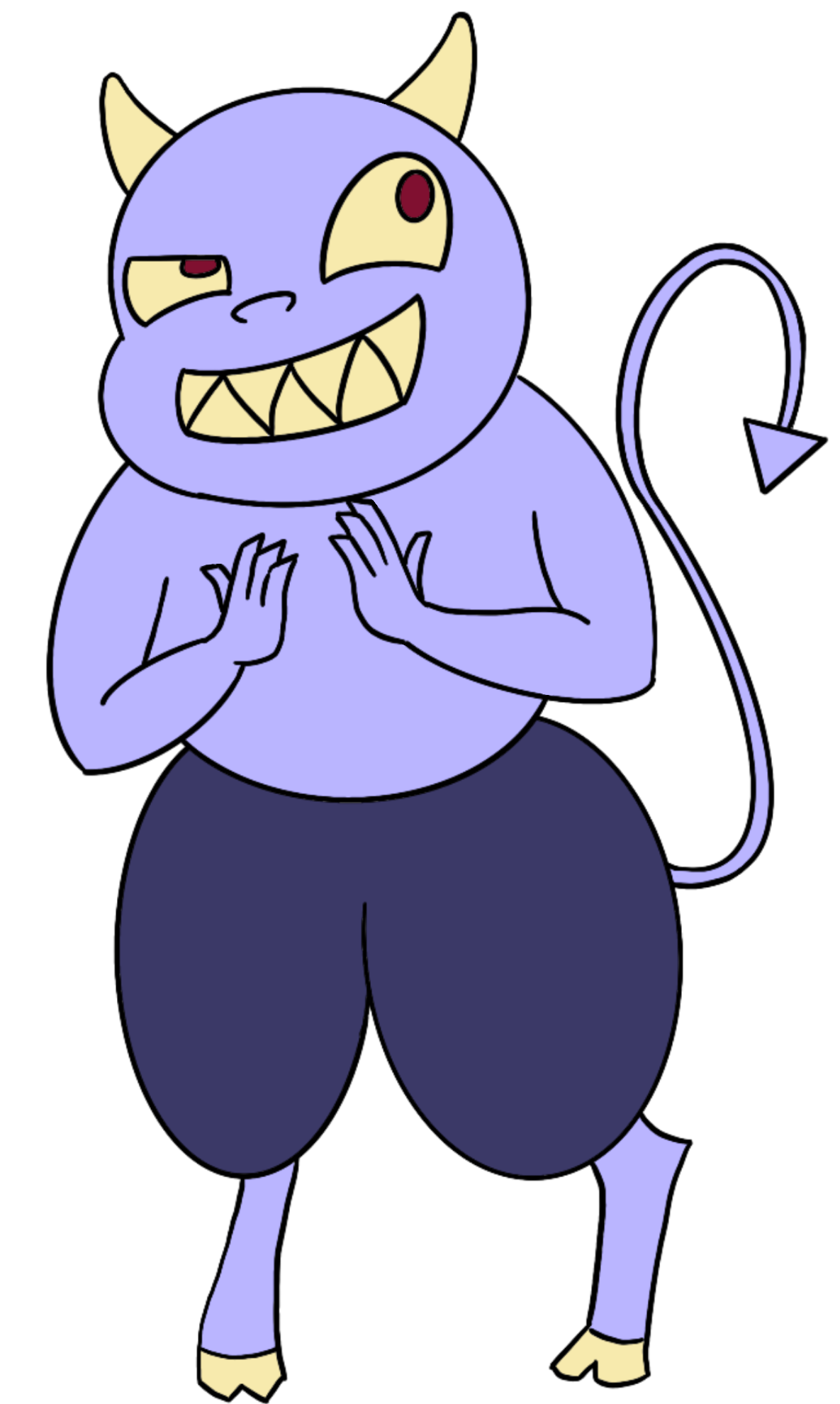}
    \caption{How the authors imagine the EHPO-running demon}
  \end{center}
\end{figure}

For more on the long (and rich) history of the use of imaginary demons and devils as adversaries---notably a different conception of an adversary than the potential real threats posed in computer security research---we refer the reader to \citet{canales2020devil}.

\section{Section \ref{sec:formalizing} Appendix: Modal Logic Formalization} \label{app:sec:logic}

\subsection{Further Background on Modal Logic}

We first provide the necessary background on modal logic, which will inform the proofs in this appendix (Appendix \ref{app:sec:kripkeaxioms}). We then describe our possibility logic---a logic for representing the possible results of the evil demon running EHPO---and prove that it is a valid modal logic (Appendix \ref{app:sec:ehpologic}). We then present a primer on modal belief logic (Appendix \ref{app:sec:belief}), and suggest a proof for the validity of combining our modal possibility logic with modal belief logic (Appendix \ref{app:sec:multimodal}).

\subsubsection{Axioms from Kripke Semantics} \label{app:sec:kripkeaxioms}
Kripke semantics in modal logic inherits all of the the axioms from propositional logic, which assigns values  $T$ and $F$ to each atom $p$, and adds two operators, one for representing \emph{necessity} ($\necessary$)  and one for \emph{possibility} ($\possible$).
\begin{itemize}
    \item $\necessary p$ reads ``It is necessary that $p$".
    \item $\possible p$ reads ``It is possible that $p$".
\end{itemize}

The $\possible$ operator is just syntactic sugar, as it can be represented in terms of $\lnot$ and $\necessary$:

\begin{equation} \label{app:eq:sugar}
    \possible p \equiv \lnot \necessary \lnot p
\end{equation}
which can be read as:
\begin{center}
``It is possible that $p$" is equivalent to ``It is not necessary that not $p$."
\end{center}

The complete set of rules is as follows:
\begin{itemize}
    \item Every atom $p$ is a sentence.
    \item If $D$ is a sentence, then 
    \begin{itemize}
        \item $\lnot D$ is a sentence.
        \item $\necessary D$ is a sentence.
        \item $\possible D$ is a sentence.
    \end{itemize}
    \item If $D$ and $E$ are sentences, then
    \begin{itemize}
        \item $D \land E$ is a sentence.
        \item $D \lor E$ is a sentence.
        \item $D \rightarrow E$ is a sentence.
        \item $D \leftrightarrow E$ is a sentence
    \end{itemize}
    \item $\necessary(\mathcal{D} \rightarrow \mathcal{E}) \rightarrow (\necessary \mathcal{D} \rightarrow \necessary \mathcal{E})$ (Distribution)
    \item $\mathcal{D} \rightarrow \necessary \mathcal{D}$ (Necessitation)
\end{itemize}

\subsubsection{Possible Worlds Semantics} \label{app:sec:pw}
Modal logic introduces a notion of \emph{possible worlds}. Broadly speaking, a possible world represents the state of how the world \emph{is} or potentially \emph{could be} \cite{chellas1980book, garson2018stanford}. Informally, $\necessary D$ means that $D$ is true at \emph{every} world (Equation \ref{app:eq:nec}); $\possible D$ means that $D$ is true at \emph{some} world (Equation \ref{app:eq:poss}).

Possible worlds give a different semantics from more familiar propositional logic. In the latter, we assign truth values $\{T, F\}$ to propositional variables $p \in \mathcal{P}$, from which we can construct and evaluate sentences $D \in \mathcal{D}$ in a truth table. In the former, we introduce a set of possible worlds, $\mathcal{W}$, for which each $w \in \mathcal{W}$ has own truth value for each $p$. This means that the value of each $p$ can differ across different worlds $w$. Modal logic introduces the idea of valuation function,
\begin{equation*}
    \mathcal{V}: (\mathcal{W}\times \mathcal{D}) \rightarrow \{T, F\}
\end{equation*}
to assign truth values to logical sentences at different worlds. This in turn allows us to express the formulas, axioms, and inference rules of propositional logic in terms of $\mathcal{V}$. For example,

\begin{equation*}
    \mathcal{V}(w, \lnot D) = T \leftrightarrow \mathcal{V}(w, D) = F
\end{equation*}

There are other rules that each correspond to a traditional truth-table sentence evaluation, but conditioned on the world in which the evaluation occurs. We omit these for brevity and refer the reader to \citet{chellas1980book}.

We do include the valuation rules for the $\necessary$ and $\possible$ operators that modal logic introduces (Equations \ref{app:eq:nec} \& \ref{app:eq:poss}). To do so, we need to introduce one more concept: The accessibility relation, $\mathcal{R}$. $\mathcal{R}$ provides a frame of reference for one particular possible world to access other possible worlds; it is a way from moving from world to world. So, for an informal example, $\mathcal{R}w_1w_2$ means that $w_2$ is possible relative to $w_1$, i.e. we can reach $w_2$ from $w_1$. Such a relation allows for a world to be possible relative potentially to some worlds but not others. More formally, 

\begin{align*}
R \subseteq \mathcal{W} \times \mathcal{W}
\end{align*}

Overall, the important point is that we have a collection of worlds $\mathcal{W}$, an accessibility relation $\mathcal{R}$, and a valuation function $\mathcal{V}$, which together defines a Kripke model, which captures this system:

\begin{align*}
    \mathcal{M} = \langle \mathcal{W}, \mathcal{R}, \mathcal{V} \rangle
\end{align*}

Finally, we can give the valuation function rules for $\necessary$ and $\possible$: 

\begin{equation} \label{app:eq:nec}
    \mathcal{V}(w, \necessary D) = T \leftrightarrow \forall w', (\mathcal{R}ww' \rightarrow \mathcal{V}(w', D) = T)
\end{equation}

\begin{equation} \label{app:eq:poss}
    \mathcal{V}(w, \possible D) = T \leftrightarrow \exists w', (\mathcal{R}ww' \land \mathcal{V}(w', D) = T)
\end{equation}

Informally, for $\necessary D$ to be true in a world, it must be true in every possible world that is reachable by that world. For $\possible D$ to be true in a world, it must be true in some possible world that is reachable by that world.

\subsection{Our Multimodal Logic Formulation} \label{app:sec:ourlogic} 

\subsubsection{A Logic for Reasoning about the Conclusion of EHPO} \label{app:sec:ehpologic}
As in Section \ref{sec:formalizing}, we can define the well-formed formulas of our indexed modal logic\footnote{For an example of another indexed modal logic concerning probability, please refer to \citet{heifeitz1998indexed}.} recursively in Backus-Naur form, where $t$ is any real number and $P$ is any atomic proposition

\begin{align} \label{app:eq:ehposyntax}
    \kappa \coloneqq P\;|\;\lnot \kappa\;|\;\kappa \land \kappa \;|\; \possible_t \kappa
\end{align}

where $\kappa$ is a well-formed formula.

As we note in Section \ref{sec:formalizing}, where we first present this form of defining modal-logic, $\necessary$ is syntactic sugar, with $\necessary p \equiv \lnot \possible \lnot p$ (which remains true for our indexed modal logic). Similarly, ``or'' has $p \lor q \equiv \lnot (\lnot p \land \lnot q)$ and ``implies'' has $p \rightarrow q \equiv \lnot p \lor q$, which is why we do not include them for brevity in this recursive definition.

We explicitly define the relevant semantics for $\possible_t$ for reasoning about the demon's behavior in running EHPO. For clarity, we replicate that definition of the semantics of expressing the possible outcomes of EHPO conducted in bounded time (Definitions \ref{def:strategy} \& \ref{def:ehpologic}, respectively) below:

\begin{definition*}
A randomized \textbf{strategy} $\boldsymbol{\sigma}$ is a function that specifies which action the demon will take. Given $\mathcal{L}$, its current set of logs, $\boldsymbol{\sigma(\mathcal{L})}$ gives a distribution over concrete actions, where each action is either 1) running a new $H$ with its choice of hyper-HPs $c$ and seed $r$ 2) erasing some logs, or 3) returning. We let $\Sigma$ denote the set of all such strategies.
\end{definition*}

We can now define what the demon can reliably bring about, in terms of executing a strategy in bounded time:

\begin{definition*}
Let $\boldsymbol{\sigma[\mathcal{L}]}$ denote the logs output from executing strategy $\sigma$ on logs $\mathcal{L}$, and let $\boldsymbol{\tau_\sigma(\mathcal{L})}$ denote the total time spent during execution. $\tau_\sigma(\mathcal{L})$ is equivalent to the sum of the times $T$ it took each HPO procedure $H \in \mathcal{H}$ executed in strategy $\sigma$ to run.
Note that both $\sigma[\mathcal{L}]$ and $\tau_\sigma(\mathcal{L})$ are random variables, as a function of the randomness of selecting $G$ and the actions sampled from $\sigma(\mathcal{L})$. For any formula $p$ and any $t \in \mathbb{R}_{>0}$, we say $\mathcal{L} \models \possible_t p$, i.e. ``$\mathcal{L}$ models that it is possible $p$ in time $t$,'' if 
\[
    \text{there exists a strategy } \sigma \in \Sigma, \text{ such that } \;\; \mathbb{P}(\sigma[\mathcal{L}] \models p) = 1 \; \text{ and } \; \mathbb{E}[\tau_\sigma(\mathcal{L})] \le t.
\]
\end{definition*}

\subsubsection{A Possible Worlds Interpretation} \label{app:sec:pw_interpretation}

Drawing on the possible worlds semantics that modal logic provides (Section \ref{app:sec:pw}), we can define specific possible worlds semantics for our logic for expressing the actions of the demon in EHPO from above. 

\begin{definition} \label{app:def:possibleworld}
    A \textbf{possible world} represents the set of logs $\mathcal{L}$ the demon has produced at time $\tau_\sigma(\mathcal{L})$ (i.e., after concluding running EHPO) and the set of formulas $\mathcal{P}$ that are modeled from $\mathcal{L}$ via $\mathcal{F}$.
\end{definition}

Therefore, different possible worlds represent the states that \emph{could have existed} if the evil demon had executed different strategies (Definition \ref{def:strategy}). In other words, if it had performed EHPO with different learning algorithms, different HPO procedures, different hyper-hyperparameter settings, different amounts of time (less than the total upper bound), different learning tasks, different models, etc... to produce a different set of logs $\mathcal{L}$ and corresponding set of conclusions $\mathcal{P}$. 

In this formulation, the demon has knowledge of all possible worlds; it is trying to fool us about the relative performance of algorithms by showing as an intentionally deceptive world. Informally, moving from world to world (via an accessibility relation) corresponds to the demon running more passes of HPO to produce more logs to include in $\mathcal{L}$.

\subsubsection{Syntax and Semantics for the Logic Modeling the Demon Running the EHPO} \label{app:sec:ehpoproperties}

We provide proofs and intuitions of the axioms of our EHPO logic in this section, based on a correspondence with un-indexed modal logic.

We remind the reader that the following are the axioms of our indexed modal logic:

\begin{align*}
\vdash (p \rightarrow q) \rightarrow (\possible_t p \rightarrow \possible_t q) && \textit{(necessitation + distribution)} \\
p \rightarrow \possible_t p && \textit{(reflexivity)} \\
\possible_t\possible_s p \rightarrow \possible_{t+s} p && \textit{(transitivity)}
\\\possible_s\necessary_t p \rightarrow \necessary_t p && \textit{(symmetry)},
\\\possible_t(p \land q) \rightarrow (\possible_t p \land \possible_t q) && \textit{(distribution over } \land \textit{)}
\end{align*}
In short, to summarize these semantics---the demon has knowledge of all possible hyper-hyperparameters, and it can pick whichever ones it wants to run EHPO within a bounded time budget $t$ to realize the outcomes it wants: $\possible_t p$ means it can realize $p$.

We inherit distribution and necessitation from un-indexed modal logic; they are axiomatic based on Kripke semantics. We provide greater intuition and proofs below.

\textbf{Notes on necessitation for $\necessary_t$}:

Necessitation for our indexed necessary operator can be written as follows:

\begin{align*}
    \vdash p \rightarrow \necessary_i p
\end{align*}

As we note in Section \ref{sec:formalizing}, $\vdash$ just means here that $p$ is a theorem of propositional logic. So, if $p$ is a theorem, then so is $\necessary_t p$. By theorem we just mean that $p$ is provable by our axioms (these being the only assumptions we can use); so whenever $p$ fits this definition, we can say $\necessary_t p$.

For our semantics, this just means that when $p$ is a theorem, it is necessary that $p$ in time $t$. 

\textbf{Distribution for $\necessary_t$}:

\begin{align*}
\necessary_t (p \rightarrow q) \rightarrow (\necessary_t p \rightarrow \necessary_t q) \end{align*}

We provide three ways to verify distribution over implication for $\necessary_t$. From this, we will prove distribution over implication for $\possible_t$

\begin{enumerate}[A.]
\item The first follows from an argument about the semantics of possible worlds from the Kripke model of our system (Sections \ref{app:sec:pw} \& \ref{app:sec:pw_interpretation}). 
    \begin{enumerate}[i.]
        \item It is fair to reason that distribution is self-evident given the definitions of implication ($\rightarrow$, formed from $\lnot$ and $\lor$ in our syntax for well-formed formulas for our EPHO logic, given at (\ref{app:eq:ehposyntax}) and necessity ($\necessary_t$, formed from $\lnot$ and $\possible_t$ in our syntax for well-formed formulas for our EHPO logic, given at (\ref{app:eq:ehposyntax})).
        \item Similarly, we can further support this via our semantics of possible worlds. 
        
        We can understand $\necessary_t p$ to mean, informally, that it an adversary does adopt a strategy $\sigma$ that is guaranteed to cause the desired conclusion $p$ to be the case while take at most time $t$ in expectation. Formally, as an ``necessary'' analog to the semantics of $\possible_t$ given in Definition \ref{def:ehpologic}:
        
        For any formula $p$, we say $\mathcal{L} \models \necessary_t p$ if and only if
        \[
        \forall \sigma \in \Sigma, \; \mathbb{P}(\sigma[\mathcal{L}] \models p) = 1 \; \land \; \mathbb{E}[\tau_\sigma(\mathcal{L})] \le t.
        \]

        Given $p \rightarrow q$ is true in \textbf{all accessible worlds} (i.e, the definition of necessary), then we can say that $q$ is true in all accessible worlds whenever $p$ is true in all accessible worlds. As in i. above, this just follows / is axiomatic from the definitions of necessity and implication for Kripke semantics. 
    \end{enumerate}
\item We can also prove distribution by contradiction.
    \begin{enumerate}[i.]
        \item Suppose that the distribution axiom does not hold. That is, the hypothesis
        \begin{align*}
            \necessary_t(p \rightarrow q)
        \end{align*}
        is true and the conclusion
        \begin{align*}
            \necessary_t p \rightarrow \necessary_t q
        \end{align*}
        is false. 
        \item By similar reasoning, from above $\necessary_t p \rightarrow \necessary_t q$ being false, we can say that $\necessary_t p$ is true and $\necessary_t q$ is false. 
        \item We can use Modal Axiom M (reflexivity, proven in the next section) to say $\necessary_t p \rightarrow p$. Since $\necessary_t p$ is true, we can use \emph{modus ponens} to determine that $p$ is true.
        \item We can also say
        \begin{align*}
            \necessary_t(p \rightarrow q) \rightarrow (p \rightarrow q) && (\textit{By Modal Axiom M (reflexivity)})
        \end{align*}
        \item Since we $\necessary_t(p \rightarrow q)$ is true from above, we can conclude via \emph{modus ponens} that $p \rightarrow q$ must also be true. 
        \item We concluded above that $p$ is true, so we can again use \emph{modus ponens} and the fact that $p \rightarrow q$ is true to conclude that $q$ is true.
        \item By necessitation, we can then also say $q \rightarrow \necessary_t q$, and conclude that $\necessary_t q$ is true. This is a contradiction, as above we said that $\necessary_t$ is false. 
        \item Therefore, by contradiction, $\necessary_t(p \rightarrow q) \rightarrow (\necessary_t p \rightarrow \necessary_t q)$ is proved. 
        
    \end{enumerate}
\item We can separately take an intuitionistic approach to verify the distribution axiom \cite{bezhanishvili2009intuitionistic, vanatten2004brouwer}:
    \begin{enumerate}[i.] 
        \item Let $b$ be an \textbf{actual proof} of $p \rightarrow q$ so that we can say $a.b$ is a proof of $\necessary_t(p \rightarrow q$).
        \item Let $d$ be an \textbf{actual proof} of $p$ so that we can say $c.d$ is a proof of $\necessary_t p$. 
        \item From i. and ii., $b(d)$ is an \textbf{actual proof} of $q$, i.e. $b$ (an actual proof of $p \rightarrow q$) is supplied $d$ (an actual proof of $p$), and therefore can conclude $q$ via an actual proof. 
        \item From iii., we can say this results in a proof of $\necessary_t q$, i.e. $e.[b(d)]$.
        \item The above i.-iv. describes a function, $f: a.b \rightarrow f_{(a.b)}$. In other words, given \textbf{any proof} $a.b$ (i.e., of $\necessary_t(p \rightarrow q)$) we can return function $f_{(a.b)}$, which turns \textbf{any proof} $c.d$ (i.e., of $\necessary_t p$) into a proof $e.[b(d)]$ (i.e., of $\necessary_t q$).
        \item $f_{(a.b)}$ is thus a proof of $\necessary_t p \rightarrow \necessary_t q$.
        \item From i.-vi., we gone from $a.b$ (a proof of $\necessary_t(p \rightarrow q)$) to a proof of $\necessary_t p \rightarrow \necessary_t q$, i.e. have intuitionistically shown that $\necessary_t(p \rightarrow q) \rightarrow (\necessary_t p \rightarrow \necessary_t q)$
    \end{enumerate}
\end{enumerate}

\textbf{Distribution and $\possible_t$}:

We provide the following axiom in our logic:

\begin{align*}
\vdash (p \rightarrow q) \rightarrow (\possible_t p \rightarrow \possible_t q) && (\textit{necessitation and distribution})
\end{align*}

and we now demonstrate it to be valid. 

\begin{align*}
    \vdash (p \rightarrow q) \rightarrow \necessary_t(p \rightarrow q) && (\textit{necessitation})
    \\\rightarrow \;
    \necessary_t (\lnot q \rightarrow \lnot p) && (\textit{modus tollens})
    \\\rightarrow \;
    (\necessary_t \lnot q \rightarrow \necessary_t  \lnot p) && (\textit{distribution})
    \\\rightarrow \;
    (\lnot \necessary_t \lnot p \rightarrow \lnot \necessary_t  \lnot q) && (\textit{modus tollens})
    \\\rightarrow \;
    (\possible_t p \rightarrow \possible_t  q) && (\possible_t a \equiv \lnot \necessary_t \lnot a)
\end{align*}

This concludes our proof, for how the axioms are jointly stated. 

Further, we could also say
\begin{align*}
    (p \rightarrow q) \rightarrow \possible_t (p \rightarrow q) && (\textit{Modal axiom M (reflexivity)})
\end{align*}

And therefore also derive distribution over implication for possibility: 

\begin{align*}
    \possible_t(p \rightarrow q) \rightarrow (\possible_t p \rightarrow \possible_t q)
\end{align*}

\textbf{Modal Axiom M: Reflexivity}

\begin{align*}
    p \rightarrow \possible_t p
\end{align*}

This axiom follows from how we have defined the semantics of our indexed modal logic (Definition \ref{def:ehpologic}). It follows from the fact that the demon could choose to do nothing.

We can provide a bit more color to the above as follows:

We can also derive this rule from necessitation, defined above (and from the general intuition / semantics of modal logic that necessity implies possibility). First, we can say that necessity implies possibility. We can see this a) from a possible worlds perspective and b) directly from our axioms. From a possible worlds perspective, this follows from the definition of the operators. Necessity means that there is truth at every accessible possible world, while possibility means there is truth at some accessible possible world, which puts that possible truth in time $t$ as a subset of necessary truth in time $t$. From the axioms, we verify

\begin{align*}
    \necessary_t p \rightarrow \possible_t p  & & (\textit{Theorem to verify, which also corresponds to Modal Axiom D (serial)}) \\
    \lnot \necessary_t p \lor \possible_t p && (\textit{Applying } p \rightarrow q \textit{ is equiv. }\lnot p \lor q) \\
    \possible_t \lnot p \lor \possible_t p && (\textit{By modal conversion, } \lnot \necessary_t p \rightarrow \possible_t \lnot p) \\
    & & (\textit{Which for our semantics is tautological}) \\
\end{align*}

That is, in time $t$ it is possible that $p$ or it is possible that $p$, which allows for us also to not conclude anything (in the case that the demon chooses to do nothing).

We can then say,

\begin{align*}
    (\necessary_t p \rightarrow p)  \rightarrow \possible_t p && (\textit{By necessitation and } \necessary_t p \rightarrow \possible_t p ) \\
    p \rightarrow \possible_t p && (\textit{By concluding } p \textit{ from necessitation})
\end{align*}

Another way to understand this axiom is again in terms of possible worlds. We can say in our system that every world is possible in relation to itself. This corresponds to the accessibility relation $\mathcal{R}ww$. As such, an equivalent way to model reflexivity is in terms of the following:

\begin{align*}
    \necessary_t p \rightarrow p
\end{align*}

That is, if $\necessary_t p$ holds in world $w$, then $p$ also holds in world $w$, as is the case for $\mathcal{R}ww$. We can see this by proving $\necessary_t p \rightarrow p$ by contradiction. Assuming this were false, we would need to construct a world $w$ in which $\necessary_t p$ is true and $p$ is false. If $\necessary_t p$ is true at world $w$, then by definition $p$ is true at every world that $w$ accesses. For our purposes, this holds, as $\necessary_t p$ means that it is necessary for $p$ to be the case in time $t$; any world that we access from this world $w$ (i.e. by say increasing time, running more HPO) would require $p$ to hold. Since $\mathcal{R}ww$ means that $w$ accesses itself, that means that $p$ must also be true at $w$, yielding the contradiction.

\textbf{Modal Axiom 4: Transitivity}

\begin{align} \label{app:ax:a4}
    \possible_t\possible_s p \rightarrow \possible_{t+s} p
\end{align}

We can similarly understand transitivity to be valid intuitively from the behavior of the demon and in relation to the semantics of our possible worlds. We do an abbreviated treatment (in relation to what we say for reflexivity above) for brevity.

In terms of the demon, we note that in our semantics $\possible_t p$ means that it is possible for the demon to bring about conclusion $p$ via its choices in time $t$. Similarly, we could say the same for $\possible_s p$; this means it is possible for the demon to bring about conclusion $p$ in time $s$. If it is possible in time $t$ that it is possible in time $s$ to bring about $p$, this is equivalent in our semantics to saying that it is possible in time $t+s$ to bring about conclusion $p$.

We can understand this rule (perhaps more clearly) in terms of possible worlds and accessibility relations.

For worlds $w_n$,
\begin{align*}
    \forall w_{1}, \forall w_{2}, \forall w_{3}, \mathcal{R}w_{1}w_{2} \land \mathcal{R}w_{2}w_{3}\rightarrow \mathcal{R}w_{1}w_{3}
\end{align*}

In other words, this accessibility relation indicates that if $w_1$ accesses $w_2$ and if $w_2$ accesses $w_3$, then $w_1$ accesses $w_3$.

For understanding this in terms of relative possibility, we could frame this as, if $w_3$ is possible relative to $w_2$ and if $w_2$ is possible relative to $w_1$, then $w_3$ is possible relative to $w_1$. For our semantics of the demon, this means that in some time if in some time $b$ we can get to some possible world $w_3$ from when we're in $w_2$ and in time $a$ we can get to some possible world $w_2$ when we're in $w_1$, then in time $a+b$ we can get to $w_3$ from $w_1$

This axiom is akin to us regarding a string of exclusively possible or exclusively necessary modal operators as just one possible or necessary modal operator, respectively; we regard then regard sum of times as the amount of time it takes to bring about $p$ (again, being necessary or possible, respectively).

\textbf{Modal Axiom 5: Symmetry}

\begin{align} \label{app:ax:a5}
    \possible_s\necessary_t p \rightarrow \necessary_t p
\end{align}

We can similarly understand that our modal logic is symmetric; this is valid intuitively from the behavior of the demon. We further abbreviate our treatment for brevity. In terms of the demon, we note that in our semantics $\possible_s p$ means that it is possible for the demon to bring about conclusion $p$ via its choices in time $s$. We can also say $\necessary_t p$ means that it is necessary for the demon to bring about $p$ in time $t$. If it is possible in time $s$ that it is necessary in time $s$ to bring about $p$, this is equivalent in our semantics to saying that it is necessary in time $t$ to bring about conclusion $p$. In other words, we can disregard would could have possibly happened in time $s$ from the demon's behavior and only regard what was necessary in time $t$ for the demon to do in order to bring about $p$.

As another example, consider our reduction of $\possible_t \lnot \possible_t p$ to $\lnot \possible_t p$ in our proof for deriving a defended reasoner in Section \ref{sec:defense}. While the intuitive English reading (``It's possible that it's not possible that $p$") does not seem equivalent to this reduction (``It's not possible that $p$), it is in fact valid for our semantics. Think of this in terms of the demon. If $p$ cannot be brought about in time $t$ in expectation (where $t$ is a reasonable upper bound on compute time), then that's the end of it; it doesn't matter which operators come before it (any number of $\possible_t$ or $\necessary_t$). Adding possibility or necessity before that condition doesn't change that fact that it, for that upper bound $t$, it is not possible to bring about $p$.

This axiom is akin to us just regarding the rightmost modal operator when we have a mix of modal operators applied iteratively; we can disregard what was possible or necessary in the time prior to the rightmost operator, and say that what we can say about a sentence $p$ (whether it is possible or necessary) just relates to how much time the last operator required to bring about $p$.

\textbf{Derived axioms}

We can similarly derive other axioms of our indexed modal logic, form the axioms above. Notably,

\textbf{$\necessary_t$ distributes over $\land$}

\begin{align*}
    \necessary_t(p \land q) \rightarrow (\necessary_t p \land \necessary_t q) & & (\textit{$\necessary_t$ distributes over $\land$}) \\
    \textbf{Inner proof 1} \\
    p \land q \\
    p \\
    (p \land q) \rightarrow p \\
    \necessary_t ((p \land q) \rightarrow p) && (\textit{Necessitation}) \\
    \necessary_t (p \land q) \rightarrow \necessary_t p && (\textit{Distribution}) \\
    \necessary_t p  && (\textit{By assuming the hypothesis}) \\
    \textbf{Inner proof 2} \\
    p \land q \\
    q \\
    (p \land q) \rightarrow q \\
    \necessary_t ((p \land q) \rightarrow q) && (\textit{Necessitation}) \\
    \necessary_t (p \land q) \rightarrow \necessary_t q && (\textit{Distribution}) \\
    \necessary_t q  && (\textit{By assuming the hypothesis}) \\
    \necessary_t p \land \necessary_t q && (\textit{By inner proof 1, inner proof 2, assuming the hypothesis})
\end{align*}

\textbf{We can show a similar result for $\possible_t$ and $\land$, omitted for brevity.}

\textbf{$\possible_t$ distributes over $\lor$}

\begin{align*}
    \possible_t(p \lor q) \rightarrow (\possible_t p \lor \possible_t q) & & (\textit{$\possible$ distributes over $\lor$}) \\
    \lnot \necessary_t \lnot (p \lor q) \rightarrow (\possible_t p \lor \possible_t q) & & (\possible_t a \equiv \lnot \necessary_t \lnot a) \\
    \lnot \necessary_t (\lnot p \land \lnot q) \rightarrow (\possible_t p \lor \possible_t q) & & (\lnot(a \lor b) \equiv (\lnot a \land \lnot b)) \\
    \lnot (\necessary_t  \lnot p \land \necessary_t  \lnot q) \rightarrow (\possible_t p \lor \possible_t q) & & (\textit{$\necessary_t$ distributes over $\land$}) \\
    (\lnot \necessary_t  \lnot p \lor \lnot \necessary_t  \lnot q) \rightarrow (\possible_t p \lor \possible_t q) & & (\lnot(a \land b) \equiv (\lnot a \lor \lnot b)) \\
    (\possible_t p \lor \possible_t q) \rightarrow (\possible_t p \lor \possible_t q) & & (\possible_t a \equiv \lnot \necessary_t \lnot a) \\
\end{align*}

\textbf{We can show a similar result for $\necessary_t$ and $\lor$, omitted for brevity.}

\subsubsection{Syntax and Semantics for the Logic of our Belief in EHPO Conclusions} \label{app:sec:belief} 
The logic of belief is a type of modal logic, called doxastic logic \cite{hintikka1962doxastic}, where the modal operator $\mathcal{B}$ is used to express belief\footnote{Computer scientists do not tend to distinguish between the logic of knowledge (epistemic) and the logic of belief (doxastic)~\cite{segerberg1999logicofbelief}.} Different types of reasoners can be defined using axioms that involve $\mathcal{B}$ \cite{smullyan1986belief}.

We can formulate the doxastic logic of belief in Backus-Naur form:

For any atomic proposition $P$, we define recursively a well-formed formula $\phi$ as
\begin{align} \label{app:eq:beliefsyntax}
    \phi \coloneqq P\;|\;\lnot \phi\;|\;\phi \land \phi\;|\; \mathcal{B}\phi
\end{align}

where $\mathcal{B}$ means ``It is believed that $\phi$". We interpret this recursively where $p$ is the base case, meaning that $\phi$ is p if it is an atom, $\lnot \phi$ is well-formed if $\phi$ is well-formed. We can also define $\lor$, $\rightarrow$, and $\leftrightarrow$ from $\lnot$ and $\land$, as in propositional logic.

As stated in the belief logic portion of Section \ref{sec:formalizing}, we model a consistent Type 1 reasoner \cite{smullyan1986belief}, which has access to all of propositional logic, has their beliefs logical closed under \emph{modus ponens}, and does not derive contradictions. In other words, we have the following axioms:

\begin{align*}
    \lnot ( \mathcal{B}p \land \mathcal{B}\lnot p )\equiv \mathcal{B}p \rightarrow \lnot \mathcal{B} \lnot p
\end{align*}
which is the consistency axiom,
\begin{align*}
    \vdash p \rightarrow \mathcal{B} p
\end{align*}
which is akin to Necessitation above in Section \ref{app:sec:kripkeaxioms} and means that we believe all tautologies, and
\begin{align*}
    \mathcal{B}(p \rightarrow q) \rightarrow (\mathcal{B}p \rightarrow \mathcal{B}q)
\end{align*}
which means that belief distributes over implication. This notably does not include

\begin{align*}
    \mathcal{B}p \rightarrow p
\end{align*}

which essentially means that we do not allow for believing $p$ to entail concluding $p$. This corresponds to us actually wanting to run hyperparameter optimization before we conclude anything to be true. We do not just want to conclude something to be true based only on \emph{a priori} information. This is akin to picking folkore parameters and concluding they are optimal without running hyperparameter optimization. 

\subsubsection{Combining Logics} \label{app:sec:multimodal}
It is a well known result that we can combine modal logics to make a multimodal logic \cite{scott1970multimodal}. In particular, we refer the reader to results on \emph{fusion} \cite{thomason1984fusion}.

For a brief intuition, we are able to combine our EHPO logic with belief logic since we are operating over the same set of possible worlds. The results of running EHPO produce a particular possible world, to which we apply our logic of belief in order to reason about the conclusions drawn in that world.

\subsubsection{Our Combined, Multimodal Logic and Expressing Hyperparameter Deception}
We develop the following multimodal logic, which we also state in Section \ref{sec:formalizing}:

\[
    \psi \coloneqq P\;|\;\lnot \psi\;|\;\psi \land \psi \;|\; \possible_t \psi \;|\; \mathcal{B} \psi
\]

\subsubsection{Axioms} \label{app:sec:multimodalaxioms}

We give this multimodal logic semantics to express our $t$-non-deceptiveness axiom, which we repeat below for completeness:

For any formula $p$,
\begin{equation*}
    \lnot \left( \possible_t \mathcal{B} p \land \possible_t \mathcal{B} \lnot p \right)
\end{equation*}

We can similarly express the $t$-non-deceptiveness axiom:

For any formula $p$,
\begin{equation*}
    \possible_t \mathcal{B}p \rightarrow \lnot \possible_t \mathcal{B} \lnot p 
\end{equation*}

We can also express a $t$-deceptiveness-axiom:

For any formula p,
\begin{equation*}
    \possible_t \mathcal{B} p \land \possible_t \mathcal{B} \lnot p
\end{equation*}

To reiterate, \emph{multimodal} just means that we have multiple different modes of reasoning, in this case our $\possible_t$ semantics for the demon doing EHPO and our consistent Type 1 reasoner operator $\mathcal{B}$.

Given a reasonable maximum time budget $t$, we say that EHPO is $t$-non-deceptive if it satisfies all of axioms above. Moreover, based on this notion of $t$-non-deceptiveness, we can express what it means to have a defense to being deceived. 

\subsubsection{Some notes on strength of belief and belief update} \label{app:sec:beliefstrength}

There are potentially interesting connections between our work on defending against hyperparameter deception and belief update \cite{ferme2011agmreview}. Notably, one could view our notion of skeptical belief as related to work done on "strength of belief" and belief update, or dynamic doxastic logic \cite{segerberg1999logicofbelief, benthem2009dup, kooi2003pdel}. Instead of picking an EHPO runtime \emph{a priori} and then running a defended EHPO and at the end evaluating whether or not we believe the conclusions we draw, we could iteratively update and test our belief and terminate if a certain belief threshold is met. In such quantitative theories of belief change, the degree of acceptance of a sentence is represented by a numerical value. Those numerical values can be updated in light of new information (so-called ``soft" information updates) \cite{benthem2007delbr, baltag2008pdbr}. Exploring this is out of scope for our work here, but could be an interesting future research direction for how to reason about empirical results that imply inconsistent conclusions.

\section{Section \ref{sec:defense} Appendix: Additional Notes on Defenses} \label{app:sec:defense}

\subsection{Proving a defended reasoners}

Suppose that we have been drawing conclusions using some ``naive'' belief operator $\mathcal{B}_{\text{n}}$ (based on a conclusion function $\mathcal{F}_{\text{n}}$) that satisfies the axioms of Section~\ref{sec:belieflogic}. We want to use $\mathcal{B}_{\text{n}}$ to construct a new operator $\mathcal{B}_{*}$, which is guaranteed to be deception-free.
One straightforward way to do this is to define the belief operator $\mathcal{B}_{*}$ such that for any statement $p$,
\[
    \mathcal{B}_{*} p \; \equiv \; \mathcal{B}_{\text{n}} p \;\land\; \lnot \possible_t \mathcal{B}_{\text{n}} \lnot p.
\]
That is, we conclude $p$ only if both our naive reasoner would have concluded $p$, and it is impossible for an adversary to get it to conclude $\lnot p$ in time $t$.
This enables us to show $t$-non-deceptiveness, following directly from the axioms in a proof by contradiction: Suppose $\mathcal{B}_{*}$ can be deceived, i.e. $\textcolor{red}{\possible_t \mathcal{B}_{*} p} \land \textcolor{red}{\possible_t \mathcal{B}_{*} \lnot p}$ is $\mathsf{True}$:

\begingroup
\setlength{\tabcolsep}{3.5pt} 
\renewcommand{\arraystretch}{1} 
\begin{table}[H]
\vspace{-.2cm}
\small
\begin{center}
      \centering
        \begin{tabular}{r c l c c}
\toprule
\; & \; & \; & \; & \textbf{Rule} \\
\midrule
$\textcolor{red}{\possible_t \mathcal{B}_{*} p}$ & $\equiv$ & $\possible_t \left( \mathcal{B}_{\text{n}} p \land \lnot \possible_t \mathcal{B}_{\text{n}} \lnot p \right)$   &  \; & \text{Applying $\possible_t$ to the definition of $\mathcal{B}_{*} p$ \; (\hyperlink{ax:defendedreasoner}{1})} \\

\midrule
\; & $\rightarrow$ & $\possible_t \left( \lnot \possible_t \mathcal{B}_{\text{n}} \lnot p \right)$  & \;   &  \text{Reducing a conjunction to either of its terms: $(a \land b) \rightarrow b$} \\

\midrule
\; & $\rightarrow$ & $\textcolor{blue}{\lnot \possible_t \mathcal{B}_{\text{n}} \lnot p}$  & \;   & \text{Symmetry; dropping all but the right-most operator: $\possible_t(\possible_t a)\rightarrow \possible_t a$} \\
\bottomrule
\end{tabular}
\end{center}
\vspace{-.5cm}
\end{table}

We provide more detail on these transformations than we do in the main text. The first application is simple; we just put parentheses around our definition of $\mathcal{B}_*$, and apply $\possible_t$ to it. The second step is also simple. We apply a change to whats inside the parentheses, i.e. just the definition of $\mathcal{B}_*$. Because this is a conjunction, in order for it to be true, both components have to be true. So, we can reduce the conjunction to just it's second term.

The part that is more unfamiliar is the application of Modal Axiom 5 (Symmetry) to reduce the number of $\possible_t$ operators. We provide this example above in Section \ref{app:sec:logic}, where we explain why Modal Axiom 5 holds for our EHPO logic semantics. We reiterate here for clarity:

While the intuitive English reading (``It's possible that it's not possible that $p$") does not seem equivalent to this reduction (``It's not possible that $p$), it is in fact valid for our semantics. Think of this in terms of the demon. If $p$ cannot be brought about in time $t$ in expectation (where $t$ is a reasonable upper bound on compute time), then that's the end of it; it doesn't matter which operators come before it (any number of $\possible_t$ or $\necessary_t$). Adding possibility or necessity before that condition doesn't change that fact that it, for that upper bound $t$, it is not possible to bring about $p$.

We then pause to apply our axioms to the right side of the conjunction, $\textcolor{red}{\possible_t \mathcal{B}_{*}\lnot p}$ :

\begingroup
\setlength{\tabcolsep}{3pt} 
\renewcommand{\arraystretch}{1} 
\begin{table}[H]
\vspace{-.2cm}
\small
\begin{center}
      \centering
        \begin{tabular}{r c l c c}
\toprule
\; & \; & \; & \; & \textbf{Rule} \\
\midrule
$\textcolor{red}{\possible_t \mathcal{B}_{*} \lnot p}$ & $\equiv$ & $\possible_t \left( \mathcal{B}_{\text{n}} \lnot p \land \lnot \possible_t \mathcal{B}_{\text{n}} p \right)$   &  \; & \text{Applying $\possible_t$ to the definition of $\mathcal{B}_{*} \lnot p$ \; (\hyperlink{ax:defendedreasoner}{1})} \\

\midrule
\; & $\rightarrow$ & $\possible_t\mathcal{B}_{\text{n}} \lnot p \land \possible_t\lnot \possible_t \mathcal{B}_{\text{n}} p$  & \;   &  \text{Distributing $\possible_t$ over $\land$: $\possible_t (a \land b) \rightarrow (\possible_t a \land \possible_t b)$} \\

\midrule
\; & $\rightarrow$ & $\textcolor{blue}{\possible_t \mathcal{B}_{\text{n}} \lnot p}$  & \;   & \text{Reducing a conjunction to either of its terms: $(a \land b) \rightarrow a$} \\
\bottomrule
\end{tabular}
\end{center}
\vspace{-.35cm}
\end{table}

This transformation is much like the one above. We similarly apply $\possible_t$ to the definition of $\mathcal{B}_*$.

We then distribute $\possible_t$ over the definition, which holds for our logic since possibility distributes over and. We prove this for our logic in Section \ref{app:sec:logic}, and provide an intuitive explanation here. If it is possible in time $t$ to bring about a particular formula, then it must also be possible to bring about the sub-conditions that compose that formula in time $t$. If this were not the case, then we would not be able to satisfy bringing about the whole formula in time $t$.

Lastly, as in the first example, we reduce the conjunction to one of its terms (this time taking the first, rather than the second).

We now bring both sides of the conjunction back together: $\textcolor{red}{\possible_t \mathcal{B}_{*} p} \land  \textcolor{red}{\possible_t \mathcal{B}_{*} \lnot p}
\; \equiv \;
\textcolor{blue}{\lnot \possible_t \mathcal{B}_{\text{n}} \lnot p} \land  \textcolor{blue}{\possible_t \mathcal{B}_{\text{n}} \lnot p}$.
The \textcolor{blue}{right-hand side} is of the form $\lnot a \land a$, which must be $\mathsf{False}$. This contradicts our initial assumption that $\mathcal{B}_{*}$ is $t$-deceptive (i.e., $\textcolor{red}{\possible_t \mathcal{B}_{*} p} \land \textcolor{red}{\possible_t \mathcal{B}_{*} \lnot p}$ is $\mathsf{True}$). Therefore, $\mathcal{B}_{*}$ is $t$-non-deceptive. 

\subsection{Theoretically Validating Defenses to Hyperparameter Deception}

We prove Theorem \ref{thm:defendedhpo}:

\begin{theorem}
Suppose that the set of allowable hyper-HPs $\mathcal{C}$ of $H$ is constrained, such that any two allowable random-search distributions $\mu$ and $\nu$ have Renyi-$\infty$-divergence at most a constant, i.e. $D_{\infty}(\mu \| \nu) \le \gamma$. The $(K,R)$-defended random-search EHPO of Definition \ref{def:defended_random} is guaranteed to be $t$-non-deceptive if we set $R \ge \sqrt{t \exp(\gamma K)/K} = O(\sqrt{t})$.
\end{theorem}

Suppose we are considering HPO via random search \cite{bergstra2011algorithms}, in which the set of allowable hyper-hyperparameters contains tuples $(\mu, M)$, where $\mu$ is a distribution over all possible hyperparameter sets $\Lambda$ and $M$ is the number of different hyperparameter configuration trials to run. This set $S$ is the Cartesian product of the set of allowable distributions $D$ ($\mu \in D$) and $M$.

Suppose that for any two allowable distributions $\mu$, $\nu \in D$ and any event $A$ (a measurable subset of $\Lambda$), $\mu(A) \le e^\gamma \cdot \nu(A)$ (i.e., the Renyi $\infty$-divergence between any pair of distributions is at most $\gamma$).
This bounds how much the choice of hyper-hyperparameter can affect the hyperparameters in HPO.

We also suppose we start from a naive reasoner (expressed via the operator $\mathcal{B}_{\text{n}}$), which draws conclusions based on a log with $K$ trials. For this scenario, we are only concerned with the reasoner's conclusions from $K$-trial logs.  We therefore assume w.l.o.g. that the reasoner draws no conclusions unless presented with exactly one log with exactly $K$ trials.

For some constant $R \in \mathbb{N}$, we
construct a new reasoner $\mathcal{B}_{*}$ that does the following:
It draws conclusions only from a single log with exactly $KR$ trials (otherwise it concludes nothing).
To evaluate a proposition $p$, it splits the log into $R$ groups of $K$ trials each, evaluates $\mathcal{B}_{n}$ on $p$ on each of those $R$ groups, and then concludes $p$ only if $\mathcal{B}_{n}$ also concluded $p$ on all $R$ groups.

Now consider a particular (arbitrary) proposition $p$.
Since $\mathcal{B}_{*}$ draws conclusions based on only a single log, any strategy $\sigma$ for the demon is equivalent to one that maintains at most one log at all times (the ``best'' log it found so far for its purposes, as it can discard the rest).

Let $Q$ be the supremum, taken over all allowable distributions $\mu$, of the probability that running a group of $K$ random search trials using that distribution will result in a log that would convince the $\mathcal{B}_{n}$ of $p$. Similarly, let $Q_{\lnot}$ be the supremum, taken over all allowable distributions $\nu$, of the probability that running a group of $K$ trials using that distribution will result in a log that would convince $\mathcal{B}_{n}$ of $\lnot p$.

Observe that $Q$ is the probability of an event in a product distribution of $K$ independent random variables each distributed according to $\mu$, and similarly for $Q_{\lnot}$, and the corresponding events are disjoint. By independent additivity of the Renyi divergence, the Renyi $\infty$-divergence between these corresponding product measures will be $\gamma K$.
It follows that 

\[
1 - Q \ge \exp(-\gamma K) Q_{\lnot}
\]

and

\[
1 - Q_{\lnot} \ge \exp(-\gamma K)  Q
\]

From here it's fairly easy to conclude that
\[
    Q + Q_{\lnot} \le \frac{2}{1 + \exp(-\gamma K) }.
\]

Now, an EHPO procedure using random search with $KR$ trials will convince $\mathcal{B}_{*}$ of $p$ with probability $Q^R$, since all $R$ independently sampled groups of $K$ trials must ``hit'' and each hit happens with probability $Q$. Thus, the expected time it will take the fastest strategy to convince us of $p$ is $Q^{-R} \cdot KR$.
Similarly, the fastest strategy to convince us of $\lnot p$ takes expected time $Q_{\lnot}^{-R} \cdot KR$.

Suppose now, by way of contradiction, that the $t$-non-deceptiveness axiom is violated, and there are strategies that can achieve both of these in time at most $t$.
That is,
\[
    Q^{-R} \cdot KR \le t
    \hspace{2em}\text{and}\hspace{2em}
    Q_{\lnot}^{-R} \cdot KR \le t.
\]
From here, it's fairly easy to conclude that
\[
    Q + Q_{\lnot} \ge 2 \left( \frac{KR}{t} \right)^{1/R}.
\]
Combining this with our conclusion above gives
\[
    \left( \frac{KR}{t} \right)^{1/R} \le \frac{1}{1 + \exp(-\gamma K) }.
\]
It's clear that we can cause this to be violated by setting $R$ to be large enough.
Observe that
\[
    \frac{1}{1 + \exp(-\gamma K) } \le \exp(-\exp(-\gamma K)),
\]
so
\[
    \left( \frac{KR}{t} \right)^{1/R} \le \exp(-\exp(-\gamma K)).
\]
Taking the root of both sides gives
\[
    \left( \frac{KR}{t} \right)^{\frac{1}{R \exp(-\gamma K)}} \le \frac{1}{e}.
\]
To simplify this expression, let $\beta$ denote
\[
    \beta = R \exp(-\gamma K).
\]
So that
\[
    \left( \frac{\beta K}{t \exp(-\gamma K)} \right)^{1/\beta} \le \frac{1}{e}.
\]
Finally, we set $R$ such that
\[
    \beta = \sqrt{\frac{t \exp(-\gamma K)}{K}}.
\]
To give
\[
    \left( \frac{1}{\beta} \right)^{1/\beta} \le \frac{1}{e}.
\]
But this is impossible, as the minimum of $x^x$ occurs above $1/e$.
This setting of $R$ gives
\[
    R = \beta \exp(\gamma K) = \sqrt{\frac{t \exp(\gamma K)}{K}} = O(\sqrt{t}).
\]
This shows that, for this task, if we run our constructed EHPO with $R = O(\sqrt{t})$ assigned in this way, it will be guaranteed to be $t$-non-deceptive.

\newpage
\subsection{Defense Experiments}\label{app:sec:experiments:defense}

In this section we provide more information about the implementation of a random-search-based defense to hyperparameter deception in \citet{wilson2017marginal}, which we discuss in Section \ref{sec:defense}. 

\subsubsection{Our Implemented Defense Algorithm} \label{app:sec:phased}

The defense we implement in our experiments is a bit different than what we describe in our theoretical results in Section \ref{sec:defense}. In particular, in practice it is easier to implement subsampling rather than resampling.

\textbf{Protocol of Selecting Hyper-HPs.}
As partially illustrated in Figure~\ref{fig:wilson_deception} and elaborated on in Appendix \ref{app:sec:prelim}, \citet{wilson2017marginal}'s choice of hyper-HPs does not capture the space where Adam effectively simulates Heavy Ball. In \citet{wilson2017marginal}i, Adam-specific HPs like numerical variable $\epsilon=10^{-8}$ \cite{kingma2014adam} are treated as constants, leading to a biased HP-search space.

In contrast, we select the hyper-HPs of $\epsilon$ following a dynamic searching protocol: 

Inspired by \cite{choi2019empirical}, we start from a wide range $\epsilon\in[10^{-12}, 10^{12}]$ as a wide search space. We iteratively select powers-of-10 grids that are uniformly spaced in the logarithmic scale of the current range. For instance, the selected grids for the prior range would be $\{10^{-12}, 10^{-11}, \cdots, 10^{11}, 10^{12}\}$. We perform a single run on each grid selected, and shrink the range towards grids where the best performance are achieved. The shrinkage follows the policy of either $\times 10$ to the lower boundary or $\times 0.1$ to the upper boundary. For example, for the prior range, we found the best performance is achieved on grid $10^{11}$, so we multiply the lower boundary $10^{-12}$ with $10$ and shift the range to $\epsilon\in[10^{-11}, 10^{12}]$.
Our protocol terminates with $\epsilon\in[10^{10}, 10^{12}]$ as the final hyper-HPs that we use for our defended random search EHPO. 

\textbf{Scaling Learning Rate $\eta$.}
Note that directly applying the hyper-HP of $\epsilon\in[10^{10}, 10^{12}]$ to Adam would lead to extremely slow convergence, since essentially large $\epsilon$ indicates a small effective learning rate $\eta$. Similar to \citet{choi2019empirical}, we explore a shifted hyper-HPs for the $\eta$, scaled proportionally with $\epsilon$. Specifically, note that a large $\epsilon$ would make the update of Adam approach the update rule in the Heavy Ball method; for any randomly selected $\epsilon\in[10^{10}, 10^{12}]$, we perform the random search of $\eta/\epsilon$ instead of $\eta$ itself in the space of $[0.5,0.7]$, which is the search space of HB's learning rate shown in \cite{wilson2017marginal}.

\subsubsection{Experimental setup}
We follow the setup from \cite{wilson2017marginal}, where the details are specified in Section~\ref{sec:wilson:setup}.

\subsubsection{Code}

The code for running these experiments can be found at \url{https://github.com/pasta41/deception}.

\subsubsection{Associated results and logs}
In line with our notion of a log, we provide heatmaps of our logs in Figures~\ref{fig:defended-ehpo-logs-adam-sgd}, \ref{fig:defended-ehpo-logs-adam-hb} and  that correspond with our results in Section \ref{sec:defense}. We note that the performance of Heavy Ball for some random seeds is very bad (e.g., $10\%$ test accuracy). The performance varies widely -- also nearing $92\%$ for different random seeds. We affirm that this is the search space that yields the best results for Heavy Ball ($~92\%$). 

The results for Heavy Ball exhibit large variance. This illustrates a strength of our defensed: it actually helps with robustness against potentially making the wrong conclusion about Heavy Ball's performance (more generally), due to not making conclusions off of a single result (and perhaps using a random seed for which performance is particularly bad). We make a different claim about relative algorithm performance than \citet{wilson2017marginal} about Heavy Ball (i.e., we do not claim that it is better than Adam); but we do not reach this conclusion for the wrong reason (i.e., that we got one bad Heavy Ball result for a particular random seed). 

\pagebreak

\begin{figure}[H]
    \centering
    \includegraphics[scale=.5]{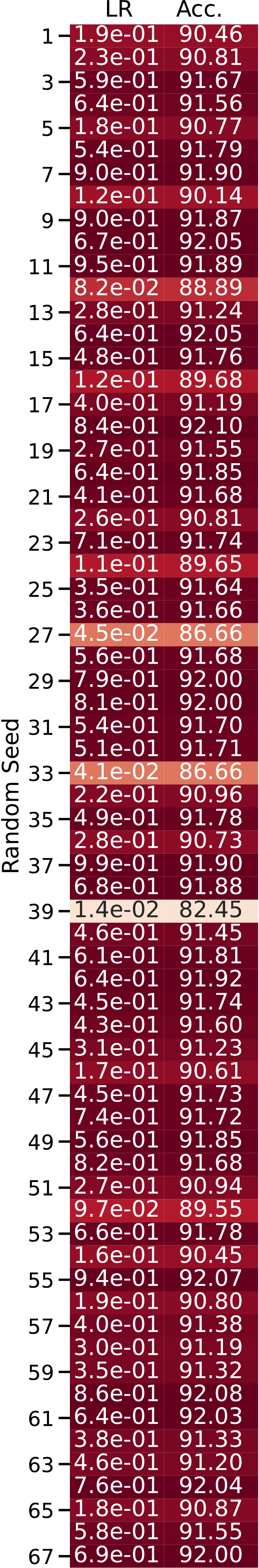} \hspace{2mm}
\includegraphics[scale=.5]{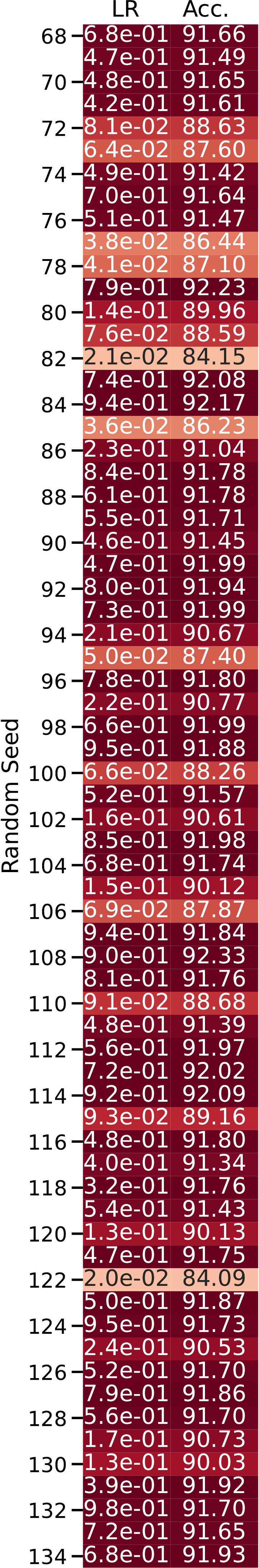} \hspace{2mm}
\includegraphics[scale=.5]{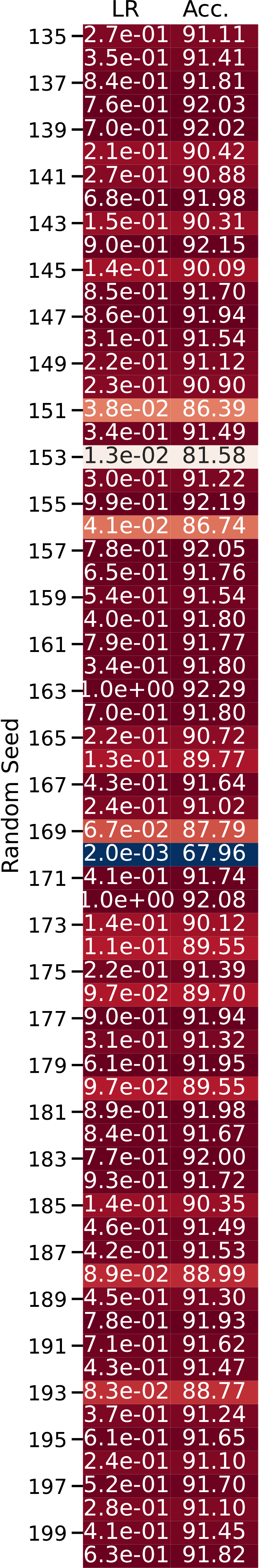} \hspace{2mm}
    \caption{Heatmap logs of SGD defended random search. Redder rows indicate higher test accuracy.}
    \label{fig:defended-ehpo-logs-sgd-only}
\end{figure}

\pagebreak

\begin{figure}[H]
    \centering
    \includegraphics[scale=.36]{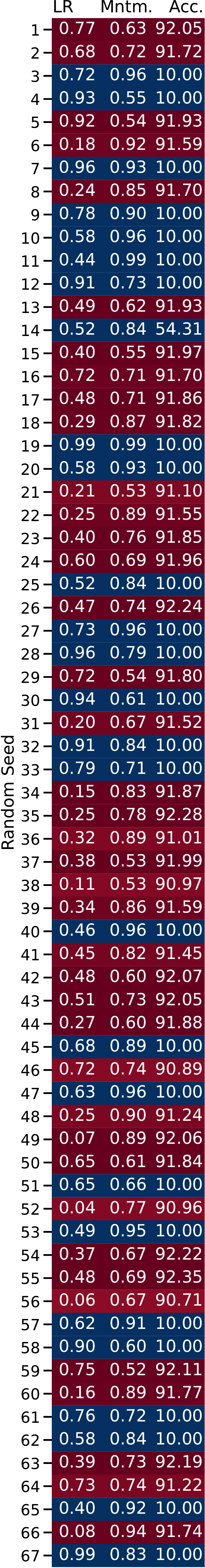} \hspace{2mm}
\includegraphics[scale=.36]{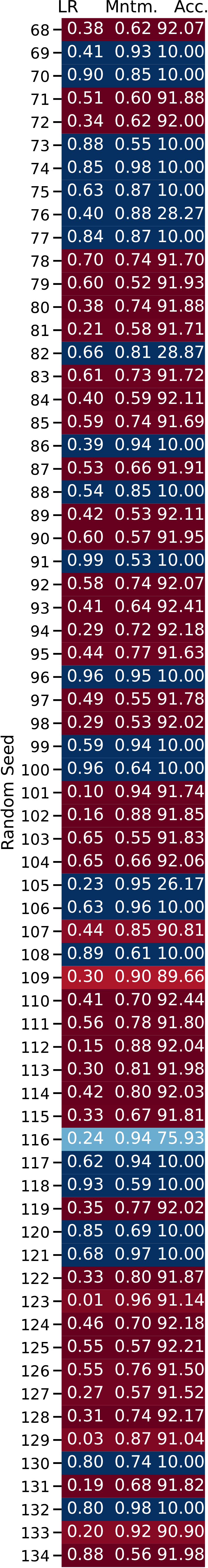} \hspace{2mm}
\includegraphics[scale=.36]{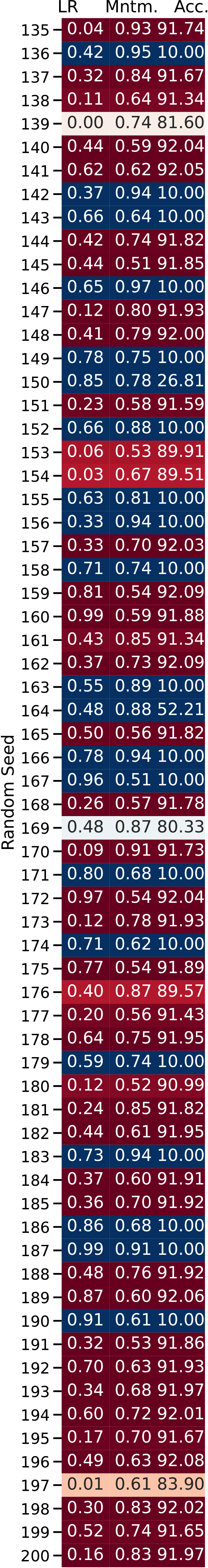} \hspace{2mm}
    \caption{Heatmap logs of Heavy Ball (HB) defended random search. Redder rows indicate higher test accuracy.}
    \label{fig:defended-ehpo-logs-hb-only}
\end{figure}

\pagebreak

\begin{figure}[H]
    \centering
    \includegraphics[scale=.35]{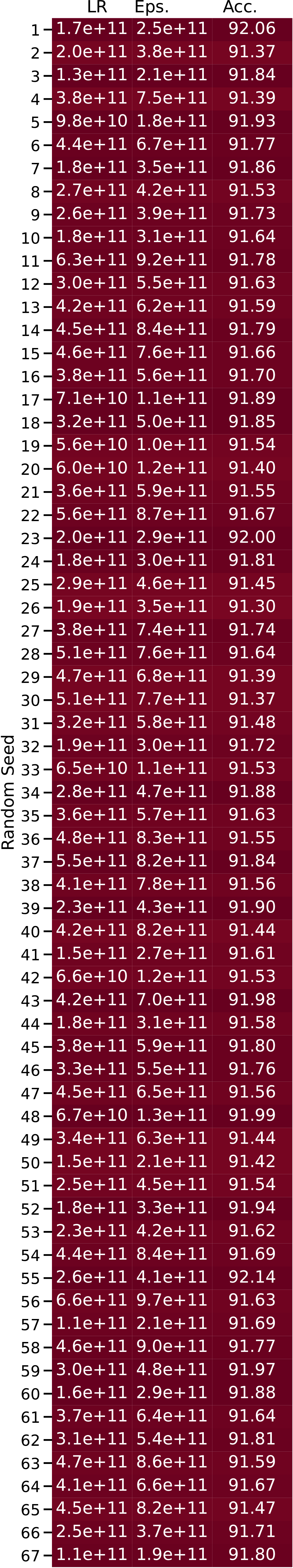} \hspace{2mm}
\includegraphics[scale=.35]{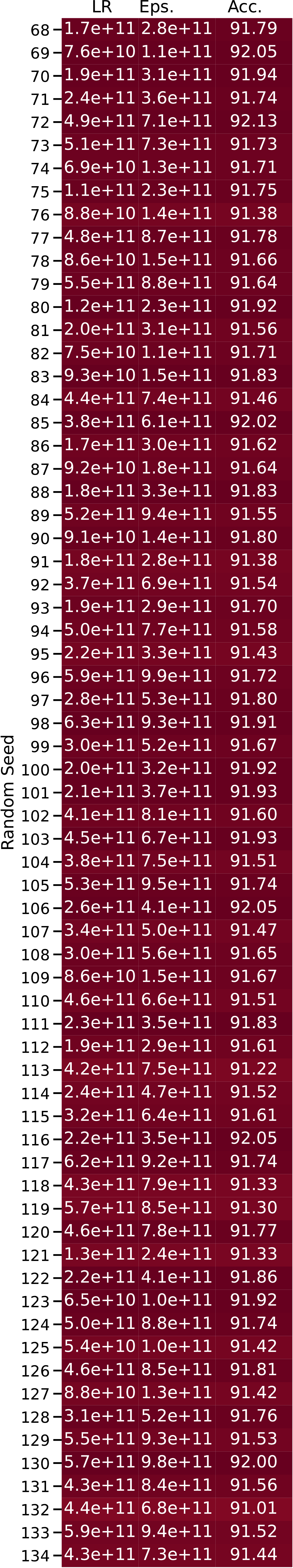} \hspace{2mm}
\includegraphics[scale=.35]{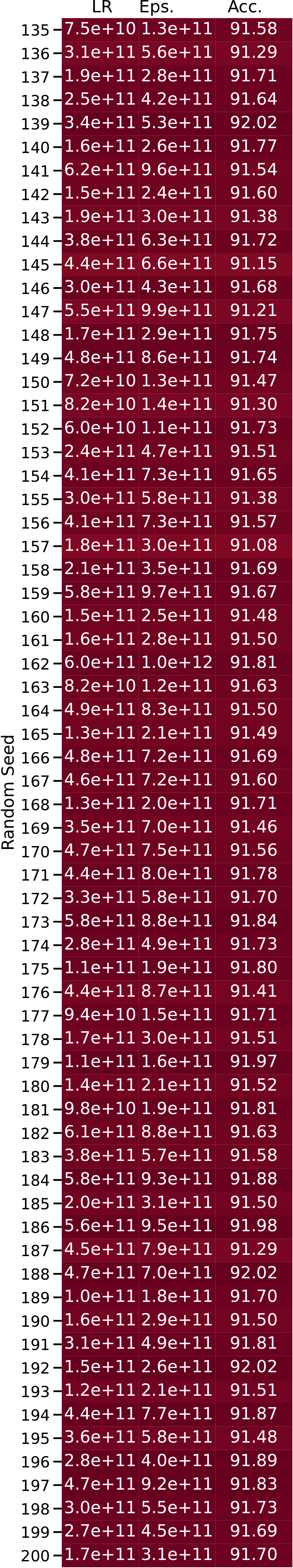} \hspace{2mm}
    \caption{Heatmap logs of Adam defended random search. Redder rows indicate higher test accuracy.}
    \label{fig:defended-ehpo-logs-adam-only}
\end{figure}

\pagebreak

\begin{figure}[ht]
    \centering
    \includegraphics[scale=.5]{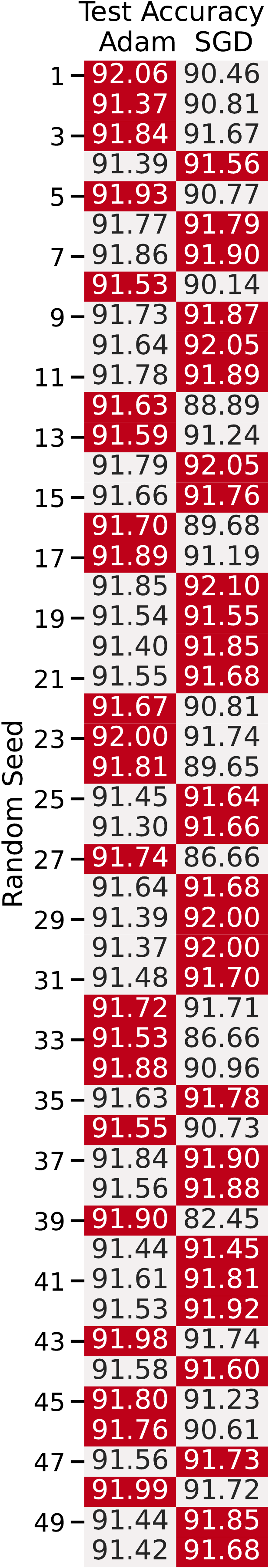} \hspace{2mm}
\includegraphics[scale=.5]{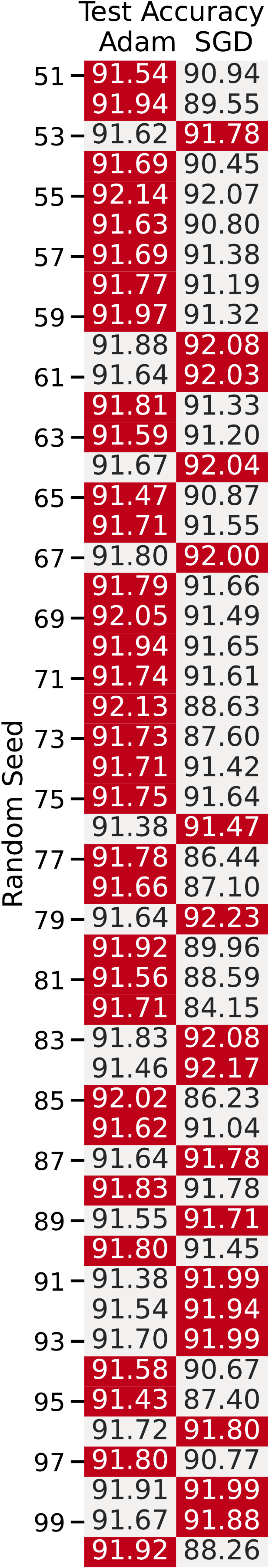} \hspace{2mm}
\includegraphics[scale=.5]{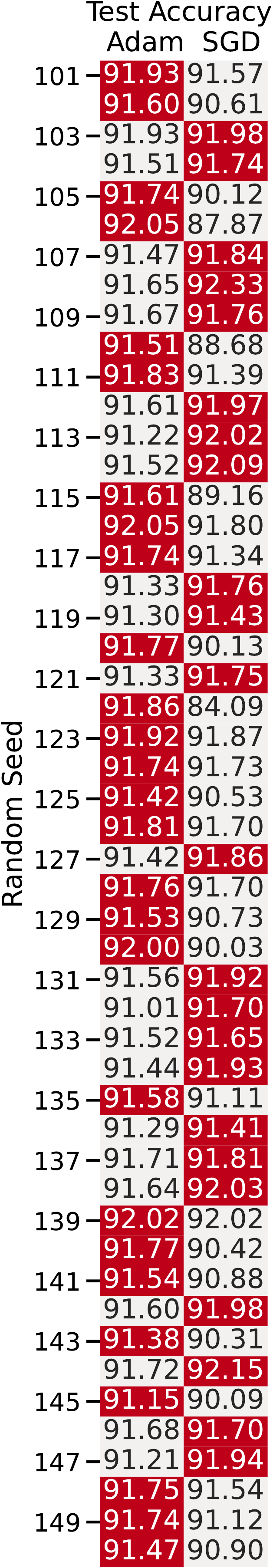} \hspace{2mm}
\includegraphics[scale=.5]{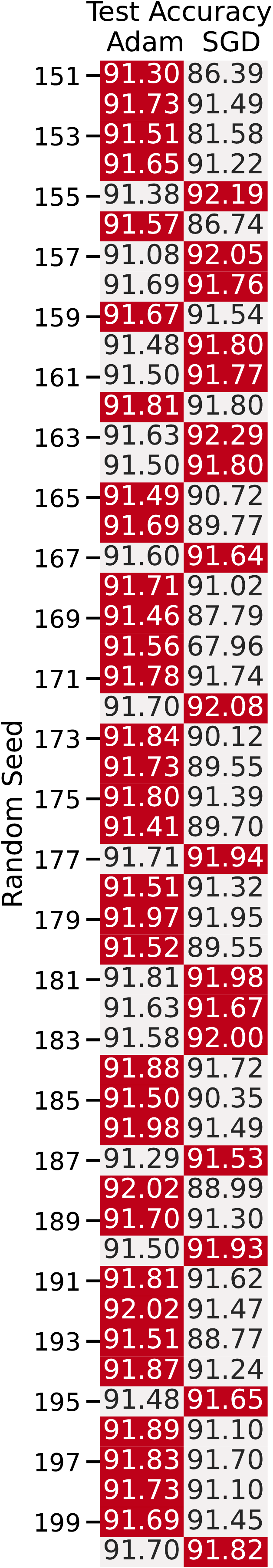} 
    \caption{Heatmap logs of test accuracy of VGG-16 on CIFAR-10 for Adam vs. SGD using our defended random search EHPO.  Red indicates higher test accuracy for the given random seed.}
    \label{fig:defended-ehpo-logs-adam-sgd}
\end{figure}

\pagebreak

\begin{figure}[ht]
    \centering
    \includegraphics[scale=.5]{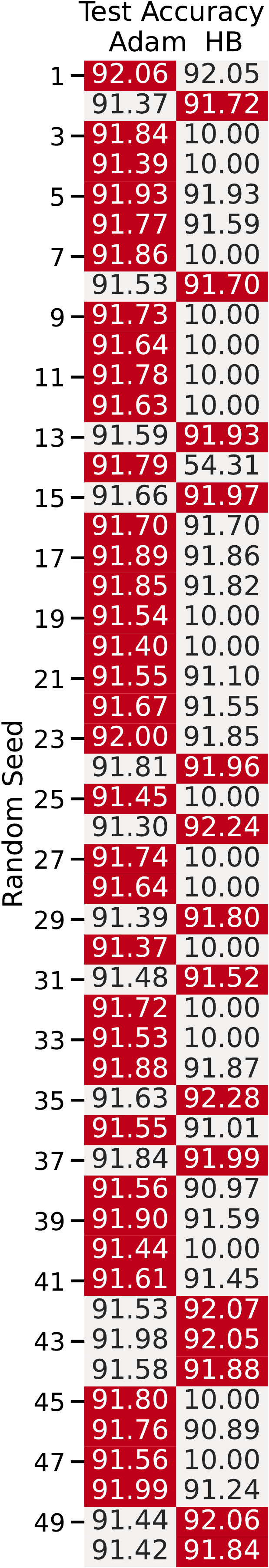} \hspace{2mm}
\includegraphics[scale=.5]{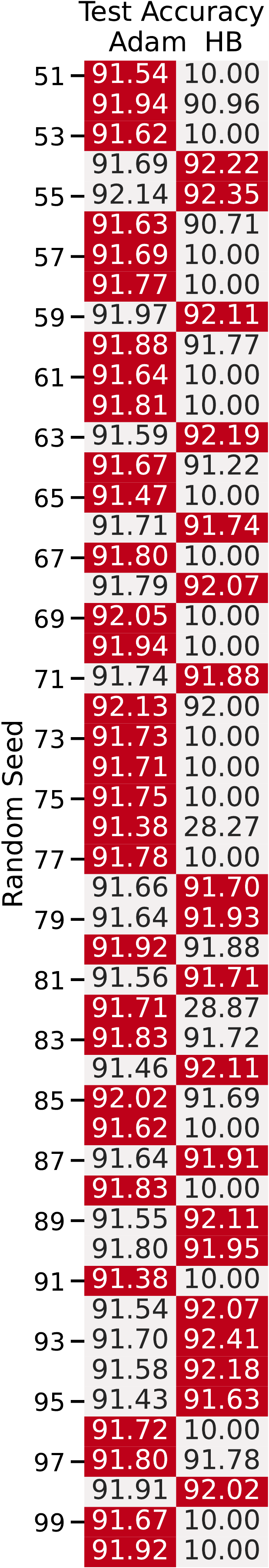} \hspace{2mm}
\includegraphics[scale=.5]{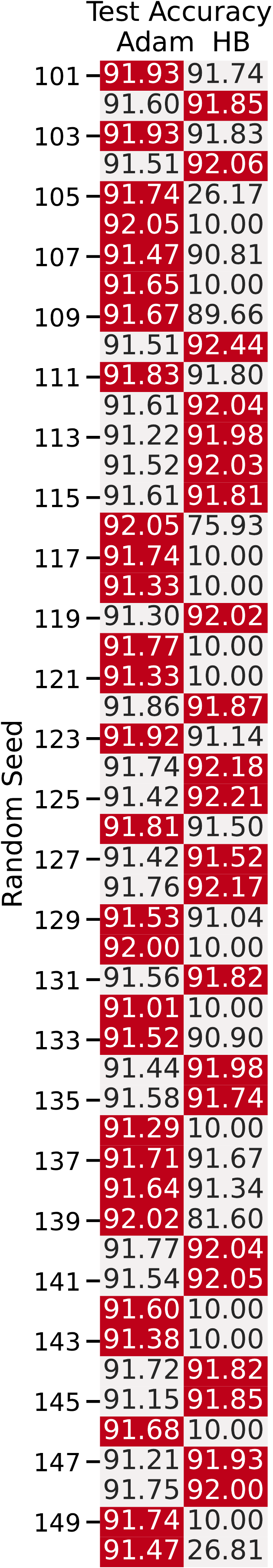} \hspace{2mm}
\includegraphics[scale=.5]{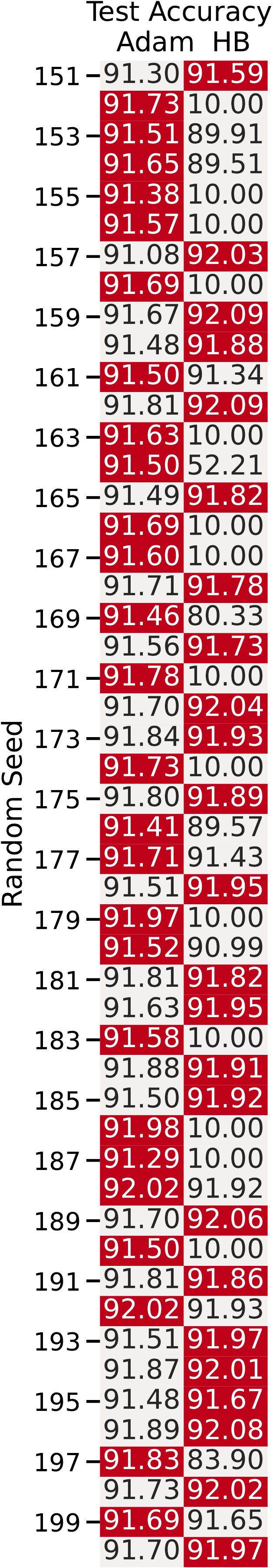} 
    \caption{Heatmap logs of test accuracy of VGG-16 on CIFAR-10 for Adam vs. Heavy Ball (HB) using our defended random search EHPO. Red indicates higher test accuracy for the given random seed.}
    \label{fig:defended-ehpo-logs-adam-hb}
\end{figure}

\newpage
\section{Section \ref{sec:conclusion} Appendix: Additional Notes on Conclusion}

\subsection{Additional Practical Takeaways}

In our conclusion in Section \ref{sec:conclusion}, we note the following practical takeaways:
\begin{itemize}[topsep=0pt, leftmargin=.25cm]
    \item \textbf{Researchers should have their own notion of skepticism, appropriate to their specific task.} There is no one-size-fits-all defense solution. Our results are \emph{\textbf{broad insights}} about defended EHPO: A defended EHPO is \emph{\textbf{always possible}}, but finding an efficient one will depend on the task.
    \item \textbf{Researchers should make explicit how they choose hyper-HPs.} 
    What is reasonable is ultimately a function of what the ML community accepts. Being explicit, rather than eliding hyper-HP choices, is essential for helping decide what is reasonable. As a heuristic, we recommend setting hyper-HPs such that they include HPs for which the optimizers' performance starts to degrade, as we do above. 
    \item \textbf{Avoiding hyperparameter deception is just as important as reproducibility}. We have shown that reproducibility \cite{henderson2017reinforcement, pineau2019checklist, gundersen2018reproducibility, bouthillier19reproducible, Sinha2020-aw} is only part of the story for ensuring reliability. While necessary for guarding against brittle findings, it is not sufficient. We can replicate results---even statistically significant ones---that suggest conclusions that are altogether wrong.
\end{itemize}

We elaborate here that our defended random search EHPO indicates a particular form of skepticism that may (or may not) be appropriate to different ML tasks. That is, we suggest a defended EHPO, but do not claim that that EHPO is optimal or suited for all tasks. Even though it may not be optimal, the guarantees it affords would translate to other tasks (so long as the assumption is maintained that there is an upper bound on how much the hyper-HPs can control the HPs). So, while we do not necessarily encourage practitioners to use our particular defended EHPO, we do not discourage it either. The main take away is that practitioners should develop their own notion of skepticism (appropriate to their particular task) and be explicit about the assumptions they rely on when selecting hyper-HPs. The way one chooses hyper-HPs should be defensible.

When in doubt, as a heuristic we recommend using a search space that includes where an algorithm's performance starts to degrade (to be assured that a maximum, even if a local one, has been found). We refer to our dynamic two-phase protocol (which we describe in detail in Appendix \ref{app:sec:phased}) as an example of how to do this. We first do a broad (but coarse) search. We used grid search for that initial sweep. Random search may be a better choice for some tasks. We were familiar with \citet{wilson2017marginal} (and many have written about it), and felt confident that grid search would capture the space well based on the results that others have also reported on this task. We then used this first sweep to determine which hyper-HPs we should use for our second, finer-grained sweep. We apply our more expensive, defended EHPO for this second sweep, using the hyper-HPs we selected from the first sweep. In other words, we spent a bit of time/our compute budget justifying to ourselves that we were picking reasonable hyperparameters -- instead of just picking one grid or range for random search to sample, and hoping that our results would be representative of other search spaces.

\subsection{Broader Impact}

\label{app:sec:impact}
As we suggest in Section \ref{sec:defense}, our work can be considered as related to (but orthogonal with) with prior studies on reproducibility as advocating for more robust scientific practices in ML research. In particular, our work complements prior empirical studies that shine a light on reliability issues in ML---issues that relate particularly to traditionally underspecified choices in hyperparameter optimization \cite{choi2019empirical, sivaprasad2019optimizer, lipton2018troubling}. In contrast to this prior work, which illustrates the issue with experiments, we provide a theoretical contribution that enables ML practitioners and researchers to defend against unreliable, inconsistent HPO. We provide a theoretically-backed mechanism to promote and facilitate more trustworthy norms and practices in ML research.

More broadly, our work can be understood as a mechanism for dealing with \emph{measurement bias}---the misalignment between what one intends to measure and what they are actually measuring---for overall ML algorithm performance. While alleviating measurement bias is by no means novel to more mature branches of science \cite{gould1996mismeasure}, including other fields of computing \cite{mytkowicz2009pl}, until recently it has been under-explored in ML. Beginning in the last couple of years, measurement bias is now coming under increased scrutiny with respect to the origins of empirical gains in ML \cite{musgrave2020metric, forde2021actionable}. In current work, it is often difficult to disentangle whether the concluded measured performance gains are due to properties of the training algorithm or to fortuitous HP selection. Our formalization, rather than allowing HPO choices to potentially obscure empirical results, provides confidence in the conclusions we can draw about overall algorithm performance.

Our work also highlights how there is a human element, not a just statistical one, to bias in ML pipelines: Practitioners make decisions about HPO that can heavily influence performance (e.g., choice of hyper-hyperparameters). The human element of biasing solution spaces has been discussed in sociotechnical writing \cite{friedman1996bias, smith1985limits, cooper2021emergent}, in AI \cite{mitchell1980bias}, in the context of ``p-hacking'' or ``fishing expeditions" for results that fit a desired pattern~\cite{gelman2019fishing}, and was also the focus of Professor \citet{neurips2020keynote}'s NeurIPS 2020 keynote. Formalizing the process for how to draw conclusions from HPO, as we do here, has the potential to alleviate the effects of this type of human bias in ML systems. 

Lastly, our insights concerning robustness also extend to growing areas in ML that use learning to guide hyperparemeter selection, such as meta-learning and neural architecture search \cite{elsken2018nas, zoph2017nas, automl_book, neurips2020meta,icml2020automl}. While the assisting learning agents in those methods guide choosing hyperparameters for the trained output model, their own hyperparameters tend to be either manually tuned or chosen with more traditional HPO strategies, like grid search \cite{zhong2018practical}. In other words, these processes can exhibit the bias problem discussed above and are therefore potentially subject to hyperparameter deception, which can be mitigated by the work we present here.

\end{document}